\documentclass{article} % For LaTeX2e
\usepackage{iclr2026_conference,times}

\usepackage{amsmath,amsfonts,bm}

\def\eqref#1{equation~\ref{#1}}
\def\1{\bm{1}}

\DeclareMathAlphabet{\mathsfit}{\encodingdefault}{\sfdefault}{m}{sl}
\SetMathAlphabet{\mathsfit}{bold}{\encodingdefault}{\sfdefault}{bx}{n}

\usepackage{microtype}
\usepackage{amsmath}
\usepackage{amssymb}
\usepackage{amsfonts}
\usepackage{amsthm}

\usepackage{xcolor}
\usepackage{graphicx}
\usepackage{float}
\usepackage{wrapfig}
\usepackage{subcaption}
\usepackage{colortbl}

\usepackage{booktabs}
\usepackage{array}
\usepackage{tabularx}
\usepackage{threeparttable}
\usepackage{multirow}
\usepackage{multicol}
\usepackage{makecell}
\usepackage{longtable}
\usepackage{ragged2e}
\usepackage{adjustbox}
\usepackage{caption}
\usepackage{calc}

\usepackage{enumitem}
\usepackage{nicefrac}

\usepackage{algorithm}
\usepackage{algpseudocode}

\usepackage[colorlinks=true, citecolor=blue]{hyperref}

\usepackage[most]{tcolorbox}
\usepackage{etoolbox}

\newtcolorbox{exbox}[1]{breakable, title=#1}
\usepackage{listings}
\usepackage{pgfplotstable}

\definecolor{lightblue1}{rgb}{0.97, 0.985, 1} 
\definecolor{lightblue2}{rgb}{0.92, 0.965, 1} 
\definecolor{lightblue3}{rgb}{0.84, 0.93, 1}
\definecolor{lightblue4}{rgb}{0.74, 0.87, 1}
\definecolor{lightblue5}{rgb}{0.64, 0.81, 1}
\definecolor{lightblue6}{rgb}{0.54, 0.75, 1}

\definecolor{lightgreen1}{rgb}{0.97, 1.00, 0.97}
\definecolor{lightgreen2}{rgb}{0.92, 0.98, 0.92}
\definecolor{lightgreen3}{rgb}{0.84, 0.95, 0.84}
\definecolor{lightgreen4}{rgb}{0.74, 0.91, 0.74}
\definecolor{lightgreen5}{rgb}{0.64, 0.86, 0.64}
\definecolor{lightgreen6}{rgb}{0.54, 0.81, 0.54}

\definecolor{lightorange1}{rgb}{1.00, 0.98, 0.95}
\definecolor{lightorange2}{rgb}{1.00, 0.95, 0.85}
\definecolor{lightorange3}{rgb}{1.00, 0.90, 0.70}
\definecolor{lightorange4}{rgb}{1.00, 0.85, 0.55}
\definecolor{lightorange5}{rgb}{1.00, 0.80, 0.40}
\definecolor{lightorange6}{rgb}{1.00, 0.75, 0.30}

\definecolor{lightpurple1}{rgb}{0.985, 0.97, 1.00}
\definecolor{lightpurple2}{rgb}{0.96, 0.92, 1.00}
\definecolor{lightpurple3}{rgb}{0.93, 0.84, 1.00}
\definecolor{lightpurple4}{rgb}{0.87, 0.74, 1.00}
\definecolor{lightpurple5}{rgb}{0.81, 0.64, 1.00}
\definecolor{lightpurple6}{rgb}{0.75, 0.54, 1.00}

\definecolor{lightred1}{rgb}{1.00, 0.97, 0.97}
\definecolor{lightred2}{rgb}{1.00, 0.92, 0.92}
\definecolor{lightred3}{rgb}{1.00, 0.84, 0.84}
\definecolor{lightred4}{rgb}{1.00, 0.74, 0.74}
\definecolor{lightred5}{rgb}{1.00, 0.64, 0.64}
\definecolor{lightred6}{rgb}{1.00, 0.54, 0.54}

\definecolor{lightcyan1}{rgb}{0.97, 1.00, 1.00}
\definecolor{lightcyan2}{rgb}{0.92, 0.98, 0.98}
\definecolor{lightcyan3}{rgb}{0.84, 0.95, 0.96}
\definecolor{lightcyan4}{rgb}{0.74, 0.91, 0.94}
\definecolor{lightcyan5}{rgb}{0.64, 0.87, 0.92}
\definecolor{lightcyan6}{rgb}{0.54, 0.83, 0.90}

\title{Co-rewarding: Stable Self-supervised RL for Eliciting Reasoning in Large Language Models}
\author{
\begin{tabular}{@{}l@{}}
\textbf{Zizhuo Zhang}$^{1}$\thanks{Equal contribution.}        \quad
\textbf{Jianing Zhu}$^{1*}$ \quad
\textbf{Xinmu Ge}$^{2,3*}$ \quad
\textbf{Zihua Zhao}$^{3*}$ \quad
\textbf{Zhanke Zhou}$^{1}$ \\
\textbf{Xuan Li}$^{1}$ \quad
\textbf{Xiao Feng}$^{1}$ \quad
\textbf{Jiangchao Yao}$^{3}$\thanks{Correspondence to Bo Han (bhanml@comp.hkbu.edu.hk) and Jiangchao Yao (Sunarker@sjtu.edu.cn).} \quad
\textbf{Bo Han}$^{1 \dagger}$
\end{tabular}
\vspace{0mm} \\
$^{1}$TMLR Group, Department of Computer Science, Hong Kong Baptist University \\ 
$^{2}$Shanghai Innovation Institute \quad $^{3}$CMIC, Shanghai Jiao Tong University \\
\texttt{\{cszzzhang,csjnzhu,cszkzhou,csxuanli,xiaofeng\}@comp.hkbu.edu.hk} \\
\texttt{bhanml@comp.hkbu.edu.hk}, \texttt{\{g3ra1d,sjtuszzh,Sunarker\}@sjtu.edu.cn}
}

\iclrfinalcopy % Uncomment for camera-ready version, but NOT for submission.
\begin{document}

\maketitle

\begin{abstract}
While reinforcement learning with verifiable rewards (RLVR) is effective to improve the reasoning ability of large language models (LLMs), its reliance on human-annotated labels leads to the scaling up dilemma, especially for complex tasks. Recent self-rewarding methods investigate a label-free alternative to unlock the reasoning capabilities of LLMs, yet they frequently encounter the non-negligible training collapse issue, as the single-view supervision signal easily forms the self-consistent illusion, yielding the reward hacking. Inspired by the success of self-supervised learning, we propose \textit{Co-rewarding}, a novel self-supervised RL framework that improves training stability by seeking complementary supervision from another views. Specifically, we instantiate Co-rewarding in two ways: (1) \textit{Co-rewarding-I} is a data-side instantiation that derives reward signals from contrastive agreement across semantically analogous questions; and (2) \textit{Co-rewarding-II} is a model-side instantiation that maintains a slowly-updated reference teacher with pseudo labels to realize self-distillation. Intuitively, such instantiations introduce different levels of discrepancy to increase the difficulty of training collapse on trivial reasoning solutions. We also explore their orthogonally combined version to further boost the performance. Empirically, Co-rewarding exhibits stable training across various setups, and outperforms other self-rewarding baselines by $+3.31\%$ improvements on average on multiple mathematical reasoning benchmarks, especially by $+7.49\%$ on Llama-3.2-3B-Instruct. Notably, Co-rewarding reaches or even surpasses RLVR with ground-truth (GT) label in several cases, such as a Pass@$1$ of $94.01\%$ on GSM8K with Qwen3-8B-Base remarkably higher than GT. Our code is released at~\url{https://github.com/tmlr-group/Co-rewarding}.
\end{abstract}

\begin{figure}[h]
  \centering
  \vspace{-4mm}
  \includegraphics[width=0.42\textwidth]{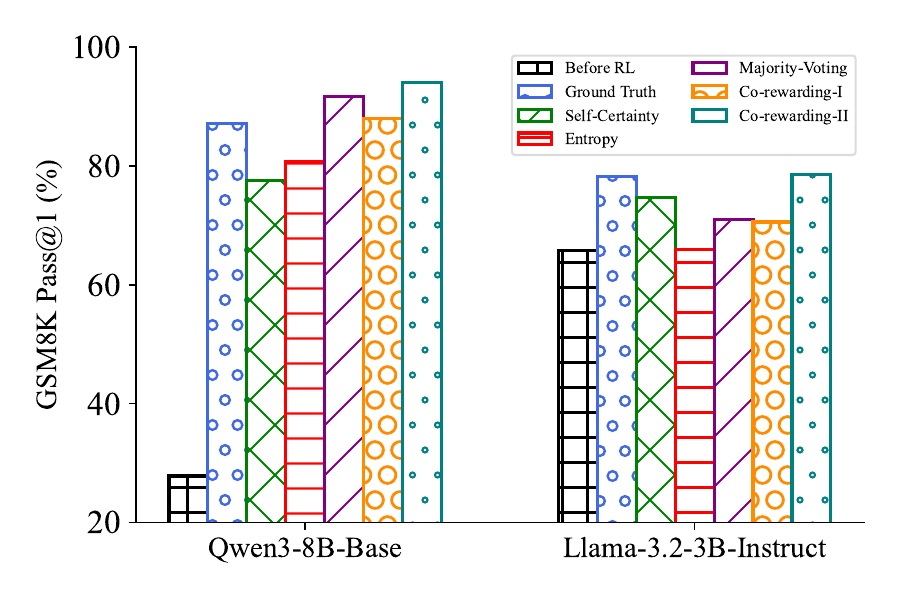}
  \includegraphics[width=0.27\textwidth]{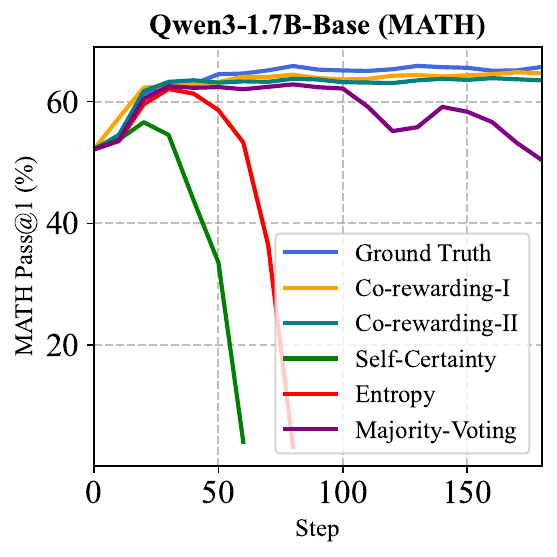}
  \includegraphics[width=0.27\textwidth]{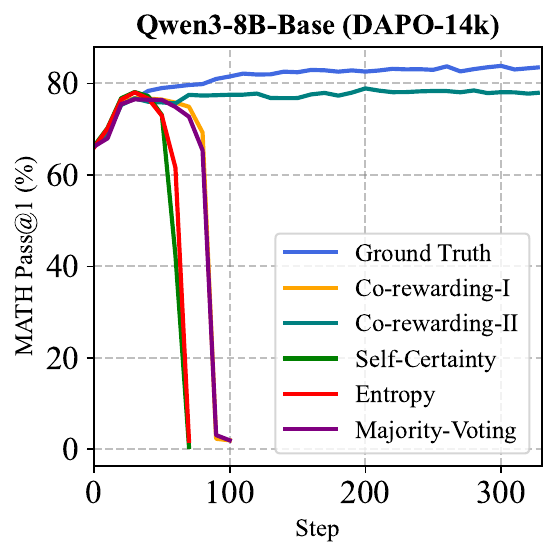}
  \vspace{-3mm}
  \caption{\textbf{Performance overview.} Reasoning comparison of Pass@1 value and validation curves. Our Co-rewarding achieves better and more stable (without collapse) training than other baselines.}
  \label{fig:comparison}
  \vspace{-2mm}
\end{figure}

\section{Introduction}
Large language models (LLMs)~\citep{achiam2023gpt,dubey2024llama} have demonstrated remarkable general-purpose capabilities in a wide range of linguistic tasks~\citep{li2026llm,chen2026eepo}. To further elicit their reasoning ability in complex scenarios, reinforcement learning with verifiable rewards (RLVR)~\citep{shao2024deepseekmath,yu2025dapo} is developed for post-training with externally verifiable signals like program execution results~\citep{deepcoder2025} or mathematical equivalence~\citep{shao2024deepseekmath}. Despite the impressive improvement, the reliance on high-quality ground-truth (GT) labels of RLVR remains as a major bottleneck~\citep{ouyang2022training,bai2022training} in the spirit of the scaling law, which subsequently motivates the emerging exploration of self-rewarding methods with unlabeled data~\citep{zhao2025learning,zuo2025ttrl,zhang2025consistent}.

One prominent line of such label-free methods leverages the internal signals (e.g., entropy~\citep{zhang2025right, prabhudesai2025maximizing} and self-certainty~\citep{zhao2025learning}) to strengthen the confidence of the model in reasoning. Another critical line seeks the answer-level consensus~\citep{zuo2025ttrl,shafayat2025can} to construct pseudo labels as reward basis. while effective initially, these self-rewarding approaches frequently exhibit non-negligible training collapse~\citep{zhang2025no} (indicated as right of Figure~\ref{fig:comparison}), which limits the scalability of such label-free training manners. 

The collapse phenomenon stems from reward hacking~\citep{laidlawcorrelated} under self-consistent illusion: the reward signal is internally produced by the policy model from a single-view data perspective, which is easily trapped by trivial solutions along with training (see Figure~\ref{Fig:Case_study}). Specifically, for entropy- or certainty-based rewards, the policy model may concentrate probability mass on a small set of tokens and produce repetitive strings that minimize entropy or maximize self-certainty~\citep{zhang2025no}. And for consensus-based rewards, the policy model can converge to a consistent yet incorrect answer that attains high consensus across rollouts~\citep{shafayat2025can}. 
Overall, the policy model continually reduces uncertainty without sustained gains in correctness, inflating the reward but eroding exploration and diversity. It ultimately collapses when a persistent hacking strategy emerges.

\begin{figure}[t]
  \centering
  \vspace{-2mm}
  % \fbox{\rule{0pt}{1.5in} \rule{0.9\linewidth}{0pt}}  % 2in tall, 90% width
  \includegraphics[width=0.98\textwidth]{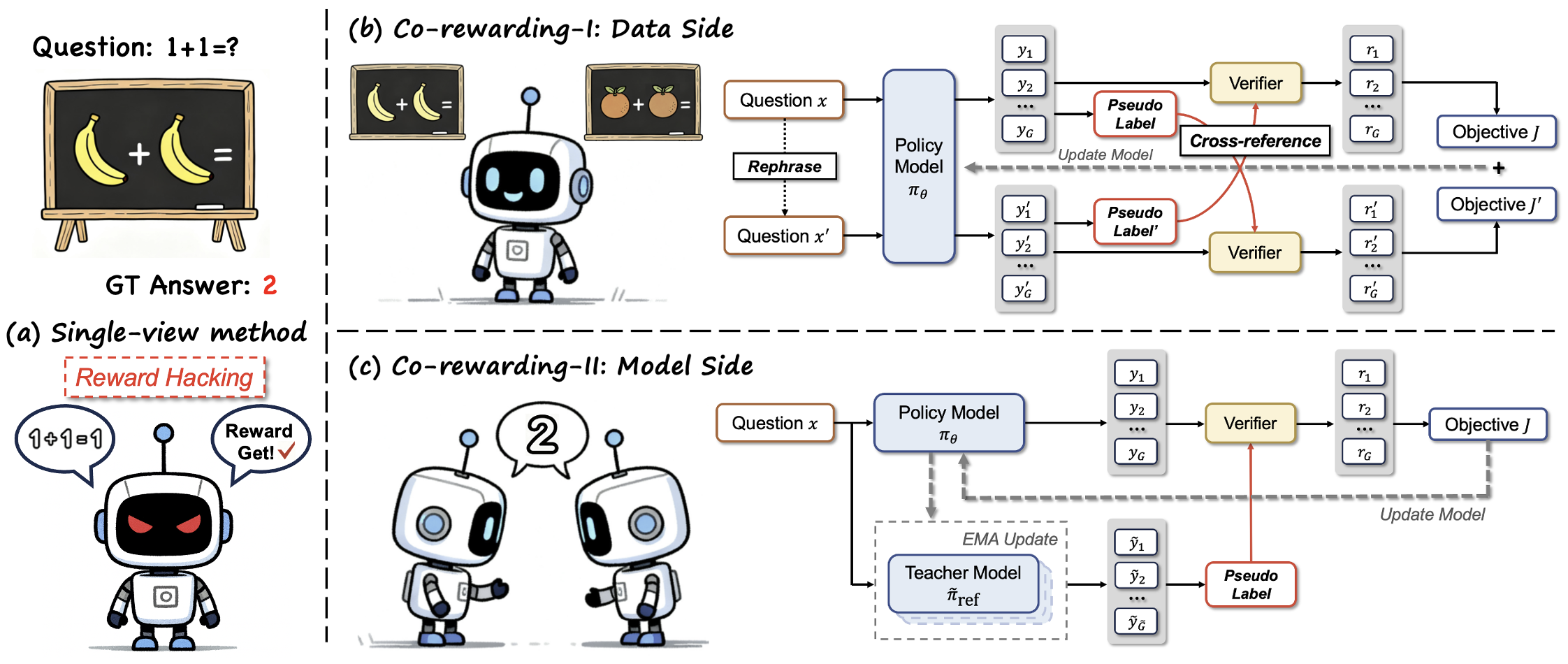}
  \vspace{-2mm}
  \caption{\textbf{Illustration of \textit{Co-rewarding} framework}: Unlike single-view methods that rely only on internal reward signal on original question (a), Co-rewarding introduces complementary supervision. On the data side (b), paraphrased questions yield pseudo-labels for cross-reference. On the model side (c), teacher model isolated from current policy provides stabilized pseudo-labels for updates.}
  \label{Fig:Co-rewarding_V1}
  \vspace{-6mm}
\end{figure}

To this end, we introduce \textit{Co-rewarding}, a self-supervised RL framework that seeks complementary supervision from another views, inspired by self-supervised learning~\citep{chen2020SimCLR, grill2020BYOL, caron2021DINO}. Conceptually, one fundamental characteristic of self-rewarding methods lies on that supervision intertwined with current policy on single-view outputs, for which we propose to seek reasoning invariance across different views (see Figure~\ref{Fig:Co-rewarding_V1}). Specifically, we investigate two initiations of Co-rewarding: (1) \textit{Co-rewarding-I}: a data-side initiation that constructs rewards via contrastive agreement across semantically analogous questions, each providing pseudo labels for the other; and (2) \textit{Co-rewarding-II}: a model-side initiation that introduces an extra teacher with dynamically updated policy and provides stable pseudo-labels insulated from current online policy. Additionally, we also explore the combined instantiation, \textit{Co-rewarding-III}, which integrates data-side cross-supervision with model-side self-distillation to further boost the performance.

By introducing cross-view supervision on data and decoupling the reward signal from the current policy, Co-rewarding effectively mitigates training collapse and yields stable self-supervised RL training. Extensive experiments across multiple datasets validate the stability and superiority of Co-rewarding, compared to several recent baselines across several LLM families including Qwen3/2.5 and Llama. Notably, both Co-rewarding-I and -II reach or exceed training with ground-truth labels in several settings, such as achieving up to $94.01\%$ Pass@1 on GSM8K. Our main contributions are
\begin{itemize}[leftmargin=0.3cm]
    \item We introduce a new perspective, from self-supervised learning, to elicit reasoning capability via another views of supervision, which prevents the model from training collapse (Section~\ref{sec:method_philosophy}). 
    \item We propose Co-rewarding, a novel self-supervised RL framework that is initiated by the data and model sides to construct self-generate rewards to promote stably reasoning elicitation (Section~\ref{sec:method_init}).
    \item We empirically demonstrate the general effectiveness of Co-rewarding to achieve superior reasoning performance on LLMs, and also present various ablation studies and further analyses (Section~\ref{sec:exp}).
\end{itemize}

% \vspace{-2mm}
\section{Preliminary}
% \vspace{-1mm}
\textbf{Problem Setups.} 
Given a LLM $\pi_\theta$ parameterized by $\theta$ and a dataset $\mathcal{D}$ of question–answer pairs $(x,a)$, the model generates a response $y \sim \pi_\theta(\cdot \mid x)$ auto\-regressively.
Let $y=(y_1,\dots,y_n)$, where each token is sampled as $y_t \sim \pi_\theta(\cdot \mid x, y_{<t})$ given the generated prefix $y_{<t}$. We consider the LLM outputs a step\-by\-step reasoning trace and a final answer. A verifiable reward function $r(a,y)$ compares the extracted answer $\mathrm{ans}(y)$ with the ground truth $a$ as follows:
\begin{equation}
    r(a, y)=\left \{
    \begin{aligned}
    &\text{1} &&\text{If}~\mathrm{ans}(y)~\text{is correct with answer}~a,\\
    &\text{0} &&\text{If}~\mathrm{ans}(y)~\text{is incorrect with answer}~a.
    \end{aligned}
    \right.
\end{equation}
Then, the general objective of training LLM for reasoning via RLVR~\citep{shao2024deepseekmath,yu2025dapo} can be formulated with the policy model $\pi_\theta$ as follows:
\begin{equation}
\label{Eq:RL_obj}
    \max_{\pi_\theta} \mathbb{E}_{(x,a)\in \mathcal{D},~y\sim\pi_\theta(x)}[r(a, y)-\beta\cdot\mathrm{KL}[\pi_\theta(y|x)||\pi_\textrm{ref}(y|x)]],
\end{equation}
where $\pi_\textrm{ref}$ is an initial reference policy, and $\beta$ is a coefficient controlling the KL divergence to prevent excessive deviation from the reference model. Intuitively, the training target is to maximize the reward in passing specific reasoning questions while maintaining the general capability of LLM.

\par
\textbf{Group Relative Policy Optimization (GRPO).}
In practice, we adopt GRPO~\citep{shao2024deepseekmath}, a widely used and representative optimization method for objective Eq.~(\ref{Eq:RL_obj}) that estimates the advantage by normalizing the reward across multiple sampled outputs for the same question. Specifically, for a given question $x$, GRPO samples $G$ outputs from the old policy $\pi_\textrm{old}$ as $\{y_i\}_{i=1}^{G} \sim \pi_{\text{old}}(\cdot | x)$. It then computes a reward for each output $y_i$ via a deterministic reward function, forming a group of rewards $\{r(a, y_i)\}_{i=1}^{G}$ to estimate the advantage $\hat{A}_{i}$ as follows:
\begin{equation}
\hat{A}_{i}
% = \bar{r}(a, y_i) 
= \frac{r(a,y_i) - \text{mean}(\{r(a,y_i)\}_{i=1}^{G})}{\text{std}(\{r(a,y_i)\}_{i=1}^{G})}.
\label{eqn: GRPO reward}
\end{equation}
Then, the target policy is optimized by maximizing the advantage while ensuring the policy model remains close to the reference policy: 
\begin{align}
\begin{split}
\label{eq:grpo}
    \mathcal{J}_\textrm{GRPO}&(\theta) = \mathbb{E}_{(x,a)\in \mathcal{D},\{{y_i}\}_{i=1}^G\sim \pi_{\theta_\textrm{old}}(\cdot|x)}\\
    &\underbrace{\frac{1}{G}\sum_{i=1}^G \frac{1}{|y_i|}\sum_{t=1}^{|y_i|}\bigg( \min\bigg[ c_{i,t}(\theta)\hat{A}_{i,t}, \textrm{clip}(c_{i,t}(\theta),1-\epsilon,1+\epsilon)\hat{A}_{i,t} \bigg]-\beta\mathbb{D}_{\textrm{KL}}(\pi_\theta||\pi_\textrm{ref}) \bigg)}_{\mathcal{R}_\theta(\hat{A})},
\end{split}
\end{align}
where
\begin{equation}
    c_{i,t}(\theta) = \frac{\pi_\theta(y_{i,t}|x,y_{i,<t})}{\pi_{\theta_\textrm{old}}(y_{i,t}|x,y_{i,<t})},\; \mathbb{D}_{\textrm{KL}}(\pi_\theta||\pi_\textrm{ref}) = \frac{\pi_\theta(y_{i,t}|x,y_{i,<t})}{\pi_\textrm{ref}(y_{i,t}|x,y_{i,<t})}-\log \frac{\pi_\textrm{ref}(y_{i,t}|x,y_{i,<t})}{\pi_\theta(y_{i,t}|x,y_{i,<t})}-1.
\end{equation}
Note that the $\textrm{clip}(\cdot,1-\epsilon,1+\epsilon)$ in Eq.~(\ref{eq:grpo}) is used to ensure that updates do not deviate excessively from the old policy by bounding the policy ratio between $1-\epsilon$ and $1+\epsilon$ in a risk function $\mathcal{R}(\hat{A})$. 
We also provide a comprehensive discussion on additional training variants for RLVR, such as DAPO~\citep{yu2025dapo} and Dr. GRPO~\citep{liu2025understanding}, which we leave in Appendix~\ref{appe:related_work} due to space limits.

\section{Co-rewarding}
\label{sec:method}

% Given the high cost of human annotation and the growing complexity of reasoning tasks~\citep{song2024mind, ouyang2022training}, an emerging and promising direction for scalable post-training of LLMs is to enhance reasoning capabilities without relying on ground-truth labels~\citep{shafayat2025can, zhao2025learning}, which relies solely on unlabeled data and aligns with the vision of self-evloving.

In the following, we present Co-rewarding in detail, a novel self-supervised RL framework for LLM to elicit the latent reasoning capability through the intuition of seeking complementary supervision.

\subsection{Conceptual Philosophy: invariance beyond the single-view}
\label{sec:method_philosophy}

At the core of self-rewarding methods lies a fundamental tension: the model derives supervisory signals from its own outputs, inevitably intertwining supervision with policy and risks collapse. 
True reasoning competence, however, cannot be reduced to the mere correctness of isolated answers. It should instead reflect invariance that extends beyond the single-view output for consistency. 
This calls for training signals that remain valid across different data views or persist throughout the temporal evolution of the model, providing a more reliable basis on which self-supervised RL can rely. In this aspect, stability arises from invariance that prevents reasoning against superficial variations in data and guides the model towards increasingly valid reasoning trajectories throughout training.

\par
This philosophy yields our Co-rewarding framework, whose core idea is to ground self-supervised RL in invariance rather than the suspicious single-view feedback.We instantiate it in two orthogonal ways and one combined version: by enforcing analogy-invariance on the data side (Co-rewarding-I), by disentangling supervision through temporal invariance on the model side (Co-rewarding-II), and by integrating both mechanisms in a unified instantiation (Co-rewarding-III).

\subsection{Two Initiations of Co-rewarding Framework}
\label{sec:method_init}

\textbf{Co-rewarding-I: \textit{on the Data Side}.}
Inspired by contrastive learning, such as SimCLR~\citep{chen2020SimCLR} and InfoNCE~\citep{oord2018infonce}, where two views of the same data are encouraged to have similar representations, we hypothesize an analogy-invariance inductive property of LLMs in eliciting reasoning capacity: questions that share the same mathematical essence but differ in surface form (e.g., via paraphrasing, background substitution, or reformatting) should elicit the comparably valid and similar reasoning results. This forms the foundation for a self-referential training signal: contrastive agreement among different question variants can serve as an optimization proxy. Co-rewarding-I defines contrastive agreement as a principle that aligns model reasoning outputs, treating consistent inter-view agreement as a signal for valid inference. This complements single-view self-rewarding strategies by introducing a form of collective validity verification with broader input consideration.

% While regularizing the latent space like traditional contrastive learning is hard in RL framework, 

\par
Building upon the discussed contrastive agreement, we initiate our \textit{Co-rewarding-I} as illustrated in Figure~\ref{Fig:Co-rewarding_V1}. Formally, its learning objective can be formulated as follows based on GRPO:
\begin{align}
\begin{split}
\label{eqn:co-reward_obj}
    \mathcal{J}_\textrm{Co-rewarding-I}(\theta) 
     \; = \;\underbrace{\mathbb{E}_{\textcolor{blue}{x\in \mathcal{D}},\{{\textcolor{blue}{y_i}}\}_{i=1}^G\sim \pi_{\theta_\textrm{old}}(\textcolor{blue}{\cdot|x})}\mathcal{R}_\theta(\hat{A})}_{\mathcal{J}_\textrm{\textcolor{blue}{original}}(\theta)}\;+\; \underbrace{\mathbb{E}_{\textcolor{brown}{x'\in \mathcal{D'}},\{\textcolor{brown}{{y_i}'}\}_{i=1}^G\sim \pi_{\theta_\textrm{old}}(\textcolor{brown}{\cdot|x'})}\mathcal{R}_\theta(\hat{A}')}_{\mathcal{J}_\textrm{\textcolor{brown}{rephrased}}(\theta)},
\end{split}
\end{align}
where the relative advantages are estimated by the \textit{cross-refereed} supervision as follows:
\begin{equation}
\hat{A}_{i} = 
\frac{r(\textcolor{brown}{y'_\textrm{v}},\textcolor{blue}{y_i}) - \text{mean}(\{r(\textcolor{brown}{y'_\textrm{v}},\textcolor{blue}{y_i})\}_{i=1}^{G})}{\text{std}(\{r(\textcolor{brown}{y'_\textrm{v}},\textcolor{blue}{y_i})\}_{i=1}^{G})},\; 
\hat{A}'_{i} = 
\frac{r(\textcolor{blue}{y_\textrm{v}},\textcolor{brown}{y_i'}) - \text{mean}(\{r(\textcolor{blue}{y_\textrm{v}},\textcolor{brown}{y_i'})\}_{i=1}^{G})}{\text{std}(\{r(\textcolor{blue}{y_\textrm{v}},\textcolor{brown}{y_i'})\}_{i=1}^{G})}.
\label{Eq:CorewardingV1_adv}
\end{equation}
Specifically, given a set of original questions, we utilize the rephrased version that keeps the semantical equivalence for the model to respond, and then collect the self-generated pseudo-labels based on the majority voting mechanism~\citep{shafayat2025can} as follows to supervise learning on the counterparts,
\begin{equation}
\label{eqn:co-reward_vote}
\textcolor{blue}{y_\textrm{v}}\leftarrow{\arg\max_{y*}\sum_{i=1}^G \mathrm{1}[\mathrm{ans}(\textcolor{blue}{y_i})=\mathrm{ans}(y*)]},\quad \textcolor{brown}{y_\textrm{v}'}\leftarrow{\arg\max_{y*}\sum_{i=1}^G \mathrm{1}[\mathrm{ans}(\textcolor{brown}{y_i'})=\mathrm{ans}(y*)]}.
\end{equation}
The overall pipeline can be viewed as a dual-path structure with cross-reference in the reward shaping process, it may also be compatible with other self-generated feedbacks~\citep{wang2022self} on the output-side information due to the generality of the core idea. While in the current version, we choose the majority voting mechanism in the implementation for the empirical effectiveness and simplicity.

\par
We summarize the pseudo code of Co-rewarding-I in Algorithm~\ref{Alg:Co-rewarding_V1}.
Our contrastive objective operates on self-generated reasoning answers, encouraging the model to align its reasoning results to different questions that share the similar semantic intent. Formally, for each input question, the signal of Co-rewarding-I increases when the model’s output is consistent with the majority answer obtained from its analogical counterparts, and decreases when it diverges. This contrastive agreement promotes semantic invariance, implicitly increasing the difficulty of reaching trivial solutions to obtain the reward (e.g., achieving the arbitrary answers but consistent on single input) by involving data-side analogy. We leave a more intuitive case study in the Appendix~\ref{app:rephrased_problem} to present the rephrased questions.

\textbf{Co-rewarding-II: \textit{on the Model Side}.}
On the data side, our Co-rewarding-I provides complementary supervision by involving question analogy, while its pseudo-labels are still generated by the current online policy and may depend on rephrasing quality; consequently, supervision remains partially entangled with the policy. Inspired by self- or weakly supervised methods like the representative BYOL~\citep{grill2020BYOL}, DINO~\citep{caron2021DINO}, and Co-teaching~\citep{han2018coteaching}, which share the common intuition of introducing an auxiliary network to provide supervision beyond the current model, we initiate \textit{Co-rewarding-II} from another view of complementary supervision: a model-side strategy that sources pseudo-labels from a teacher reference, which disentangle the self-supervision reward from the online policy. To avoid the heavy cost of adding and maintaining another LLM in training, Co-rewarding-II reuses the GRPO reference model as the teacher to generate the rollouts and produce pseudo-labels. In particular, the teacher is dynamically updated as an exponential moving average (EMA) of the student policy to ensure pseudo-label quality improving as the policy improves. 

\par
Intuitively, we illustrate \textit{Co-rewarding-II} in Figure~\ref{Fig:Co-rewarding_V1}. Its learning objective can be formulated as:
\begin{equation}
\label{Eq:Co-rewarding_obj}
    \mathcal{J}_\textrm{Co-rewarding-II}^{(k)}(\theta)\;=\;\mathbb{E}_{x\in\mathcal{D},\underbrace{\{y_i\}_{i=1}^G\sim \textcolor{blue}{\pi_{\theta_\textrm{old}}^{(k)}}(\cdot|x),}_{\textcolor{blue}{\text{policy student rollout}}}\underbrace{\{\tilde{y}_j\}_{j=1}^{\tilde{G}}\sim \textcolor{brown}{\tilde{\pi}_\textrm{ref}^{(k)}}(\cdot|x)}_{\textcolor{brown}{\text{reference teacher rollout}}}} \mathcal{R}_\theta(\hat{A}^{(k)}),
\end{equation}
where $\{\textcolor{blue}{y_i}\}_{i=1}^G$ are policy rollouts and $\{\textcolor{brown}{\tilde{y}_j}\}_{j=1}^{\tilde{G}}$ are reference teacher rollouts at the $k$-th training step, and the estimated advantage $\mathcal{R}(\hat{A}^{(k)})$ is computed as follows:
\begin{align}
    \hat{A}_i^{(k)} = \frac{r(\textcolor{brown}{\tilde{y}_\textrm{v}^{(k)}}, \textcolor{blue}{y_i}) - \textrm{mean} (\{r(\textcolor{brown}{\tilde{y}_\textrm{v}^{(k)}}, \textcolor{blue}{y_i})\}_{i=1}^G)}{\textrm{std}(\{r(\textcolor{brown}{\tilde{y}_\textrm{v}^{(k)}}, \textcolor{blue}{y_i})\}_{i=1}^G)}, \; \textcolor{brown}{\tilde{y}_\textrm{v}^{(k)}} = \arg \max_{y*} \sum_{j=1}^{\tilde{G}} \mathbf{1}[\textrm{ans}({\textcolor{brown}{\tilde{y}_j}})=\textrm{ans}(y*)], \label{Eq:CorewardingV2_label}
\end{align}
where the pseudo label $\tilde{y}_\textrm{v}^{(k)}$ is obtained via majority voting from reference rollouts, and the reference model is updated via an EMA with the policy to play a role of a slowly updated teacher:
\begin{equation}
\label{Eq:CorewardingV2_updateTeacher}
    \textcolor{brown}{\tilde{\pi}_{\textrm{ref}}^{(k)}} \leftarrow \alpha^{(k)} \cdot \textcolor{brown}{\tilde{\pi}_{\textrm{ref}}^{(k-1)}} + (1-\alpha^{(k)})\cdot \textcolor{blue}{\pi_{\theta_{\textrm{old}}}^{(k)}},\; \alpha^{(k)} = 1 - \frac{(\alpha_\textrm{end} - \alpha_\textrm{start})}{2}\left(1+ \cos \left(\frac{\pi k}{K}\right)\right)
\end{equation}
where $\alpha^{(k)}\in (0,1)$ is the EMA weight, updated according to a cosine annealing schedule from $\alpha_\textrm{start}$ to $\alpha_\textrm{end}$, such that the teacher is updated rapidly at the beginning and progressively more slowly, thereby evolving smoothly and remaining temporally decoupled from the current online policy.

\par
We summarize the pseudo code of Co-rewarding-II in Algorithm~\ref{alg:corewarding_v2}. This design can be interpreted as a kind of self-distillation, in which a slowly updated teacher supervises a faster-moving student. Such a paradigm breaks the single-step on-policy feedback loop inherent in existing self-rewarding methods~\citep{zhao2025learning,prabhudesai2025maximizing,shafayat2025can}, raises the cost of exploiting trivial low-entropy shortcuts or spurious consensus, and offers a stable reward source without introducing an additional LLM or optimizer. In this way, it effectively overcomes reward hacking and prevents training collapse by implicitly seeking a temporal invariance for true reasoning.

\textbf{Co-rewarding-III: \textit{Data-side + Model-side}.}
Given that Co-rewarding-I and Co-rewarding-II provide two complementary perspectives for constructing stable self-supervised signals, a natural exploration is to integrate both data-side cross-supervision and model-side self-distillation into a unified instantiation. We introduce \textit{Co-rewarding-III}, which leverages analogy-invariance between each original question and its rephrased counterparts while producing pseudo-labels from the EMA-updated reference teacher. Specifically, the teacher generates rollouts for both original and rephrased questions, and the resulting pseudo-label from one side is used to supervise the other. This combination further boosts the resistance of the reward signal to hacking, promoting more stable training dynamics.

\par
Formally, its learning objective can be formulated as:
\begin{align}
\label{Eq:Co-rewarding-III_obj}
    &\mathcal{J}_\textrm{Co-rewarding-III}^{(k)}(\theta)\;=\;\mathbb{E}_{\textcolor{blue}{x\in\mathcal{D}},\underbrace{\{\textcolor{blue}{y_i}\}_{i=1}^G\sim \pi_{\theta_\textrm{old}}^{(k)}(\textcolor{blue}{\cdot|x}),}_{\textcolor{blue}{\text{policy student rollout from original question}}} \textcolor{brown}{x'\in\mathcal{D'}}, \underbrace{\{\textcolor{brown}{\tilde{y}_j'}\}_{j=1}^{\tilde{G}}\sim \tilde{\pi}_\textrm{ref}^{(k)}(\textcolor{brown}{\cdot|x'})}_{\textcolor{brown}{\text{reference teacher rollout from rephrased question}}}} \mathcal{R}_\theta(\hat{A}^{(k)}) \nonumber \\
    &\quad + \mathbb{E}_{\textcolor{brown}{x'\in\mathcal{D'}},\underbrace{\{\textcolor{brown}{y_i'}\}_{i=1}^G\sim \pi_{\theta_\textrm{old}}^{(k)}(\textcolor{brown}{\cdot|x'}),}_{\textcolor{brown}{\text{policy student rollout from rephrased question}}}\textcolor{blue}{x\in\mathcal{D}}, \underbrace{\{\textcolor{blue}{\tilde{y}_j}\}_{j=1}^{\tilde{G}}\sim \tilde{\pi}_\textrm{ref}^{(k)}(\textcolor{blue}{\cdot|x})}_{\textcolor{blue}{\text{reference teacher rollout from original question}}}} \mathcal{R}_\theta(\hat{A}'^{(k)}),
\end{align}
where the first term supervises the original question via pseudo labels generated from its rephrased counterpart, and the second term, symmetrically, supervises the rephrased question via pseudo labels generated from the original question. The estimated advantages $\mathcal{R}_\theta(\hat{A}^{(k)})$ and $\mathcal{R}_\theta(\hat{A}'^{(k)})$ are computed in the similar way as in Co-rewarding-I and Co-rewarding-II. The reference teacher is also updated via EMA, as Eq.~(\ref{Eq:CorewardingV2_updateTeacher}) in Co-rewarding-II. The other formulations and pseudo code of Co-rewarding-III are supplemented in Appendix~\ref{app:formulation_III} and Algorithm~\ref{alg:corewarding_v3}.

\par
\textbf{Remark 1.} Overall, the two instantiations of Co-rewarding embody our core idea from different perspectives: I leverages data-side analogy-invariance to provide cross supervision, while II employs model-side self-distillation to stabilize learning. Together, they reflect that stable self-supervised reasoning elicitation can emerge from both the diversity of data perspectives and the disentanglement of supervision signals. Co-rewarding-III further explores an orthogonally combined instantiation of these two sides. Moreover, Co-rewarding offers a flexible framework, in which key components, such as pseudo-labeling strategies, data rephrasing techniques, teacher model update rules, and policy optimization, can be seamlessly substituted with other advanced approaches~\citep{yu2025dapo}.

\begin{table}[t]
    \centering
    \caption{\textbf{Main Results (\%) of Co-rewarding and baselines trained on MATH.} Cell background colors indicate relative performance: darker colors denote better results within each model group. Additional results of Qwen2.5-3B/7B and Qwen3-1.7B-Base trained on MATH refer to Table~\ref{Tble:Appe_Exp_CorewardingV1}.}
    \label{Tble:Main_Exp_CorewardingV1}
    \vspace{-3mm}
    \resizebox{\textwidth}{!}{
    \begin{tabular}{l|cccccccc}
    \toprule[1.6pt]
        \textbf{Training Set: MATH} & \multicolumn{4}{c}{\textbf{Mathematics}} & \multicolumn{2}{c}{\textbf{Code}} & \textbf{Instruction} & \textbf{Multi-Task} \\ \cmidrule{1-9}
        \textbf{Methods} & \textbf{MATH500} & \textbf{GSM8K} & \textbf{AMC} & \textbf{AIME24} & \textbf{LiveCode} & \textbf{CRUX} & \textbf{IFEval} & \textbf{MMLU-Pro} \\ 
        % Metric & ~ & Pass@1 & Pass@1 & Accuracy & pass@1 & pass@1 \\ 
        \midrule
            \multicolumn{9}{c}{\textit{\textbf{Qwen3-8B-Base}}} \\
        \midrule
        Before RL & 72.4 & 27.82 & 20.93 & 3.75 & 23.41 & 54.75 & 50.89 & 52.92 \\ 
        - GT-Reward~\citep{shao2024deepseekmath} & 82.6 & 87.26 & 54.22 & 17.15 & 30.52 & 63.25 & 52.78 & 57.11 \\
        \midrule
        - Self-Certainty~\citep{zhao2025learning}   & \cellcolor{lightred2}{80.2} & \cellcolor{lightred1}{80.74} & \cellcolor{lightred3}{50.75} & \cellcolor{lightred5}{15.73} & \cellcolor{lightred1}{27.20} & \cellcolor{lightred5}{64.38} & \cellcolor{lightred1}{50.98} & \cellcolor{lightred1}{54.17} \\
        - Entropy~\citep{prabhudesai2025maximizing} & \cellcolor{lightred2}{80.2} & \cellcolor{lightred2}{87.19} & \cellcolor{lightred2}{49.54} & \cellcolor{lightred4}{15.63} & \cellcolor{lightred2}{29.38} & \cellcolor{lightred1}{62.00} & \cellcolor{lightred3}{51.81} & \cellcolor{lightred2}{54.86} \\
        - Majority-Voting~\citep{shafayat2025can}   & \cellcolor{lightred1}{79.8} & \cellcolor{lightred3}{89.76} & \cellcolor{lightred1}{49.09} & \cellcolor{lightred6}{15.83} & \cellcolor{lightred4}{30.52} & \cellcolor{lightred3}{63.38} & \cellcolor{lightred2}{51.80} & \cellcolor{lightred3}{56.93} \\
        - Co-rewarding-I (Ours)                     & \cellcolor{lightred5}{81.2} & \cellcolor{lightred6}{93.70} & \cellcolor{lightred4}{51.20} & \cellcolor{lightred3}{15.10} & \cellcolor{lightred6}{30.81} & \cellcolor{lightred6}{66.00} & \cellcolor{lightred5}{55.79} & \cellcolor{lightred6}{59.95} \\
        - Co-rewarding-II (Ours)                    & \cellcolor{lightred4}{80.8} & \cellcolor{lightred5}{92.42} & \cellcolor{lightred5}{53.46} & \cellcolor{lightred2}{14.48} & \cellcolor{lightred3}{30.23} & \cellcolor{lightred2}{62.83} & \cellcolor{lightred6}{60.70} & \cellcolor{lightred4}{57.50} \\
        - Co-rewarding-III (Ours)                   & \cellcolor{lightred6}{81.4} & \cellcolor{lightred4}{90.98} & \cellcolor{lightred6}{54.07} & \cellcolor{lightred1}{13.33} & \cellcolor{lightred5}{30.71} & \cellcolor{lightred4}{63.75} & \cellcolor{lightred4}{53.69} & \cellcolor{lightred5}{59.10} \\ 
        
        \specialrule{1pt}{0.4ex}{0.4ex}

        \multicolumn{9}{c}{\textit{\textbf{Qwen3-4B-Base}}} \\
        \midrule
        Before RL & 71.2 & 26.15 & 21.08 & 4.58 & 11.00 & 38.88 & 46.43 & 47.23 \\ 
        - GT-Reward~\citep{shao2024deepseekmath} & 78.6 & 89.76 & 51.20 & 15.00 & 26.07 & 55.38 & 47.80 & 53.96 \\
        \midrule
        - Self-Certainty~\citep{zhao2025learning}   & \cellcolor{lightorange1}{71.6} & \cellcolor{lightorange1}{71.79} & \cellcolor{lightorange1}{38.86} & \cellcolor{lightorange3}{11.67} & \cellcolor{lightorange1}{22.37} & \cellcolor{lightorange3}{57.00} & \cellcolor{lightorange1}{48.15} & \cellcolor{lightorange1}{48.93} \\
        - Entropy~\citep{prabhudesai2025maximizing} & \cellcolor{lightorange2}{77.0} & \cellcolor{lightorange2}{88.10} & \cellcolor{lightorange5}{47.44} & \cellcolor{lightorange2}{10.94} & \cellcolor{lightorange2}{25.59} & \cellcolor{lightorange1}{52.88} & \cellcolor{lightorange5}{50.44} & \cellcolor{lightorange2}{49.90} \\
        - Majority-Voting~\citep{shafayat2025can}   & \cellcolor{lightorange3}{77.4} & \cellcolor{lightorange4}{90.07} & \cellcolor{lightorange2}{45.33} & \cellcolor{lightorange1}{10.10} & \cellcolor{lightorange5}{26.54} & \cellcolor{lightorange6}{57.50} & \cellcolor{lightorange2}{48.78} & \cellcolor{lightorange6}{54.35} \\
        - Co-rewarding-I (Ours)                     & \cellcolor{lightorange6}{78.8} & \cellcolor{lightorange6}{91.28} & \cellcolor{lightorange4}{46.08} & \cellcolor{lightorange6}{13.85} & \cellcolor{lightorange6}{26.64} & \cellcolor{lightorange4}{56.50} & \cellcolor{lightorange4}{50.35} & \cellcolor{lightorange4}{53.26} \\
        - Co-rewarding-II (Ours)                    & \cellcolor{lightorange4}{78.0} & \cellcolor{lightorange3}{88.86} & \cellcolor{lightorange3}{45.93} & \cellcolor{lightorange4}{12.17} & \cellcolor{lightorange4}{26.25} & \cellcolor{lightorange2}{55.00} & \cellcolor{lightorange6}{51.30} & \cellcolor{lightorange5}{53.88} \\
        - Co-rewarding-III (Ours)                   & \cellcolor{lightorange5}{78.6} & \cellcolor{lightorange5}{90.75} & \cellcolor{lightorange6}{48.80} & \cellcolor{lightorange5}{12.71} & \cellcolor{lightorange3}{26.16} & \cellcolor{lightorange3}{56.00} & \cellcolor{lightorange3}{49.23} & \cellcolor{lightorange3}{53.08} \\
        
        \specialrule{1pt}{0.4ex}{0.4ex}

            \multicolumn{9}{c}{\textit{\textbf{Llama-3.2-3B-Instruct}}} \\
        \midrule
        Before RL & 39.2 & 65.73 & 10.54 & 3.75 & 9.86 & 25.37 & 57.32 & 31.14 \\ 
        - GT-Reward~\citep{shao2024deepseekmath} & 47.0 & 77.94 & 22.14 & 11.67 & 9.57 & 31.87 & 47.51 & 34.32 \\
        \midrule
        - Self-Certainty~\citep{zhao2025learning}   & \cellcolor{lightcyan1}{43.4} & \cellcolor{lightcyan2}{74.91} & \cellcolor{lightcyan1}{18.83} & \cellcolor{lightcyan2}{6.88} & \cellcolor{lightcyan1}{9.95} & \cellcolor{lightcyan2}{25.87} & \cellcolor{lightcyan5}{54.88} & \cellcolor{lightcyan2}{33.34} \\
        - Entropy~\citep{prabhudesai2025maximizing} & \cellcolor{lightcyan1}{43.4} & \cellcolor{lightcyan1}{66.19} & \cellcolor{lightcyan2}{20.18} & \cellcolor{lightcyan1}{6.56} & \cellcolor{lightcyan6}{11.66} & \cellcolor{lightcyan1}{24.62} & \cellcolor{lightcyan4}{54.70} & \cellcolor{lightcyan3}{33.52} \\
        - Majority-Voting~\citep{shafayat2025can}   & \cellcolor{lightcyan3}{46.8} & \cellcolor{lightcyan3}{78.77} & \cellcolor{lightcyan3}{20.48} & \cellcolor{lightcyan3}{9.27} & \cellcolor{lightcyan4}{11.00} & \cellcolor{lightcyan5}{31.25} & \cellcolor{lightcyan2}{47.96} & \cellcolor{lightcyan1}{33.18} \\
        - Co-rewarding-I (Ours)                     & \cellcolor{lightcyan5}{50.2} & \cellcolor{lightcyan5}{79.45} & \cellcolor{lightcyan5}{23.80} & \cellcolor{lightcyan4}{10.00} & \cellcolor{lightcyan5}{11.28} & \cellcolor{lightcyan3}{29.88} & \cellcolor{lightcyan3}{48.89} & \cellcolor{lightcyan5}{33.77} \\
        - Co-rewarding-II (Ours)                    & \cellcolor{lightcyan4}{49.8} & \cellcolor{lightcyan4}{79.30} & \cellcolor{lightcyan4}{22.59} & \cellcolor{lightcyan6}{10.73} & \cellcolor{lightcyan3}{10.80} & \cellcolor{lightcyan4}{30.63} & \cellcolor{lightcyan4}{49.90} & \cellcolor{lightcyan4}{33.61} \\
        - Co-rewarding-III (Ours)                   & \cellcolor{lightcyan6}{51.6} & \cellcolor{lightcyan6}{79.91} & \cellcolor{lightcyan6}{25.45} & \cellcolor{lightcyan5}{10.42} & \cellcolor{lightcyan2}{10.43} & \cellcolor{lightcyan6}{32.50} & \cellcolor{lightcyan1}{46.37} & \cellcolor{lightcyan6}{34.50} \\
        
        \bottomrule[1.6pt]
    \end{tabular}}
    \vspace{-7mm}
\end{table}

\vspace{-2mm}
\section{Experiments}
\label{sec:exp}
\vspace{-2mm}
\subsection{Setups}
\label{sec:exp_setup}

\textbf{Backbone Models and Baselines.} We employ a diverse set of LLMs from different families and scales in our experiments, including the Qwen2.5 series (Qwen2.5-3B/7B)~\citep{qwen2025qwen25technicalreport}, the Qwen3 series (Qwen3-1.7B/4B/8B-Base)~\citep{yang2025qwen3}, and the Llama3 series (Llama-3.2-3B-Instruct)~\citep{meta2024llama32}. Beyond the vanilla GRPO that utilized the GT label for rewarding, we compare our Co-rewarding against several recent state-of-the-art (SoTA) self-reward reasoning approaches, denoted as Self-Certainty~\citep{zhao2025learning}, Entropy~\citep{prabhudesai2025maximizing} and Majority Voting~\citep{shafayat2025can}. The details of all baselines are summarized in Appendix~\ref{app:details_baseline}.

\begin{table}[t!]
    \centering
    \caption{\textbf{Main Results (\%) of Co-rewarding and baselines trained on DAPO-14k}. Cell background colors indicate relative performance: darker colors denote better results within each model group. Additional Results of Qwen3-8B-Base and Qwen3-4B-Base trained on OpenRS refer to Table~\ref{Tble:Supplement_Exp_CorewardingV2_OpenRS}.
    }
    \vspace{-3mm}
    \label{Tble:Main_Exp_CorewardingV2}
    \resizebox{\textwidth}{!}{
    \begin{tabular}{l|cccccccc}
    \toprule[1.6pt]
        \textbf{Training Set: DAPO-14k} & \multicolumn{4}{c}{\textbf{Mathematics}} & \multicolumn{2}{c}{\textbf{Code}} & \textbf{Instruction} & \textbf{Multi-Task} \\ \cmidrule{1-9}
        \textbf{Methods} & \textbf{MATH500} & \textbf{GSM8K} & \textbf{AMC} & \textbf{AIME24} & \textbf{LiveCode} & \textbf{CRUX} & \textbf{IFEval} & \textbf{MMLU-Pro} \\ 
        % Metric & ~ & Pass@1 & Pass@1 & Accuracy & pass@1 & pass@1 \\ 

        % ## Qwen3-8B-Base
        \midrule
            \multicolumn{9}{c}{\textit{\textbf{Qwen3-8B-Base}}} \\
        \midrule
        Before RL & 72.4 & 27.82 & 20.93 & 3.75 & 23.41 & 54.75 & 50.89 & 52.92 \\ 
        - GT-Reward~\citep{shao2024deepseekmath} & 86.6 & 87.19 & 61.75 & 24.58 & 30.52 & 63.75 & 53.11 & 60.27 \\
        \midrule
        - Self-Certainty~\citep{zhao2025learning}   & \cellcolor{lightred6}{82.0} & \cellcolor{lightred1}{77.63} & \cellcolor{lightred2}{49.85} & \cellcolor{lightred3}{15.00} & \cellcolor{lightred1}{27.77} & \cellcolor{lightred1}{60.75} & \cellcolor{lightred2}{50.58} & \cellcolor{lightred1}{54.24} \\
        - Entropy~\citep{prabhudesai2025maximizing} & \cellcolor{lightred3}{79.4} & \cellcolor{lightred2}{80.82} & \cellcolor{lightred1}{45.48} & \cellcolor{lightred3}{15.00} & \cellcolor{lightred3}{30.14} & \cellcolor{lightred3}{62.00} & \cellcolor{lightred4}{51.56} & \cellcolor{lightred2}{54.57} \\
        - Majority-Voting~\citep{shafayat2025can}   & \cellcolor{lightred2}{78.6} & \cellcolor{lightred4}{91.66} & \cellcolor{lightred3}{50.00} & \cellcolor{lightred1}{11.25} & \cellcolor{lightred4}{30.33} & \cellcolor{lightred2}{61.62} & \cellcolor{lightred3}{51.54} & \cellcolor{lightred4}{55.65} \\
        - Co-rewarding-I (Ours)                     & \cellcolor{lightred1}{78.4} & \cellcolor{lightred3}{88.02} & \cellcolor{lightred4}{51.20} & \cellcolor{lightred2}{11.88} & \cellcolor{lightred2}{29.38} & \cellcolor{lightred4}{62.50} & \cellcolor{lightred1}{50.17} & \cellcolor{lightred3}{55.39} \\
        - Co-rewarding-II (Ours)                    & \cellcolor{lightred4}{80.6} & \cellcolor{lightred6}{94.01} & \cellcolor{lightred6}{54.37} & \cellcolor{lightred5}{16.35} & \cellcolor{lightred5}{31.66} & \cellcolor{lightred6}{67.12} & \cellcolor{lightred5}{53.31} & \cellcolor{lightred5}{59.83} \\
        - Co-rewarding-III (Ours)                   & \cellcolor{lightred5}{81.6} & \cellcolor{lightred5}{92.27} & \cellcolor{lightred5}{53.77} & \cellcolor{lightred6}{17.71} & \cellcolor{lightred6}{32.70} & \cellcolor{lightred5}{66.75} & \cellcolor{lightred6}{55.85} & \cellcolor{lightred6}{60.02} \\
        
        % ## Qwen3-4B-Base
        \specialrule{1pt}{0.4ex}{0.4ex}
            \multicolumn{9}{c}{\textit{\textbf{Qwen3-4B-Base}}} \\
        \midrule
        Before RL & 71.2 & 26.15 & 21.08 & 4.58 & 11.00 & 38.88 & 46.43 & 47.23 \\ 
        - GT-Reward~\citep{shao2024deepseekmath} & 83.6 & 85.14 & 52.86 & 20.63 & 18.58 & 56.88 & 47.70 & 55.35 \\
        \midrule
        - Self-Certainty~\citep{zhao2025learning}   & \cellcolor{lightorange1}{68.4} & \cellcolor{lightorange1}{44.81} & \cellcolor{lightorange1}{35.39} & \cellcolor{lightorange1}{8.85} & \cellcolor{lightorange1}{25.88} & \cellcolor{lightorange1}{50.12} & \cellcolor{lightorange1}{45.58} & \cellcolor{lightorange1}{48.84} \\
        - Entropy~\citep{prabhudesai2025maximizing} & \cellcolor{lightorange4}{76.6} & \cellcolor{lightorange4}{82.79} & \cellcolor{lightorange3}{43.37} & \cellcolor{lightorange4}{12.81} & \cellcolor{lightorange4}{26.35} & \cellcolor{lightorange2}{50.75} & \cellcolor{lightorange3}{48.20} & \cellcolor{lightorange2}{50.22} \\
        - Majority-Voting~\citep{shafayat2025can}   & \cellcolor{lightorange2}{73.4} & \cellcolor{lightorange2}{64.06} & \cellcolor{lightorange2}{40.81} & \cellcolor{lightorange2}{9.17} & \cellcolor{lightorange2}{26.16} & \cellcolor{lightorange3}{53.00} & \cellcolor{lightorange4}{48.91} & \cellcolor{lightorange3}{51.06} \\
        - Co-rewarding-I (Ours)                     & \cellcolor{lightorange3}{73.8} & \cellcolor{lightorange3}{75.89} & \cellcolor{lightorange4}{43.83} & \cellcolor{lightorange3}{10.63} & \cellcolor{lightorange3}{26.25} & \cellcolor{lightorange1}{50.12} & \cellcolor{lightorange2}{46.84} & \cellcolor{lightorange4}{51.51} \\
        - Co-rewarding-II (Ours)                    & \cellcolor{lightorange5}{77.8} & \cellcolor{lightorange6}{91.89} & \cellcolor{lightorange5}{48.49} & \cellcolor{lightorange5}{14.27} & \cellcolor{lightorange5}{26.64} & \cellcolor{lightorange6}{54.87} & \cellcolor{lightorange5}{48.90} & \cellcolor{lightorange5}{52.83} \\
        - Co-rewarding-III (Ours)                   & \cellcolor{lightorange6}{79.2} & \cellcolor{lightorange5}{90.45} & \cellcolor{lightorange6}{48.95} & \cellcolor{lightorange6}{15.10} & \cellcolor{lightorange6}{27.58} & \cellcolor{lightorange6}{54.87} & \cellcolor{lightorange6}{50.30} & \cellcolor{lightorange6}{54.79} \\

        % ## Llama-3.2-3B-Instruct
        \specialrule{1pt}{0.4ex}{0.4ex}
            \multicolumn{9}{c}{\textit{\textbf{Llama-3.2-3B-Instruct}}} \\
        \midrule
        Before RL & 39.2 & 65.73 & 10.54 & 3.75 & 9.86 & 25.37 & 57.32 & 31.14 \\ 
        - GT-Reward~\citep{shao2024deepseekmath} & 49.4 & 78.17 & 25.90 & 9.17 & 10.33 & 31.37 & 53.10 & 33.83 \\
        \midrule
        - Self-Certainty~\citep{zhao2025learning}   & \cellcolor{lightcyan1}{42.4} & \cellcolor{lightcyan4}{74.71} & \cellcolor{lightcyan1}{17.32} & \cellcolor{lightcyan1}{4.79} & \cellcolor{lightcyan6}{11.18} & \cellcolor{lightcyan3}{28.38} & \cellcolor{lightcyan5}{54.50} & \cellcolor{lightcyan4}{33.51} \\
        - Entropy~\citep{prabhudesai2025maximizing} & \cellcolor{lightcyan3}{44.0} & \cellcolor{lightcyan1}{65.85} & \cellcolor{lightcyan1}{17.32} & \cellcolor{lightcyan2}{6.56} & \cellcolor{lightcyan3}{9.95} & \cellcolor{lightcyan1}{25.00} & \cellcolor{lightcyan6}{55.78} & \cellcolor{lightcyan1}{31.95} \\
        - Majority-Voting~\citep{shafayat2025can}   & \cellcolor{lightcyan2}{42.8} & \cellcolor{lightcyan3}{70.96} & \cellcolor{lightcyan2}{17.62} & \cellcolor{lightcyan6}{8.74} & \cellcolor{lightcyan4}{10.14} & \cellcolor{lightcyan4}{29.50} & \cellcolor{lightcyan4}{54.07} & \cellcolor{lightcyan3}{32.95} \\
        - Co-rewarding-I (Ours)                     & \cellcolor{lightcyan4}{46.0} & \cellcolor{lightcyan2}{70.58} & \cellcolor{lightcyan5}{20.93} & \cellcolor{lightcyan3}{7.08} & \cellcolor{lightcyan1}{9.57} & \cellcolor{lightcyan2}{27.25} & \cellcolor{lightcyan3}{53.04} & \cellcolor{lightcyan2}{32.61} \\
        - Co-rewarding-II (Ours)                    & \cellcolor{lightcyan6}{49.8} & \cellcolor{lightcyan6}{78.62} & \cellcolor{lightcyan4}{19.73} & \cellcolor{lightcyan4}{8.02} & \cellcolor{lightcyan5}{10.43} & \cellcolor{lightcyan6}{32.25} & \cellcolor{lightcyan2}{51.92} & \cellcolor{lightcyan6}{34.46} \\
        - Co-rewarding-III (Ours)                   & \cellcolor{lightcyan5}{48.6} & \cellcolor{lightcyan5}{76.95} & \cellcolor{lightcyan6}{21.84} & \cellcolor{lightcyan5}{8.13} & \cellcolor{lightcyan2}{9.86} & \cellcolor{lightcyan5}{30.50} & \cellcolor{lightcyan1}{49.92} & \cellcolor{lightcyan5}{34.01} \\
        \bottomrule[1.6pt]
    \end{tabular}}
    \vspace{-4mm}
\end{table}

\begin{figure}[t!]
  \centering
  % 左列（约4）——方图
  \begin{minipage}[t]{0.435\textwidth}
    \centering
    \includegraphics[width=0.49\linewidth]{fig/val_curve/val_curve_qwen3-1.7b_math.pdf}
    \includegraphics[width=0.49\linewidth]{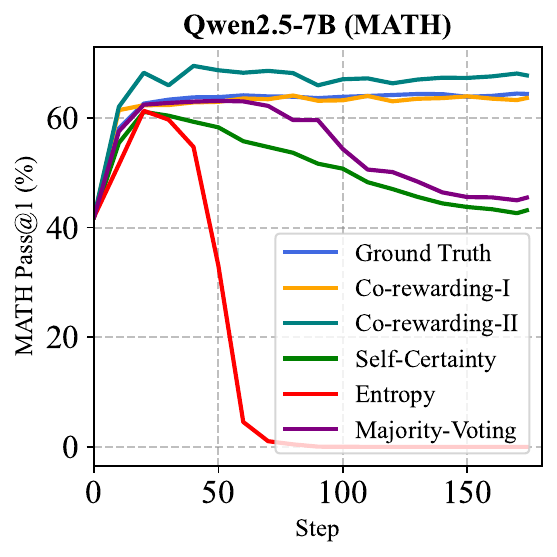}\\
    \includegraphics[width=0.49\linewidth]{fig/val_curve/val_curve_qwen3-8b.pdf}
    \includegraphics[width=0.49\linewidth]{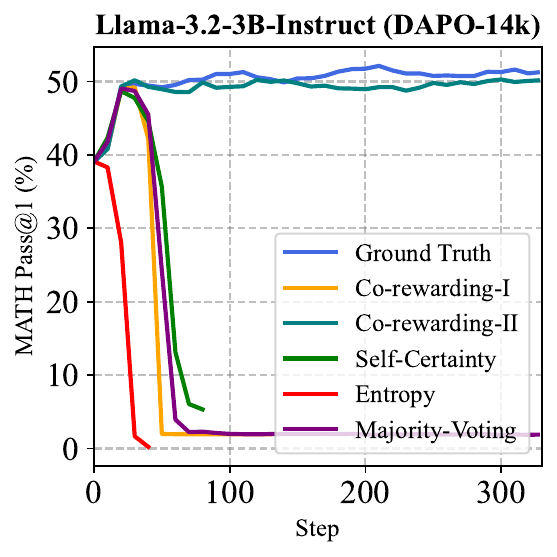}
    \vspace{-6.5mm}
    \caption{\textbf{Performance curves comparison} on validation set. \textit{Top:} Qwen3-1.7B-Base and Qwen2.5-7B trained on the MATH set. \textit{Bottom:} Qwen3-8B-Base and Llama-3.2-3B-Instruct trained on the DAPO-14k set.}
    \label{fig:curve_validation}
  \end{minipage}\hfill
  % 右列（约5）——5:4 图
  \begin{minipage}[t]{0.545\textwidth}
    \centering
    \includegraphics[width=0.49\linewidth]{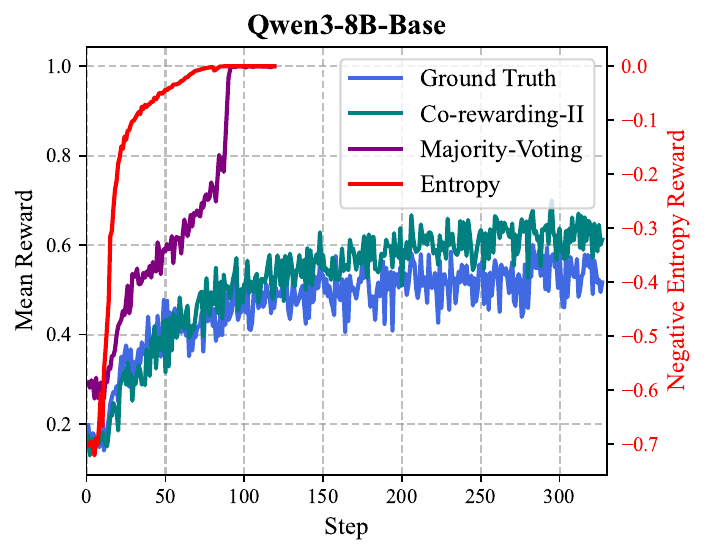}
    \includegraphics[width=0.49\linewidth]{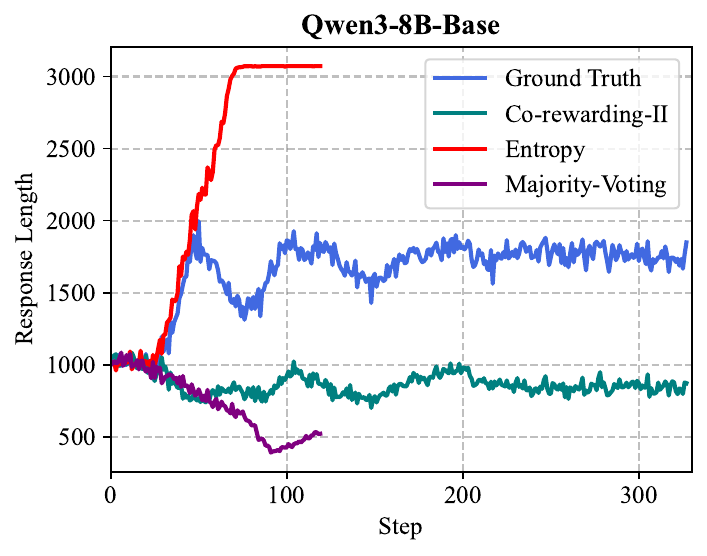}\\
    \includegraphics[width=0.49\linewidth]{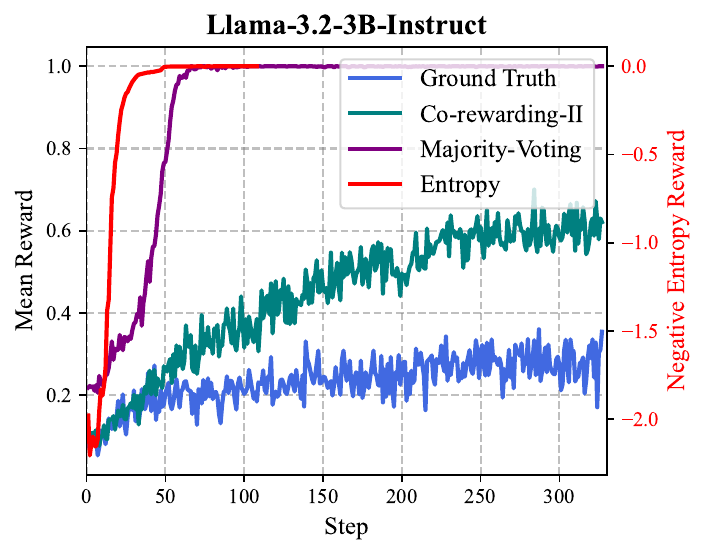}
    \includegraphics[width=0.49\linewidth]{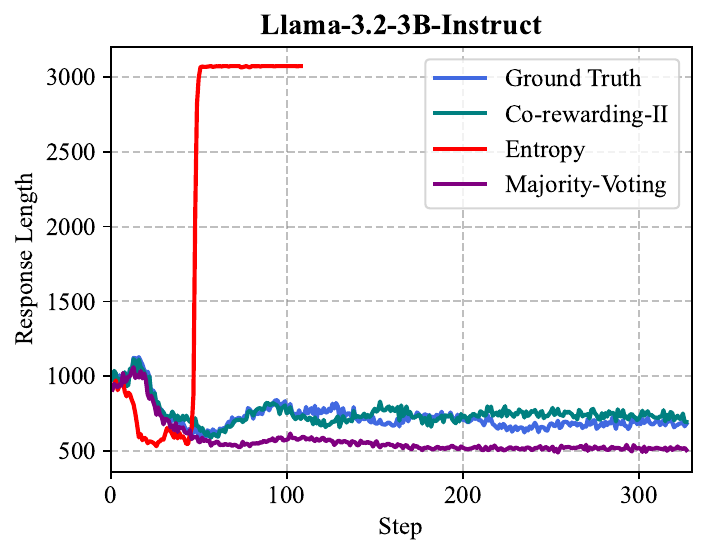}
    \vspace{-6mm}
    \caption{\textbf{Reward} (left) and \textbf{response length} (right) of Qwen3-8B-Base and Llama-3.2-3B-Instruct trained on DAPO-14k. Due to different reward scale from other methods, Entropy reward is plotted on the right $y$-axis of left panels, where the reward is the negative entropy.}
    \label{fig:reward_length_qwen3-8b_llama3b}
  \end{minipage}
  \vspace{-6mm}
\end{figure}

\par
\textbf{Implementation Details.} We implement our algorithms based on the VeRL framework~\citep{sheng2024hybridflow}, and experiments are conducted on 4 $\times$ H100-80GB GPUs. For our experiments, we totally use three training sets: MATH~\citep{hendrycksmath2021} (7,500 questions), DAPO-14k~\citep{yu2025dapo} (en-version of DAPO-Math-17k, about 14.1k questions), and OpenRS~\citep{dang2025reinforcement} (7,000 questions). During RL training, we use a global batch size of 128, set the number of rollouts to $G=\tilde{G}=8$ per question for Co-rewarding-I, II and III, and adopt AdamW with a learning rate of $3\times10^{-6}$. In Co-rewarding-I and III, question rephrasing is performed by the open-source Qwen3-32B model. In Co-rewarding-II and III, the EMA weight is scheduled from $\alpha_\textrm{start}=0.99$ to $\alpha_\textrm{end}=0.9999$ using cosine annealing. More implementation details are reported in Appendix~\ref{app:exp_train_details}.

\par
\textbf{Evaluation Details.} To provide a comprehensive evaluation of model capabilities, we utilize a diverse set of benchmarks spanning mathematical reasoning, code generation, instruction-following, and general multi-task abilities. Specifically: (1) Mathematical reasoning: MATH500~\citep{lightman2023let}, GSM8K~\citep{cobbe2021gsm8k}, AMC~\citep{li2024amc}, and AIME24~\citep{aime24}. (2) Code generation: LiveCodeBench~\citep{jainlivecodebench} release\_v6 and CRUX~\citep{gu2024cruxeval}. (3) Instruction-following and multi-task abilities: IFEval~\citep{zhou2023IFEval} and MMLU-Pro~\citep{wang2024mmlupro}. Additional evaluation details are provided in Appendix~\ref{app:eval_details}.

\vspace{-2mm}
\subsection{Experimental Results}

\subsubsection{Main Performance of Co-rewarding}
\label{Sec:exp_performance}
\textbf{Superior Performance of Co-rewarding over self-rewarding baselines.}
Table~\ref{Tble:Main_Exp_CorewardingV1} and Table~\ref{Tble:Main_Exp_CorewardingV2} report the experimental results trained on MATH and DAPO-14k, respectively. We observe that all three Co-rewarding instantiations (I, II, and III)  occupy more darker cells in the tables, demonstrating stronger performance than other self-rewarding SoTA baselines. Specifically, Co-rewarding-I achieves an average relative performance gain of $+4.42\%$ over the best baselines across four mathematical benchmarks and models in Table~\ref{Tble:Main_Exp_CorewardingV1}, while Co-rewarding-II achieves a larger average relative gain of $+12.90\%$ in Table~\ref{Tble:Main_Exp_CorewardingV2}. Moreover, Co-rewarding-III achieves improvements on average of $+7.11\%$ and $1.72\%$ over Co-rewarding-I and Co-rewarding-II, respectively, suggesting that integrating data-side cross-supervision with model-side self-distillation can further boost performance. Additional results on other training sets and LLMs are provided in Appendix~\ref{app:more_exp_LLM_trainingset}.

\par
\textbf{Surpassing GT-Reward on certain benchmarks.}
Surprisingly, we observe that all three Co-rewarding instantiations (I, II, and III) outperform GT-Reward in certain cases. For example, on GSM8K, they together achieve an average improvement of $+2.77\%$ over GT-Reward in Table~\ref{Tble:Main_Exp_CorewardingV1}, while Co-rewarding-II further delivers a larger gain of $+5.44\%$ in Table~\ref{Tble:Main_Exp_CorewardingV2}. Notably, Co-rewarding-II reaches a remarkably high Pass@1 of 94.01\% with Qwen3-8B-Base. This may be because GSM8K is a relatively easier benchmark, where self-supervised RL is sufficient to elicit latent reasoning abilities without GT labels. Additionally, Co-rewarding also shows advantages on coding benchmark CRUX in several cases, possibly due to distribution mismatch between training and evaluation data. This may offer opportunities for self-supervised methods to generalize on par with, or even surpass GT-supervised methods in some cases. These findings highlight the potential of self-supervised RL to elicit reasoning capabilities, particularly with Co-rewarding mitigating training collapse.

\begin{figure}[t]
  \centering
  \vspace{-2mm}
  % ---- 图片 ----
  \includegraphics[width=0.24\textwidth]{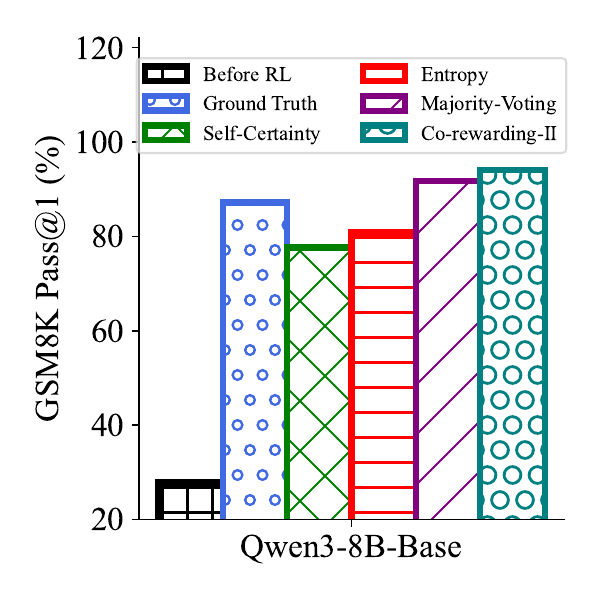}
  \includegraphics[width=0.24\textwidth]{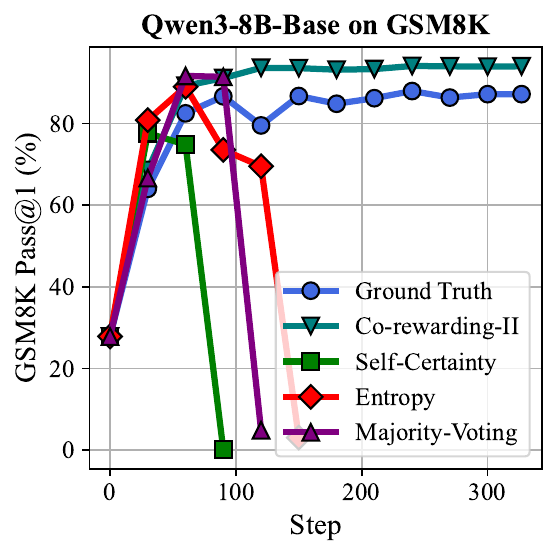}
  \includegraphics[width=0.24\textwidth]{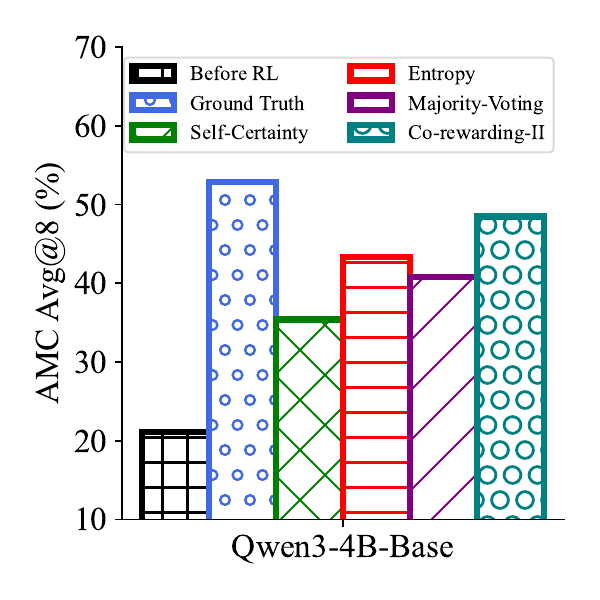}
  \includegraphics[width=0.24\textwidth]{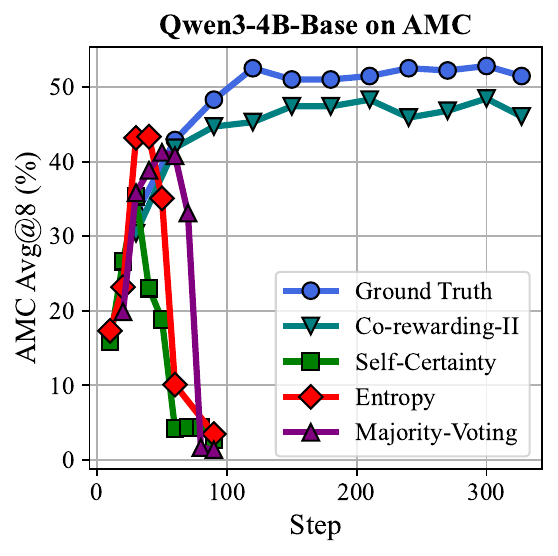}
  \vspace{-4mm}
  \caption{\textbf{Performance and Stability on GSM8K and AMC.} 
  The gains of Co-rewarding arise from its training stability, which supports continuous improvements throughout learning.}
  \label{fig:performance_stability}

  % ---- 表格 ----
  \vspace{-3mm}
  \captionof{table}{\textbf{Ablation study of Co-rewarding.} For Co-rewarding-I, ablations train only on original or rephrased data. For Co-rewarding-II, ablation removes EMA updates of the reference teacher.}
  \renewcommand\arraystretch{0.95}
  \resizebox{0.99\textwidth}{!}{
    \begin{tabular}{l|l|cccccccc}
    \toprule[1.6pt]
        \textbf{Training Set} & \textbf{Methods} & \textbf{MATH500} & \textbf{GSM8K} & \textbf{AMC} & \textbf{AIME24} & \textbf{LiveCode} & \textbf{CRUX} & \textbf{IFEval} & \textbf{MMLU-Pro} \\
        \midrule
        \multirow{14}{*}{\textbf{MATH}} & \multicolumn{9}{c}{\textit{\textbf{Qwen3-8B-Base}}} \\
        \cmidrule{2-10}
        ~ & Co-rewarding-I & \textbf{81.2} & \textbf{93.70} & \textbf{51.20} & 15.10 & 30.81 & \textbf{66.00} & \textbf{55.79} & \textbf{59.95} \\
        ~ & - Majority-Voting w/ Union & 80.2 & 93.48 & 49.70 & 15.63 & \textbf{31.94} & 64.88 & 54.25 & 59.51 \\
        ~ & - Majority-Voting w/ Original & 79.8 & 89.76 & 49.09 & \textbf{15.83} & 30.52 & 63.38 & 51.80 & 56.93 \\
        ~ & - Majority-Voting w/ Rephrased & 79.2 & 91.51 & 50.75 & 14.17 & 31.66 & 60.38 & 52.24 & 57.26 \\
        \cmidrule{2-10}
        ~ & Co-rewarding-II & \textbf{80.8} & \textbf{92.42} & \textbf{53.46} & \textbf{14.48} & 30.23 & \textbf{62.83} & \textbf{60.70} & \textbf{57.50} \\
        ~ & - w/o Updating Reference & 79.2 & 89.46 & 51.51 & 13.96 & \textbf{30.62} & 61.75 & 56.93 & 51.85 \\
        \cmidrule{2-10}
        ~ & \multicolumn{9}{c}{\textit{\textbf{Llama-3.2-3B-Instruct}}} \\
        \cmidrule{2-10}
        ~ & Co-rewarding-I & \textbf{50.2} & 79.45 & \textbf{23.80} & \textbf{10.00} & \textbf{11.28} & 29.88 & \textbf{48.89} & \textbf{33.77} \\
        ~ & - Majority-Voting w/ Union & 48.0 & \textbf{80.52} & 21.84 & 9.69 & 10.14 & 30.00 & 43.35 & 34.05 \\
        ~ & - Majority-Voting w/ Original & 46.8 & 78.77 & 20.48 & 9.27 & 11.00 & \textbf{31.25} & 47.96 & 33.18 \\
        ~ & - Majority-Voting w/ Rephrased & 44.0 & 78.85 & 21.23 & 8.85 & 10.04 & 17.25 & 47.84 & 33.72 \\
        \cmidrule{2-10}
        ~ & Co-rewarding-II & \textbf{49.8} & \textbf{79.30} & \textbf{22.59} & \textbf{10.73} & \textbf{10.80} & 30.63 & \textbf{49.90} & \textbf{33.61} \\
        ~ & - w/o Updating Reference & 47.0 & 78.92 & 22.29 & 9.06 & 5.50 & \textbf{31.25} & 47.88 & 33.32 \\
        \midrule
        \multirow{6}{*}{\textbf{DAPO-14k}} & \multicolumn{9}{c}{\textit{\textbf{Qwen3-8B-Base}}} \\
        \cmidrule{2-10}
        ~ & Co-rewarding-II & \textbf{80.6} & \textbf{94.01} & \textbf{54.37} & \textbf{16.35} & \textbf{31.66} & \textbf{67.12} & \textbf{53.31} & \textbf{59.83} \\
        ~ & - w/o Updating Reference & 78.0 & 88.40 & 51.66 & 15.94 & 30.62 & 63.75 & 52.48 & 58.01 \\
        \cmidrule{2-10}
        ~ & \multicolumn{9}{c}{\textit{\textbf{Llama-3.2-3B-Instruct}}} \\
        \cmidrule{2-10}
        ~ & Co-rewarding-II & \textbf{49.8} & \textbf{78.62} & \textbf{19.73} & \textbf{8.02} & \textbf{10.43} & \textbf{32.25} & \textbf{51.92} & \textbf{34.46} \\
        ~ & - w/o Updating Reference & 45.0 & 76.72 & 17.92 & \textbf{8.02} & 10.05 & 30.63 & 51.33 & 33.94 \\
        \bottomrule[1.6pt]
    \end{tabular}
  }
  \label{Tble:AblationStudy}
  \vspace{-7mm}
\end{figure}

\textbf{Code generalization with preserved general performance.} Although trained solely on math-oriented datasets, the models show improvements on coding benchmarks, suggesting a cross-domain generalization from math to code in self-supervised reasoning elicitation. Moreover, Co-rewarding preserves general instruction-following and multi-task ability on MMLU-Pro and IFEval. As shown in Table~\ref{Tble:MMLU-Pro_Qwen8B}, Co-rewarding-II outperforms other self-rewarding baselines in 12 of 14 MMLU-Pro categories, demonstrating that its gains do not come at the expense of broader general-domain performance. More detailed results of MMLU-Pro and IFEval refer to Appendix~\ref{appe:Detail_MMLU-Pro} and~\ref{appe:Detail_IFEval}.

\textbf{Importance of stability for performance gain.} 
As shown in Table~\ref{Tble:Main_Exp_CorewardingV2}, self-rewarding baselines exhibit noticeably limited performance gain in certain cases, such as Self-Certainty with Qwen3-4B-Base on GSM8K. Figure~\ref{fig:performance_stability} further reflects this by showing that baselines improve quickly at the beginning but soon collapse on GSM8K and AMC, whereas Co-rewarding sustains steady progress. This collapse restricts the baselines to effective training on only a small portion of the data, preventing further improvements with continued training. These observations underscore the importance of avoiding training collapse in self-supervised RL to unlock further performance gains.

\newlength{\rightimgheight}
\setlength{\rightimgheight}{0.1\textheight} % 自己微调这一项

\begin{figure}[t]
  \centering
  % 左：表
  \vspace{-4mm}
  \begin{minipage}[t]{0.73\textwidth}
    \vspace{-22mm}
    \centering 
    \captionof{table}{\textbf{Detailed performance of MMLU-Pro} with Qwen3-8B-Base trained on DAPO-14k. More results refer to Appendix~\ref{appe:Detail_MMLU-Pro}.} 
    \label{Tble:MMLU-Pro_Qwen8B} 
    \vspace{-2mm} 
    \resizebox{\linewidth}{!}{ 
      \begin{tabular}{l|ccccccc}
      \toprule[1.6pt]
          \multicolumn{8}{c}{\textit{\textbf{MMLU-Pro (Qwen3-8B-Base)}}} \\
      \midrule
          \textbf{Methods} 
          & \textbf{biology} & \textbf{business} & \textbf{chemistry} & \textbf{computer sci.} & \textbf{economics} & \textbf{health} & \textbf{history} \\ 
          \midrule
      - GT-Reward        & 77.96 & 70.85 & 60.42 & 61.95 & 71.33 & 59.79 & 51.44 \\
      \midrule
      - Self-Certainty   & 75.73 & 58.05 & 50.53 & 56.83 & 69.31 & 54.77 & 50.40 \\
      - Entropy          & 74.76 & 59.70 & 51.33 & 56.10 & 67.90 & 55.87 & 48.04 \\
      - Majority-Voting  & 75.32 & 61.47 & 54.24 & 58.29 & 69.67 & 58.20 & 49.34 \\
      - Co-rewarding-I   & 76.85 & 61.22 & 53.45 & 59.02 & 66.82 & 55.62 & 48.29 \\
      - Co-rewarding-II  & 76.71 & 68.69 & 64.58 & 61.71 & 68.25 & 56.85 & 51.71 \\
      \specialrule{1pt}{0.4ex}{0.4ex}
          \textbf{Methods} 
          & \textbf{law} & \textbf{math} & \textbf{other} & \textbf{philosophy} & \textbf{physics} & \textbf{psychology} & \textbf{engineering} \\ 
          \midrule
      - GT-Reward        & 31.52 & 73.28 & 56.28 & 52.71 & 61.97 & 67.30 & 46.14 \\
      \midrule
      - Self-Certainty   & 30.43 & 63.06 & 51.63 & 46.29 & 51.73 & 66.42 & 41.07 \\
      - Entropy          & 28.97 & 63.96 & 51.51 & 48.90 & 53.04 & 66.80 & 42.32 \\
      - Majority-Voting  & 31.16 & 64.62 & 52.27 & 48.90 & 53.27 & 66.92 & 40.97 \\
      - Co-rewarding-I   & 30.34 & 66.17 & 51.73 & 48.90 & 55.19 & 66.42 & 39.63 \\
      - Co-rewarding-II  & 31.16 & 72.17 & 52.49 & 52.10 & 63.21 & 68.17 & 47.16 \\
      \specialrule{1pt}{0.4ex}{0.4ex}
      \end{tabular}
    }
  \end{minipage}\hfill
  % 右：两图
  \begin{minipage}[t]{0.25\textwidth}
    \centering
    \includegraphics[width=\linewidth,height=\rightimgheight]{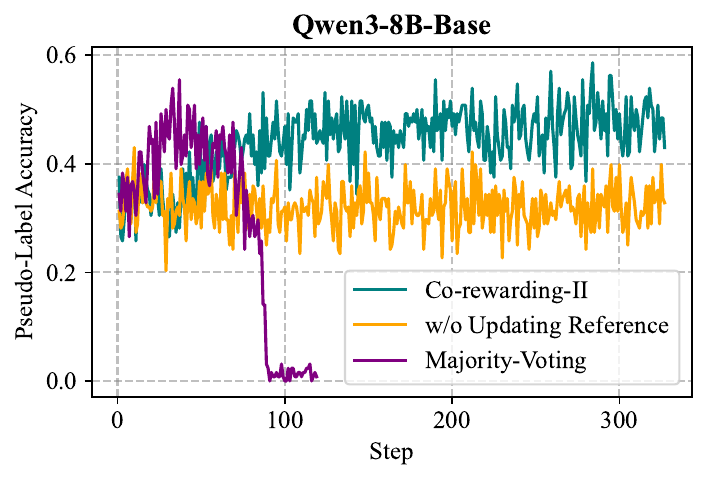}
    \includegraphics[width=\linewidth,height=\rightimgheight]{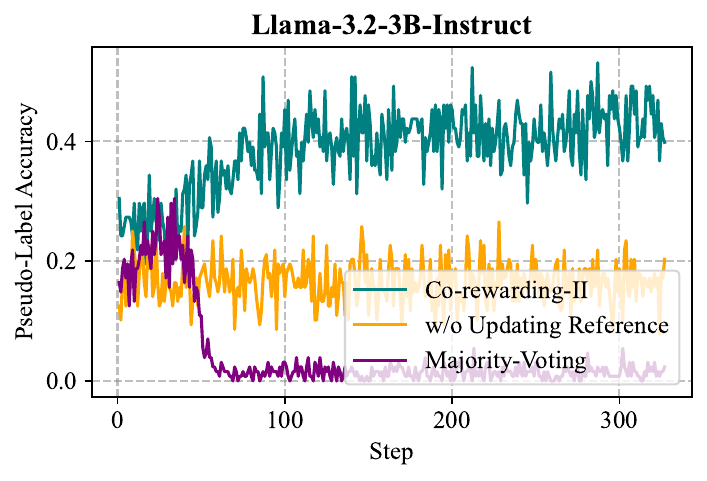}
    \vspace{-7mm}
    \captionof{figure}{\textbf{Pseudo label accuracy comparison.} 
    % of Co-rewarding-II, its ablation on ``w/o updating reference", and compared with Majority-Voting.
    }
    \label{fig:pseudo_acc_corewardv2}
  \end{minipage}
  \vspace{-2mm}
\end{figure}

\begin{figure}[t!]
  \centering
  \vspace{-2mm}
  \includegraphics[width=0.98\textwidth]{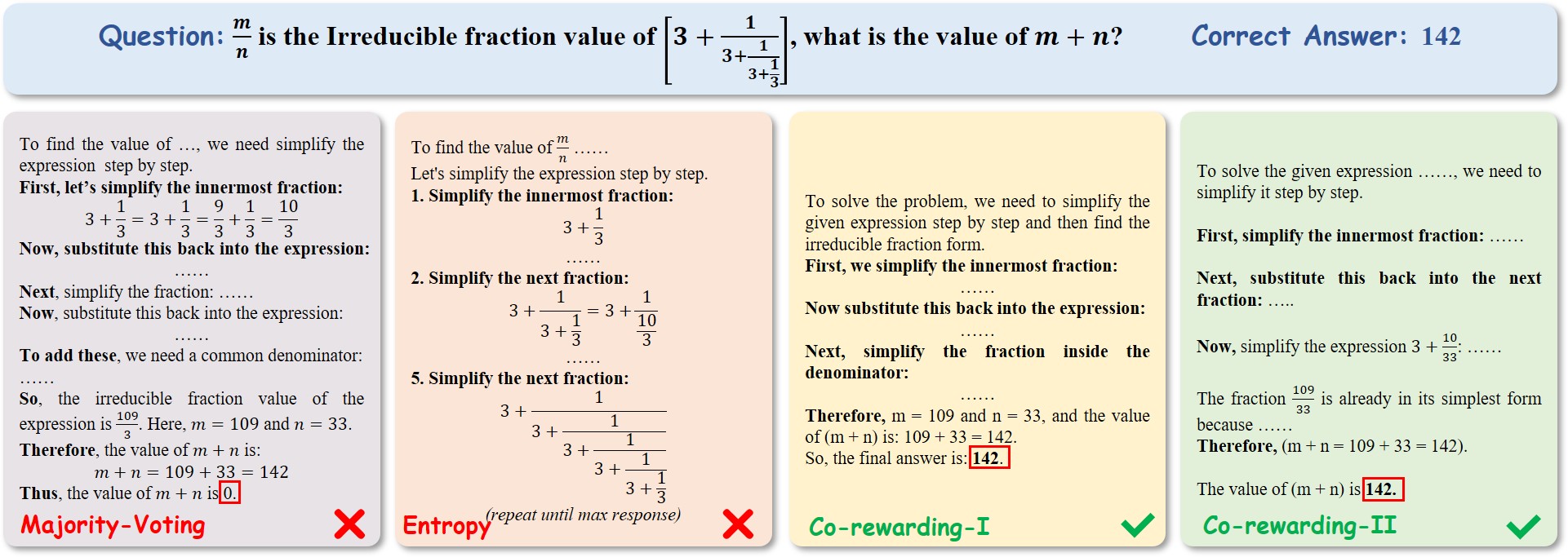}
  \vspace{-2mm}
  \caption{\textbf{Case study}: An example comparing the generations from \textit{Majority-Voting}, \textit{Entropy}, and our proposed \textit{Co-rewarding}. The results clearly reveal the reward hacking behavior exhibited by Majority-Voting and Entropy, while ours generate the correct answer. Full results refer to Appendix~\ref{app:case_study_generation}.}
  \label{Fig:Case_study}
  \vspace{-6mm}
\end{figure}

% We can find the learning of Co-rewarding is generally stable without significant performance drop across different LLMs.
\subsubsection{Further Analysis}
\textbf{Co-rewarding alleviates collapse and provides stable self-supervised RL.}
We use 5,000 questions from the MATH test split as a validation set to monitor training process. Figure~\ref{fig:curve_validation} shows that all three self-rewarding baselines collapse on both MATH and DAPO-14k. Co-rewarding-I remains stable on MATH but still collapses on DAPO-14k, suggesting that its stability depends on the property of training data. A plausible explanation is that the questions in MATH may provide favorable conditions for promoting diverse rephrasing variability, which is beneficial for the effectiveness of contrastive agreement in Co-rewarding-I. More discussions are provided in Appendix~\ref{app:discuss_MATH_DAPO14k}. In contrast, Co-rewarding-II consistently maintains stability across datasets, as its design decouples supervision from the online policy and thus breaks the entanglement between supervision and the policy itself.

\textbf{Co-rewarding attempts to balance exploration-exploitation.} 
Figure~\ref{fig:reward_length_qwen3-8b_llama3b} shows reward and response length curves. Entropy and Majority-Voting quickly reach the highest reward, indicating reward hacking rather than genuine reasoning improvement. In contrast, GT-Reward and Co-rewarding exhibit smoother, gradually increasing rewards, reflecting stable training. The response length curves further illustrate this difference: GT-Reward lengthens responses to explore correct reasoning paths; Majority-Voting collapses to short outputs, restricting exploration; and Entropy collapses its probability mass onto a small set of tokens, repeatedly generating them until truncation. Co-rewarding instead maintains moderate response lengths throughout training, suggesting a balanced exploration–exploitation trade-off. Additional curves for other LLMs are provided in Appendix~\ref{app:reward_response_length}.

\textbf{Each part contributes to Co-rewarding.}
Table~\ref{Tble:AblationStudy} summarizes the ablations across two training sets. For Co-rewarding-I, we observe that it typically outperforms all three variants of Majority-Voting: models trained only on original questions, only on rephrased questions, or on their union. This indicates that the cross-supervision between original and rephrased questions plays a key role in mitigating training collapse, whereas simply adding more data does not resolve the inherent instability of single-view self-rewarding methods. Notably, training only on the original or rephrased data yields comparable results, reflecting that the quality of original and rephrased data is similar. For Co-rewarding-II, removing the EMA update of the reference teacher model causes clear degradation, highlighting the necessity of teacher updates for improving pseudo-label quality.

\textbf{EMA is essential in Co-rewarding-II for improving pseudo-label quality.} 
Figure~\ref{fig:pseudo_acc_corewardv2} compares pseudo-label accuracy across Co-rewarding-II, ``w/o Updating Reference”, and Majority-Voting. Co-rewarding-II steadily improves accuracy as training progresses, while ``w/o Updating Reference” remains nearly flat, underscoring the role of EMA updates in allowing the teacher to co-evolve with the policy and generate higher-quality pseudo labels. By contrast, Majority-Voting briefly improves but then collapses to near zero, evidencing reward hacking through consistent yet incorrect outputs.

\textbf{Case Study of the model reasoning with different learning methods.} 
Figure~\ref{Fig:Case_study} provides a concrete example to illustrate the qualitative difference between self-rewarding baselines and our Co-rewarding. Majority-Voting exhibits reward hacking by boxing an incorrect answer ``0" to pursue consensus, even though the reasoning steps are correct. Entropy produces repetitive outputs as its decoding probability distribution collapses onto a narrow set of tokens during entropy minimization. In contrast, Co-rewarding generates coherent step-by-step reasoning and correctly boxes the final answer, showing its capacity to overcome reward hacking and elicit genuine reasoning. Full results are provided in Appendix~\ref{app:case_study_generation} and additional case studies on code benchmark are discussed in Appendix~\ref{app:case_study_code}.

\vspace{-2mm}
\section{Related Work}
\vspace{-1mm}
Reinforcement learning with verifiable reward (RLVR) has recently become a mainstream post-training paradigm for eliciting strong reasoning abilities in LLMs~\citep{guo2025deepseek}, achieving remarkably encouraging success particularly on mathematical~\citep{shao2024deepseekmath} and coding~\citep{deepcoder2025} tasks. However, RLVR fundamentally depends on high-quality and annotated GT labels to supervise reward signals, which remains a major bottleneck for scalability under the spirit of the scaling laws. To break this limitation, recent efforts have explored RL without external reward from multiple perspectives. For instance, methods such as TTRL~\citep{zuo2025ttrl} and SRT~\citep{shafayat2025can} pursue self-consistency to generate pseudo labels for rewards, where agreement among multiple rollouts is treated as optimization objective. Additionally, another technical line such as EMPO~\citep{zhang2025right}, Intuitor~\citep{zhao2025learning} and RENT~\citep{prabhudesai2025maximizing}, enhances the LLM confidence by optimizing internal signals of reasoning, such as entropy minimization or self-certainty maximization. Different from these studies, Co-rewarding focuses on mitigating inherent training collapse in existing methods and enables stable self-supervised RL training. More detailed discussions of related work are in Appendix~\ref{appe:related_work}.

\vspace{-2mm}

\section{Conclusion}
\vspace{-1mm}
In this work, we introduced Co-rewarding, a self-supervised RL framework that elicits the reasoning capability of LLMs through complementary supervision. Unlike prior self-rewarding methods that entangle rewards with single-view outputs and risk collapse, Co-rewarding establishes stability by decoupling the reward signal from the current online policy with the single-view output. Specifically, Co-rewarding-I leverages contrastive agreement across semantically analogous questions; Co-rewarding-II employs a dynamically updated teacher to provide insulated pseudo-labels; and Co-rewarding-III combines the data-side cross-supervision from Co-rewarding-I and the model-side teacher-based pseudo labels from Co-rewarding-II to further boost performance. Together, these designs construct cross-referable reward signals without explicit labels, aligning RL with invariances in reasoning rather than the mere correctness of isolated outputs. We hope this work will inspire further exploration into self-supervised RL for reasoning to advance the development.

\subsubsection*{Acknowledgments}
ZZZ, JNZ, ZKZ, XL, XF and BH were supported by NSFC Major Research Plan No. 92570109.

\section*{Ethics Statement}
This work complies with the Code of Ethics. It uses only publicly available datasets, involves no human or sensitive data, and raises no foreseeable risks related to privacy, security, or fairness issues. The research is conducted solely for scientific advancement, with no conflicts of interest.

\section*{Reproducibility Statement}

We are committed to ensure the reproducibility of our proposed method. A detailed description of our approach is provided in the Co-rewarding Framework section, and the corresponding source code has been submitted in an anonymous repository at~\url{https://github.com/tmlr-group/Co-rewarding}. Both backbone models and datasets used in our work are publicly available. Furthermore, all parameters, hyper-parameters, and procedural steps required to reproduce our results are thoroughly recorded in the Implementation Details. We believe that these components provide the community with details necessary to verify and reproduce our work.

\bibliography{iclr2026_conference}
\bibliographystyle{iclr2026_conference}

\clearpage
\appendix

\section*{LLM Usage Statement}

Here we clarify the usage of Large Language Models (LLMs) in this work. For the preparation of this paper, LLMs are limited to the role of a general-purpose writing assistant and are not used for research ideation or core content generation. For research methodology, LLM is a core component of our proposed method. Specifically, we utilize the Qwen3-32B model to perform question rephrasing in Co-rewarding-I, which is thoroughly detailed in the Implementation Details section of the main paper. The authors take full responsibility for all content written under their name.

\section{Related Work}
\label{appe:related_work}
\textbf{Large Language Model Reasoning.} 
% Reviewing advancements and challenges in LLM reasoning
LLMs have shown impressive performance on vast tasks that require reasoning, including solving mathematical problems, writing code, and answering logical questions. One of the key techniques that has improved LLM reasoning is Chain-of-Thought (CoT) prompting~\citep{wei2022chain,zhou2024can,zhou2025from}. CoT encourages the model to generate intermediate reasoning steps before producing the final answer~\citep{zhou2025landscape}, which has been shown to enhance performance on tasks like arithmetic, commonsense reasoning, and symbolic reasoning. Subsequent work has extended CoT by integrating it with various strategies, including compositional generalization~\citep{zhou2022least, khot2022decomposed} and employing structural reasoning approaches~\citep{yao2023tree, besta2024graph, yang2024buffer}. In addition, CoT serves as a fundamental framework for techniques like fine-tuninig~\citep{zelikman2022star}, argentic workflow~\citep{yao2023react}, and paving the way for inference-time scaling~\citep{snell2024scaling}.

\textbf{RL for Large Language Models.} 
Several RL algorithms have been developed primarily for alignment tasks~\citep{yue2025what, zhang2026towards, zhang2025minds}. 
Specifically, DPO~\citep{rafailov2023direct}, CPO~\citep{xu2024contrastive}, and their variants~\citep{li2023remax, guo2024direct, munos2024nash, hong2024orpo, xie2024exploratory} rely on preference pairs labeled by annotators. In contrast, KTO~\citep{ethayarajh2024kto} and BCO~\citep{jung2024binary} require only a single binary label (like or dislike) for each output. Besides, PRM~\citep{uesato2022solving, lightman2023let} and Step-KTO~\citep{lin2025step} offer step-by-step guidance by incorporating feedback at each reasoning step rather than focusing solely on the final outputs.
Recently, the follow-up work of GRPO improves the optimization objective, \textit{e.g.}, DAPO~\citep{yu2025dapo}, Dr. GRPO~\citep{liu2025understanding}, REINFORCE++~\citep{hu2025reinforce++}, CPPO~\citep{lin2025cppo}, and GPG~\citep{chu2025gpg}.
Another line of research generalizes GRPO to broader applications such as multimodal reasoning~\citep{zhou2025r1, huang2025vision, chu2025gpg, liu2025visual, zhang2025r1} and logical reasoning~\citep{xie2025logic}.

\textbf{RL without External Reward.} 
RL methods have shown promising scaling capabilities to enhance the reasoning abilities of LLMs~\citep{guo2025deepseek}, yet they are often limited by the availability of training data for reward signals~\citep{gao2023scaling, liu2023your}. Notably, Wang et al.~\citep{wang2025reinforcement} demonstrate that RL can effectively bootstrap LLM reasoning with as little as a single training example, highlighting the potential to minimize or even eliminate reliance on external reward signals during training. Recent efforts leverage distinct strategies for reward assignment. For instance, SIRLC~\citep{pang2024language} and AZR~\citep{zhao2025absolute} utilize an LLM-as-the-judge approach to assign rewards. In contrast, methods like SRT, TTRL, and their variants~\citep{shafayat2025can, zuo2025ttrl, fang2025serl, zhang2025consistent} employ self-consistency~\citep{wang2022self} to generate pseudo-rewards, reducing dependence on external annotations. Meanwhile, INTUITOR, RLSC, and RENT~\citep{zhao2025learning, li2025confidence, prabhudesai2025maximizing} harness the internal confidence scores of LLMs as intrinsic reward signals. Additionally, EMPO and its variants~\citep{zhang2025right, agarwal2025unreasonable} promote reasoning by minimizing entropy of reasoning path, further diversifying the approaches to incentivize robust LLM performance~\citep{han2025trustworthy}.

\begin{algorithm}[t]
   \caption{\textit{Co-rewarding-I}}
   \label{Alg:Co-rewarding_V1}
\begin{algorithmic}[1]
  \State \textbf{Input:} policy model $\pi_\theta$, learning rate $\eta$, training dataset $\mathcal{D}$, rephrased training dataset $\mathcal{D'}$, total iterations $K$.
  \State \textbf{Output:} trained policy model $\pi_\theta$.
  \ForAll{iteration $k=1,\dots,K$}
    \State Sample mini-batch inputs $\mathcal{B}\subseteq\mathcal{D}$ and $\mathcal{B}'\subseteq\mathcal{D}'$.
    \ForAll{input question $x \in \mathcal{B}$ and $x' \in \mathcal{B}'$}
      \State Sample rollouts $\{y_i\}_{i=1}^G \sim \pi_{\theta_{\textrm{old}}}(\cdot\mid x)$.
      \State Sample rollouts $\{y'_i\}_{i=1}^{G'} \sim \pi_{\theta_{\textrm{old}}}(\cdot\mid x')$.
      \State Obtain pseudo labels by Eq.~(\ref{eqn:co-reward_vote}).
      \State Estimate relative advantages by Eq.~(\ref{Eq:CorewardingV1_adv}).
      \State Compute the objective by Eq.~(\ref{eqn:co-reward_obj}).
      \State Update $\theta \leftarrow \theta - \eta \nabla_\theta \mathcal{J}_{\mathrm{Co\text{-}rewarding\text{-}I}}(\theta)$.
    \EndFor
  \EndFor
\end{algorithmic}
\end{algorithm}

\begin{algorithm}[!ht]
   \caption{\textit{Co-rewarding-II}}
   \label{alg:corewarding_v2}
\begin{algorithmic}[1]
  \State \textbf{Input:} policy model $\pi_\theta$, learning rate $\eta$, training dataset $\mathcal{D}$, total iterations $K$.
  \State \textbf{Output:} trained policy model $\pi_\theta$.
  \For{iteration $k = 1, \dots, K$}
    \State Sample mini-batch $\mathcal{B}\subseteq\mathcal{D}$.
    \ForAll{$x \in \mathcal{B}$}
      \State Sample rollouts $\{y_i\}_{i=1}^G \sim \pi_{\theta_{\mathrm{old}}}^{(k)}(\cdot \mid x)$.
      \State Update the reference teacher by Eq.~(\ref{Eq:CorewardingV2_updateTeacher}).
      \State Sample rollouts $\{\tilde{y}_j\}_{j=1}^{\tilde{G}} \sim \tilde{\pi}_{\mathrm{ref}}^{(k)}(\cdot \mid x)$.
      \State Obtain pseudo label from $\{\tilde{y}_j\}_{j=1}^{\tilde{G}}$ by Eq.~(\ref{Eq:CorewardingV2_label}).
      \State Estimate the relative advantage by Eq.~(\ref{Eq:CorewardingV2_label}).
      \State Compute the objective by Eq.~(\ref{Eq:Co-rewarding_obj}).
      \State Update $\theta \leftarrow \theta - \eta \nabla_\theta \mathcal{J}_{\mathrm{Co\text{-}rewarding\text{-}II}}^{(k)}(\theta)$.
    \EndFor
  \EndFor
\end{algorithmic}
\end{algorithm}

\begin{algorithm}[!ht]
   \caption{\textit{Co-rewarding-III}}
   \label{alg:corewarding_v3}
\begin{algorithmic}[1]
  \State \textbf{Input:} policy model $\pi_\theta$, learning rate $\eta$, original training dataset $\mathcal{D}$, rephrased training dataset $\mathcal{D'}$, total iterations $K$.
  \State \textbf{Output:} trained policy model $\pi_\theta$.
  \For{iteration $k = 1, \dots, K$}
    \State Sample mini-batch inputs $\mathcal{B}\subseteq\mathcal{D}$ and $\mathcal{B}'\subseteq\mathcal{D}'$.
    \ForAll{$x \in \mathcal{B}$ and $x' \in \mathcal{B}'$}
      \State Sample rollouts $\{y_i\}_{i=1}^G \sim \pi_{\theta_{\mathrm{old}}}^{(k)}(\cdot \mid x)$ and $\{y_i'\}_{i=1}^G \sim \pi_{\theta_{\mathrm{old}}}^{(k)}(\cdot \mid x')$.
      \State Update the reference teacher by Eq.~(\ref{Eq:CorewardingV2_updateTeacher}).
      \State Sample rollouts $\{\tilde{y}_j\}_{j=1}^{\tilde{G}} \sim \tilde{\pi}_{\mathrm{ref}}^{(k)}(\cdot \mid x)$ and $\{\tilde{y}_j'\}_{j=1}^{\tilde{G}} \sim \tilde{\pi}_{\mathrm{ref}}^{(k)}(\cdot \mid x')$.
      \State Obtain pseudo label from $\{\tilde{y}^\prime_j\}_{j=1}^{\tilde{G}}$ and $\{\tilde{y}_j\}_{j=1}^{\tilde{G}}$ by Eq.~(\ref{Eq:CorewardingV3_label_1}) and Eq.~(\ref{Eq:CorewardingV3_label_2}).
      \State Estimate the relative advantages by Eq.~(\ref{Eq:CorewardingV3_label_1}) and Eq.~(\ref{Eq:CorewardingV3_label_2}).
      \State Compute the objective by Eq.~(\ref{Eq:Co-rewarding-III_obj}).
      \State Update $\theta \leftarrow \theta - \eta \nabla_\theta \mathcal{J}_{\mathrm{Co\text{-}rewarding\text{-}III}}^{(k)}(\theta)$.
    \EndFor
  \EndFor
\end{algorithmic}
\end{algorithm}

\section{Pseudo Code of Co-rewarding}
\subsection{Formulation of Co-rewarding-III}
\label{app:formulation_III}
The relative advantages $\mathcal{R}_\theta(\hat{A}^{(k)})$ and $\mathcal{R}_\theta(\hat{A}'^{(k)})$ are computed as:
\begin{align}
    &\hat{A}_i^{(k)} = \frac{r(\textcolor{brown}{\tilde{y}_\textrm{v}^{\prime (k)}}, \textcolor{blue}{y_i}) - \textrm{mean} (\{r(\textcolor{brown}{\tilde{y}_\textrm{v}^{\prime (k)}}, \textcolor{blue}{y_i})\}_{i=1}^G)}{\textrm{std}(\{r(\textcolor{brown}{\tilde{y}_\textrm{v}^{\prime (k)}}, \textcolor{blue}{y_i})\}_{i=1}^G)}, \; \textcolor{brown}{\tilde{y}_\textrm{v}^{\prime (k)}} = \arg \max_{y*} \sum_{j=1}^{\tilde{G}} \mathbf{1}[\textrm{ans}({\textcolor{brown}{\tilde{y}^\prime_j}})=\textrm{ans}(y*)], \label{Eq:CorewardingV3_label_1} \\
    &\hat{A}_i^{\prime (k)} = \frac{r(\textcolor{blue}{\tilde{y}_\textrm{v}^{(k)}}, \textcolor{brown}{y^\prime_i}) - \textrm{mean} (\{r(\textcolor{blue}{\tilde{y}_\textrm{v}^{(k)}}, \textcolor{brown}{y^\prime_i})\}_{i=1}^G)}{\textrm{std}(\{r(\textcolor{blue}{\tilde{y}_\textrm{v}^{(k)}}, \textcolor{brown}{y^\prime_i})\}_{i=1}^G)}, \; \textcolor{blue}{\tilde{y}_\textrm{v}^{(k)}} = \arg \max_{y*} \sum_{j=1}^{\tilde{G}} \mathbf{1}[\textrm{ans}({\textcolor{blue}{\tilde{y}_j}})=\textrm{ans}(y*)], \label{Eq:CorewardingV3_label_2}
\end{align}
where the pseudo label $\tilde{y}_\textrm{v}^{\prime (k)}$ is the majority-vote pseudo label obtained from reference rollouts on the rephrased question, and $\tilde{y}_\textrm{v}^{(k)}$ is the corresponding pseudo label obtained from reference rollouts on original question. The reference model is slowly updated via EMA as in Eq.~(\ref{Eq:CorewardingV2_updateTeacher}).

\subsection{Pseudo Code}
To intuitively present the pipeline of Co-rewarding, we summarize the pseudo codes of Co-rewarding-I, Co-rewarding-II and Co-rewarding-III in Algorithm~\ref{Alg:Co-rewarding_V1}, Algorithm~\ref{alg:corewarding_v2} and Algorithm~\ref{alg:corewarding_v3}, respectively.

\section{Additional Experimental Details}
\label{app:exp}
\subsection{Details of Baselines}
\label{app:details_baseline}
We compare our proposed Co-rewarding-I and II against GT-reward and several recent state-of-the-art (SoTA) self-reward approaches:
\begin{itemize}[left=5pt]
    \item \textbf{GT-Reward}~\citep{shao2024deepseekmath}: Originally introduced by DeepSeek-R1~\citep{guo2025deepseek}, GT-Reward supervises training using ground-truth (GT) answers, determining whether model rollouts are correct or not, to guide RL optimization.
    \item \textbf{Self-Certainty}~\citep{zhao2025learning}: This method maximizes \textit{self-certainty}, defined as the KL-divergence between the uniform distribution and the model’s decoding distribution, serving as reward to encourage more confident predictions.
    \item \textbf{Entropy}~\citep{prabhudesai2025maximizing}: This method minimizes the entropy of the model's rollout distribution, using negative entropy as reward to maximize model confidence.
    \item \textbf{Majority-Voting}~\citep{shafayat2025can}: By generating multiple rollouts per question, Majority-Voting selects the most frequent answer as a pseudo-label to supervise training.
\end{itemize}
For all methods, we adopt the widely used GRPO as the policy optimization algorithm.

\subsection{More Implementation Details}
\label{app:exp_train_details}
The detailed training configurations are summarized in Table~\ref{Tble:app_training_details}, and all baseline methods are trained under the same setup for fairness. For the training system prompt, we adopt the official default prompt provided by VeRL\footnote{https://github.com/volcengine/verl}, shown below:
\begin{lstlisting}
Let's think step by step and output the final answer within \boxed{}.
\end{lstlisting}
In addition, the semantically analogical questions used in Co-rewarding-I are generated by Qwen3-32B through a rewriting prompt. The exact rewriting instruction is provided as follows:

\begin{lstlisting}
You are given a math problem. Please rewrite it using different wording and a different real-world scenario, while keeping the underlying mathematical meaning and answer exactly the same.

Guidelines:
1. Do not change the math logic or the final answer.
2. Use different words and a new context to make it look like a different problem.
3. Avoid copying phrases or sentence structures from the original.
4. Make sure the rewritten question is natural, clear, and solvable.
5. Output ONLY between the following markers, and strictly in this format (no extra explanation):

### RESULT_START
ORIGINAL:
<original question>
REWRITE:
<rewritten question>
### RESULT_END
\end{lstlisting}

\begin{table}[t]
\caption{\textbf{Detailed training settings.}}
\begin{center}
% \resizebox{0.8\columnwidth}{!}{
    \begin{tabular}{l|ccc}
    \toprule[1.5pt]
    \textbf{Settings} & \textbf{Co-rewarding-I} & \textbf{Co-rewarding-II} & \textbf{Co-rewarding-III} \\
    \midrule
    Batch Size & 128 & 128 & 128 \\
    Max Prompt Length & 512 & 512 & 512 \\
    Max Response Length & 3,072 & 3,072 & 3,072 \\
    Train Steps & \multicolumn{3}{c}{170-220 (MATH), 300-330 (DAPO-14k), 100-130 (Open-RS)} \\
    Learning Rate & 3e-6 & 3e-6 & 3e-6 \\
    \# Policy Rollout $G$ & 8 & 8 & 8 \\
    \# Reference Rollout $\tilde{G}$ & - & 8 & 8 \\
    Clip Ratio & 0.2 & 0.2 & 0.2 \\
    Warmup Style & Cosine & Cosine & Cosine \\
    Warmup Steps Ratio & 0.1 & 0.1 & 0.1 \\
    KL Loss Coefficient & 0.005 & 0.001 & 0.001 \\
    Optimizer & \multicolumn{3}{c}{AdamW ($\beta_1=0.9$, $\beta_2=0.999$, $\epsilon=10^{-8}$)} \\
    Training Temperature & 1.0 & 1.0 & 1.0 \\
    Evaluation Temperature & 0.8 & 0.8 & 0.8 \\
    EMA $\alpha_\textrm{start}$ & - & 0.99 & 0.99 \\
    EMA $\alpha_\textrm{end}$ & - & 0.9999 & 0.9999 \\
    \bottomrule[1.5pt]
    \end{tabular}
\label{Tble:app_training_details}
\end{center}
\end{table}

\begin{table}[t!]
\caption{\textbf{Statistics and usages of datasets} used in our experiments.}
\begin{center}
% \resizebox{0.8\columnwidth}{!}{
    \begin{tabular}{l|cc}
    \toprule[1.5pt]
    \textbf{Dataset Name} & \textbf{\# Data Size} & \textbf{Usage} \\
    \midrule
    MATH-Train~\citep{hendrycksmath2021} & 7,500 & Training Set \\
    MATH-Test~\citep{hendrycksmath2021} & 5,000 & Validation Set \\
    DAPO-14k~\citep{yu2025dapo} & 14,109 & Training Set \\
    Open-RS~\citep{dang2025reinforcement} & 7,000 & Training Set \\
    \midrule
    MATH500~\citep{lightman2023let} & 500 & Evaluation Benchmark \\
    GSM8K~\citep{cobbe2021gsm8k} & 1,319 & Evaluation Benchmark \\
    AMC~\citep{li2024amc} & 83 & Evaluation Benchmark \\
    LiveCodeBench~\citep{jainlivecodebench} & 1,055 & Evaluation Benchmark \\
    CRUX~\citep{gu2024cruxeval} & 800 & Evaluation Benchmark \\
    MMLU-Pro~\citep{wang2024mmlupro} & 12,032 & Evaluation Benchmark \\
    IFEval~\citep{zhou2023IFEval} & 541 & Evaluation Benchmark \\
    \bottomrule[1.5pt]
    \end{tabular}
\label{Tble:dataset_details}
\end{center}
\end{table}

\subsection{More Evaluation Details}
\label{app:eval_details}
We conduct the evaluation across a diverse set of benchmarks, spanning mathematical reasoning, code generation, instruction-following, and general multi-task abilities. Specifically: (1) Mathematical reasoning: We evaluate on MATH500~\citep{lightman2023let}, GSM8K~\citep{cobbe2021gsm8k}, and AMC~\citep{li2024amc}. For MATH500 and GSM8K, we report pass@1 accuracy using the \texttt{lighteval} library\footnote{https://github.com/huggingface/lighteval}. For AMC, we use the \texttt{ttrl}\footnote{https://github.com/ruixin31/Spurious\_Rewards/tree/main/code/ttrl} library and report avg@8 as the metric. (2) Code generation: We assess coding ability using LiveCodeBench~\citep{jainlivecodebench} release\_v6 and CRUX~\citep{gu2024cruxeval}. LiveCodeBench is evaluated with its \texttt{official} evaluation library\footnote{https://github.com/LiveCodeBench/LiveCodeBench}, and CRUX is evaluated via the \texttt{ZeroEval} library\footnote{https://github.com/WildEval/ZeroEval}; for both datasets, we report pass@1 accuracy. (3) Instruction-following and multi-task abilities: We evaluate on IFEval~\citep{zhou2023IFEval} and MMLU-Pro~\citep{wang2024mmlupro}, using the \texttt{lm-evaluation-harness} library\footnote{https://github.com/EleutherAI/lm-evaluation-harness} for both. Overall, we summarize the statistics of the datasets used in this paper in Table~\ref{Tble:dataset_details}.

\begin{table}[t]
    \centering
    \caption{\textbf{Supplement Results (\%) of Co-rewarding and baselines trained on MATH}. Cell background colors: darker colors denote better results within each model group.}
    \label{Tble:Appe_Exp_CorewardingV1}
    % \vspace{2mm}
    \resizebox{\textwidth}{!}{
    \begin{tabular}{l|cccccccc}
    \toprule[1.6pt]
        \textbf{Training Set: MATH} & \multicolumn{4}{c}{\textbf{Mathematics}} & \multicolumn{2}{c}{\textbf{Code}} & \textbf{Instruction} & \textbf{Multi-Task} \\ \cmidrule{1-9}
        \textbf{Methods} & \textbf{MATH500} & \textbf{GSM8K} & \textbf{AMC} & \textbf{AIME24} & \textbf{LiveCode} & \textbf{CRUX} & \textbf{IFEval} & \textbf{MMLU-Pro} \\ 
        % Metric & ~ & Pass@1 & Pass@1 & Accuracy & pass@1 & pass@1 \\ 
        \specialrule{1pt}{0.4ex}{0.4ex}
            \multicolumn{9}{c}{\textit{\textbf{Qwen2.5-3B}}} \\
        \midrule
        Before RL & 53.6 & 19.48 & 10.69 & 0.52 & 9.95 & 18.50 & 29.83 & 32.50 \\ 
        - GT-Reward~\citep{shao2024deepseekmath} & 65.4 & 82.18 & 32.98 & 6.77 & 13.93 &  32.12 & 33.66 & 36.74 \\
        \midrule
        - Self-Certainty~\citep{zhao2025learning}   & \cellcolor{lightblue2}{64.2} & \cellcolor{lightblue2}{80.52} & \cellcolor{lightblue1}{28.92} & \cellcolor{lightblue2}{5.00} & \cellcolor{lightblue2}{10.90} & \cellcolor{lightblue2}{29.00} & \cellcolor{lightblue1}{32.22} & \cellcolor{lightblue2}{33.88} \\
        - Entropy~\citep{prabhudesai2025maximizing} & \cellcolor{lightblue1}{63.2} & \cellcolor{lightblue1}{80.44} & \cellcolor{lightblue2}{29.67} & \cellcolor{lightblue5}{5.94} & \cellcolor{lightblue1}{9.05} &  \cellcolor{lightblue2}{29.00} & \cellcolor{lightblue3}{32.94} & \cellcolor{lightblue3}{35.35} \\
        - Majority-Voting~\citep{shafayat2025can}   & \cellcolor{lightblue3}{64.6} & \cellcolor{lightblue4}{82.41} & \cellcolor{lightblue5}{33.13} & \cellcolor{lightblue3}{5.10} & \cellcolor{lightblue3}{14.03} & \cellcolor{lightblue3}{36.38} & \cellcolor{lightblue5}{35.19} & \cellcolor{lightblue4}{35.50} \\
        - Co-rewarding-I (Ours)                     & \cellcolor{lightblue5}{65.4} & \cellcolor{lightblue5}{84.53} & \cellcolor{lightblue3}{30.57} & \cellcolor{lightblue4}{5.31} & \cellcolor{lightblue4}{16.40} & \cellcolor{lightblue4}{36.88} & \cellcolor{lightblue4}{33.86} & \cellcolor{lightblue5}{36.38} \\
        - Co-rewarding-II (Ours)                    & \cellcolor{lightblue4}{65.2} & \cellcolor{lightblue3}{81.72} & \cellcolor{lightblue4}{32.38} & \cellcolor{lightblue1}{4.47} & \cellcolor{lightblue5}{22.25} & \cellcolor{lightblue5}{40.25} & \cellcolor{lightblue2}{32.74} & \cellcolor{lightblue1}{30.79} \\
        
        \specialrule{1pt}{0.4ex}{0.4ex}
            \multicolumn{9}{c}{\textit{\textbf{Qwen2.5-7B}}} \\
        \midrule
        Before RL & 69.4 & 24.71 & 15.81 & 2.81 & 3.79 & 26.38 & 38.19 & 44.76 \\ 
        - GT-Reward~\citep{shao2024deepseekmath} & 76.4 & 88.02 & 45.63 & 14.06 & 15.92 & 45.12 & 41.49 & 41.12 \\
        \midrule
        - Self-Certainty~\citep{zhao2025learning}   & \cellcolor{lightpurple2}{72.8} & \cellcolor{lightpurple2}{84.31} & \cellcolor{lightpurple1}{38.55} & \cellcolor{lightpurple1}{8.75} & \cellcolor{lightpurple2}{12.04} & \cellcolor{lightpurple4}{54.12} & \cellcolor{lightpurple1}{37.24} & \cellcolor{lightpurple4}{43.30} \\
        
        - Entropy~\citep{prabhudesai2025maximizing} & \cellcolor{lightpurple1}{72.2} & \cellcolor{lightpurple1}{81.43} & \cellcolor{lightpurple2}{39.61} & \cellcolor{lightpurple3}{10.73} & \cellcolor{lightpurple5}{16.49} & \cellcolor{lightpurple3}{51.88} & \cellcolor{lightpurple3}{40.33} & \cellcolor{lightpurple3}{42.79} \\
        
        - Majority-Voting~\citep{shafayat2025can}   & \cellcolor{lightpurple4}{74.4} & \cellcolor{lightpurple3}{84.53} & \cellcolor{lightpurple3}{40.96} & \cellcolor{lightpurple4}{11.04} & \cellcolor{lightpurple3}{15.45} & \cellcolor{lightpurple2}{51.00} & \cellcolor{lightpurple2}{38.60} & \cellcolor{lightpurple5}{43.35} \\
        
        - Co-rewarding-I (Ours)                     & \cellcolor{lightpurple5}{74.6} & \cellcolor{lightpurple5}{89.61} & \cellcolor{lightpurple4}{41.27} & \cellcolor{lightpurple3}{10.73} & \cellcolor{lightpurple4}{15.73} & \cellcolor{lightpurple5}{55.58} & \cellcolor{lightpurple5}{42.86} & \cellcolor{lightpurple2}{40.51} \\
        
        - Co-rewarding-II (Ours)                    & \cellcolor{lightpurple3}{73.6} & \cellcolor{lightpurple4}{89.31} & \cellcolor{lightpurple5}{42.77} & \cellcolor{lightpurple5}{11.98} & \cellcolor{lightpurple1}{8.25} & \cellcolor{lightpurple1}{47.50} & \cellcolor{lightpurple4}{41.82} & \cellcolor{lightpurple1}{37.45} \\

        \specialrule{1pt}{0.4ex}{0.4ex}
            \multicolumn{9}{c}{\textit{\textbf{Qwen3-1.7B-Base}}} \\
        \midrule
        Before RL & 57.0 & 19.56 & 8.43 & 1.15 & 4.45 & 7.50 & 33.65 & 33.00 \\ 
        - GT-Reward~\citep{shao2024deepseekmath} & 69.6 & 81.57 & 35.54 & 8.23 & 13.74 & 35.25 & 36.16 & 39.12 \\
        \midrule
        - Self-Certainty~\citep{zhao2025learning}   & \cellcolor{lightgreen1}{58.2} & \cellcolor{lightgreen1}{40.25} & \cellcolor{lightgreen1}{23.04} & \cellcolor{lightgreen1}{3.02} & \cellcolor{lightgreen1}{9.86} & \cellcolor{lightgreen1}{18.00} & \cellcolor{lightgreen1}{32.96} & \cellcolor{lightgreen1}{35.13} \\
        
        - Entropy~\citep{prabhudesai2025maximizing} & \cellcolor{lightgreen2}{63.6} & \cellcolor{lightgreen2}{71.79} & \cellcolor{lightgreen2}{31.63} & \cellcolor{lightgreen2}{6.88} & \cellcolor{lightgreen4}{13.74} & \cellcolor{lightgreen2}{31.37} & \cellcolor{lightgreen2}{35.37} & \cellcolor{lightgreen4}{36.67} \\
        
        - Majority-Voting~\citep{shafayat2025can}   & \cellcolor{lightgreen3}{65.2} & \cellcolor{lightgreen4}{81.57} & \cellcolor{lightgreen5}{34.78} & \cellcolor{lightgreen4}{7.50} & \cellcolor{lightgreen2}{13.08} & \cellcolor{lightgreen5}{34.25} & \cellcolor{lightgreen3}{35.45} & \cellcolor{lightgreen3}{36.00} \\
        
        - Co-rewarding-I (Ours)                     & \cellcolor{lightgreen5}{67.6} & \cellcolor{lightgreen5}{83.01} & \cellcolor{lightgreen3}{32.22} & \cellcolor{lightgreen5}{8.65} & \cellcolor{lightgreen3}{13.50} & \cellcolor{lightgreen3}{32.38} & \cellcolor{lightgreen4}{35.56} & \cellcolor{lightgreen2}{35.53} \\
        
        - Co-rewarding-II (Ours)                    & \cellcolor{lightgreen4}{66.2} & \cellcolor{lightgreen3}{80.89} & \cellcolor{lightgreen4}{33.28} & \cellcolor{lightgreen4}{7.50} & \cellcolor{lightgreen5}{14.40} & \cellcolor{lightgreen4}{32.88} & \cellcolor{lightgreen5}{36.94} & \cellcolor{lightgreen5}{37.59} \\
            
        \bottomrule[1.6pt]
    \end{tabular}}
    % \vspace{-4mm}
\end{table}

% ===================================================================

\begin{table}[t!]
    \centering
    \caption{\textbf{Supplement Results (\%) of Co-rewarding and baselines trained on OpenRS}. Cell background colors: darker colors denote better results within each model group. 
    % Bolded values indicate results that surpass both the ``GT-Reward" baseline and the ``Before RL".
    }
    \label{Tble:Supplement_Exp_CorewardingV2_OpenRS}
    % \vspace{2mm}
    \resizebox{\textwidth}{!}{
    \begin{tabular}{l|ccccccc}
    \toprule[1.6pt]
        \textbf{Training Set: Open-RS} & \multicolumn{3}{c}{\textbf{Mathematics}} & \multicolumn{2}{c}{\textbf{Code}} & \textbf{Instruction} & \textbf{Multi-Task} \\ \cmidrule{1-8}
        \textbf{Methods} & \textbf{MATH500} & \textbf{GSM8K} & \textbf{AMC} & \textbf{LiveCode} & \textbf{CRUX} & \textbf{IFEval} & \textbf{MMLU-Pro} \\ 
        % Metric & ~ & Pass@1 & Pass@1 & Accuracy & pass@1 & pass@1 \\ 
        \midrule
                \multicolumn{8}{c}{\textit{\textbf{Qwen3-8B-Base}}} \\ 
        \midrule
        Before RL & 72.40 & 27.82 & 20.93 & 23.41 & 54.75 & 50.89 & 52.92 \\ 
        - GT-Reward~\citep{shao2024deepseekmath} & 80.20 & 89.76 & 54.97 & 39.00 & 63.00 & 52.94 & 55.49 \\ 
        \midrule
        - Self-Certainty~\citep{zhao2025learning}   & \cellcolor{lightred5}82.60 & \cellcolor{lightred2}85.22 & \cellcolor{lightred2}50.00 & \cellcolor{lightred3}37.00 & \cellcolor{lightred5}64.62 & \cellcolor{lightred2}52.12 & \cellcolor{lightred2}56.03 \\ 
        
        - Entropy~\citep{prabhudesai2025maximizing} & \cellcolor{lightred4}80.60 & \cellcolor{lightred3}87.41 & \cellcolor{lightred1}48.95 & \cellcolor{lightred4}38.00 & \cellcolor{lightred2}61.25 & \cellcolor{lightred3}52.53 & \cellcolor{lightred4}56.80 \\
        
        - Majority-Voting~\citep{shafayat2025can}   & \cellcolor{lightred1}78.00 & \cellcolor{lightred1}84.23 & \cellcolor{lightred4}51.96 & \cellcolor{lightred2}36.75 & \cellcolor{lightred1}58.00 & \cellcolor{lightred1}51.13 & \cellcolor{lightred1}54.92 \\
        
        - Co-rewarding-I (Ours)                     & \cellcolor{lightred2}78.20 & \cellcolor{lightred5}92.65 & \cellcolor{lightred3}50.60 & \cellcolor{lightred1}28.91 & \cellcolor{lightred4}63.12 & \cellcolor{lightred5}53.11 & \cellcolor{lightred5}57.21 \\
        
        - Co-rewarding-II (Ours)                    & \cellcolor{lightred3}80.00 & \cellcolor{lightred4}90.90 & \cellcolor{lightred5}53.01 & \cellcolor{lightred5}39.75 & \cellcolor{lightred3}62.75 & \cellcolor{lightred4}52.92 & \cellcolor{lightred3}56.55 \\ 
        \specialrule{1pt}{0.4ex}{0.4ex}
            \multicolumn{8}{c}{\textit{\textbf{Qwen3-4B-Base}}} \\ 
        \midrule
        Before RL & 71.20 & 26.15 & 21.08 & 11.00 & 38.88 & 46.43 & 47.23 \\ 
        - GT-Reward~\citep{shao2024deepseekmath} & 78.80 & 85.22 & 49.55 & 33.50 & 55.12 & 46.41 & 50.12 \\ 
        \midrule
        - Self-Certainty~\citep{zhao2025learning}   & \cellcolor{lightorange2}73.20 & \cellcolor{lightorange1}33.43 & \cellcolor{lightorange1}35.84 & \cellcolor{lightorange3}32.50 & \cellcolor{lightorange1}49.50 & \cellcolor{lightorange2}46.47 & \cellcolor{lightorange1}48.24 \\ 
        
        - Entropy~\citep{prabhudesai2025maximizing} & \cellcolor{lightorange5}76.80 & \cellcolor{lightorange4}87.57 & \cellcolor{lightorange4}42.62 & \cellcolor{lightorange5}35.00 & \cellcolor{lightorange5}53.87 & \cellcolor{lightorange3}47.61 & \cellcolor{lightorange5}52.42 \\ 
        
        - Majority-Voting~\citep{shafayat2025can}   & \cellcolor{lightorange3}76.00 & \cellcolor{lightorange2}64.14 & \cellcolor{lightorange5}44.58 & \cellcolor{lightorange2}32.25 & \cellcolor{lightorange2}50.25 & \cellcolor{lightorange1}46.35 & \cellcolor{lightorange2}48.75 \\ 
        
        - Co-rewarding-I (Ours)                     & \cellcolor{lightorange1}72.80 & \cellcolor{lightorange3}83.93 & \cellcolor{lightorange2}39.41 & \cellcolor{lightorange1}26.54 & \cellcolor{lightorange4}53.25 & \cellcolor{lightorange4}48.11 & \cellcolor{lightorange3}50.82 \\ 
        
        - Co-rewarding-II (Ours)                    & \cellcolor{lightorange4}76.60 & \cellcolor{lightorange5}89.23 & \cellcolor{lightorange3}42.32 & \cellcolor{lightorange4}34.00 & \cellcolor{lightorange3}51.50 & \cellcolor{lightorange5}48.45 & \cellcolor{lightorange4}51.80 \\ 
        \bottomrule[1.6pt]
    \end{tabular}}
    % \vspace{-4mm}
\end{table}

\begin{figure}[t!]
  \centering
  \includegraphics[width=0.19\textwidth]{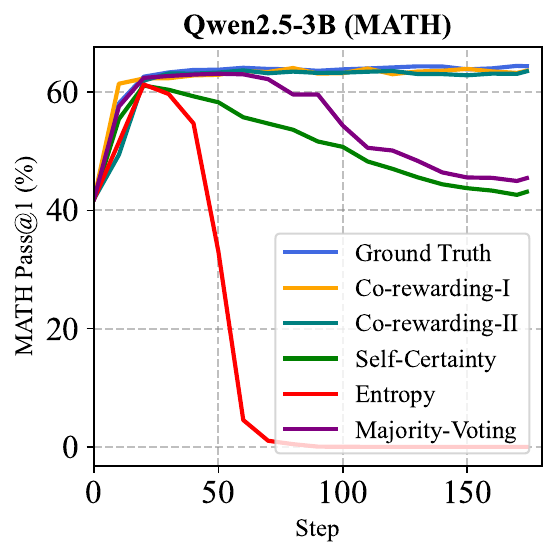}
  \includegraphics[width=0.19\textwidth]{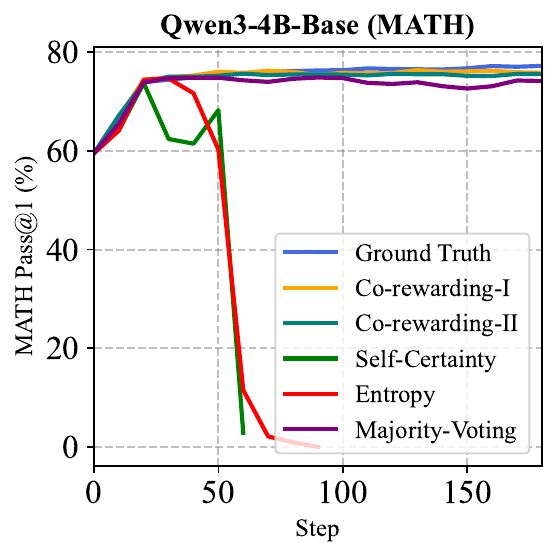}
  \includegraphics[width=0.19\textwidth]{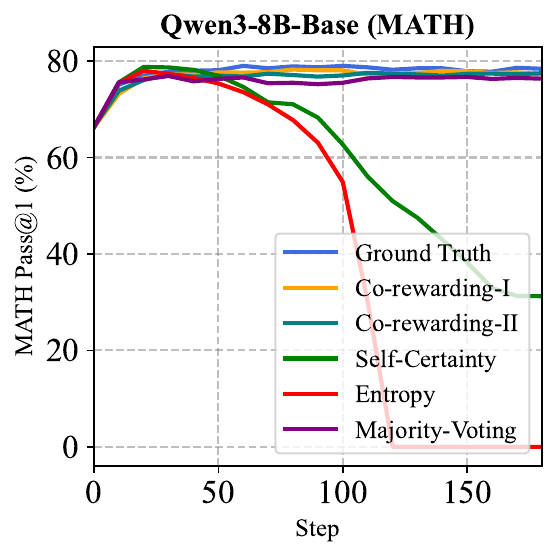}
  \includegraphics[width=0.19\textwidth]{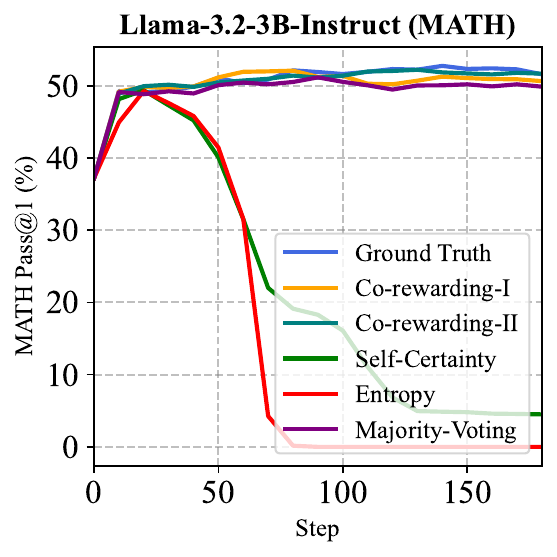}
  \includegraphics[width=0.19\textwidth]{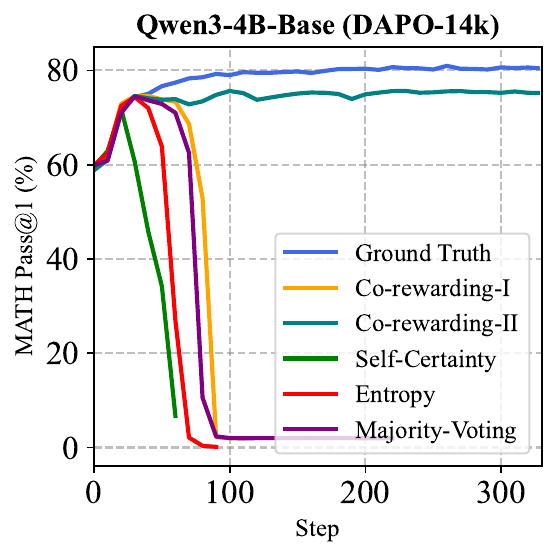}
  \caption{\textbf{Performance curves on validation set.} \textit{Left to Right:} \{Qwen2.5-3B, Qwen3-4B-Base, Qwen3-8B-Base, Llama-3.2-3B-Instruct\} trained on MATH, Qwen3-4B-Base trained on DAPO-14k. }
  \label{fig:val_curve_qwen3-4b_2.5-3b}
\end{figure}

% ==============================================

\section{Additional Experimental Results}
\label{app:add_exp_results}

\subsection{More Results on Other Training Sets and LLMs}
\label{app:more_exp_LLM_trainingset}
Table~\ref{Tble:Appe_Exp_CorewardingV1} reports additional results of Qwen2.5-3B and Qwen3-1.7B-Base and Qwen3-4B-Base trained on MATH, while Table~\ref{Tble:Supplement_Exp_CorewardingV2_OpenRS} extends the experiments of Qwen3-8B-Base and Qwen3-4B-Base to another training set OpenRS~\citep{dang2025reinforcement}. It can be observed that Co-rewarding occupies relatively darker areas. Across models and training sets, Co-rewarding-I and II achieve an average relative improvement of $+2.23\%$ on GSM8K, with notably high pass@1 scores of $92.65\%$ and $90.90\%$ for Qwen3-8B-Base trained on OpenRS, respectively. Moreover, thanks to its stability, Co-rewarding-II delivers more reliable gains than self-rewarding baselines, which occasionally suffer lower performance on certain models or benchmarks, e.g., Self-Certainty on Qwen3-1.7B-Base in Table~\ref{Tble:Appe_Exp_CorewardingV1} or Majority-Voting on Qwen3-4B-Base in Table~\ref{Tble:Supplement_Exp_CorewardingV2_OpenRS}. These results further demonstrate the effectiveness of Co-rewarding.

\begin{table}[t]
    \centering
    \caption{\textbf{Performance (\%) of test-time training (TTT).} Since self-supervised methods are label-free, they can be leveraged during inference for test-time training to further enhance performance.}
    \label{Tble:app_exp_TTT}
    \resizebox{0.99\textwidth}{!}{
    \begin{tabular}{l|l|cccccccc}
    \toprule[1.6pt]
        \multirow{2}{*}{\textbf{LLMs}} & \multirow{2}{*}{\textbf{Methods}} & \multicolumn{8}{c}{\textbf{AMC}} \\
        \cmidrule{3-10}
        ~ & ~ & avg@8 & pass@8 & avg@16 & pass@16 & avg@32 & pass@32 & avg@64 & pass@64 \\ 
    \midrule
        \multirow{6}{*}{\textit{\textbf{Qwen2.5-7B}}} 
          & Before-TTT      & 15.81 & 46.99 & 17.55 & 66.27 & 16.34 & 74.70 & 17.32 & 75.90 \\
        \cmidrule{2-10}
        ~ & Self-Certainty  & 41.57 & 74.70 & 39.23 & 74.70 & 39.68 & 78.31 & 39.95 & 87.95 \\
        ~ & Entropy         & 38.70 & 56.63 & 39.76 & 68.67 & 39.57 & 79.52 & 39.34 & 81.93 \\
        ~ & Majority-Voting & 43.67 & 63.86 & 43.67 & 67.47 & 43.49 & 78.31 & 44.35 & 85.54 \\
        ~ & Co-rewarding-I  & 44.88 & 60.24 & 45.33 & 60.24 & 45.44 & 71.08 & 45.76 & 73.49 \\
        ~ & Co-rewarding-II & 43.22 & 69.88 & 41.34 & 75.90 & 40.36 & 78.31 & 41.64 & 87.95 \\
        \specialrule{1pt}{0.4ex}{0.4ex}
        \multirow{6}{*}{\textit{\textbf{Qwen3-8B-Base}}} 
          & Before-TTT      & 20.93 & 61.45 & 21.31 & 73.49 & 19.58 & 79.52 & 20.97 & 86.75 \\
        \cmidrule{2-10}
        ~ & Self-Certainty  & 49.85 & 78.31 & 50.68 & 78.31 & 50.41 & 84.34 & 49.55 & 89.16 \\
        ~ & Entropy         & 48.64 & 74.70 & 49.92 & 80.72 & 49.96 & 87.95 & 50.23 & 89.16 \\
        ~ & Majority-Voting & 50.90 & 73.49 & 50.00 & 72.29 & 50.60 & 80.72 & 51.36 & 85.54 \\
        ~ & Co-rewarding-I  & 52.86 & 68.67 & 53.46 & 74.70 & 53.24 & 81.93 & 53.58 & 84.34 \\
        ~ & Co-rewarding-II & 48.64 & 72.29 & 48.19 & 73.49 & 50.19 & 83.13 & 49.28 & 91.57 \\
    \bottomrule[1.6pt]
    \end{tabular}}
\end{table}
% ============================================================================

% \subsection{Results of Test-time Training (TTT)}
% Self-supervised methods, which do not require GT labels, are naturally compatible with test-time training (TTT). This allows models to further refine themselves during inference, which is an advantage compared to GT-Reward training. We conduct test-time training experiments on the challenging competition-level dataset AMC. Table~\ref{Tble:app_exp_TTT} summarizes the experimental results. It can be observed that our Co-rewarding achieves consistently strong performance across all baselines. In particular, on Qwen3-8B-Base, Co-rewarding outperforms all baselines by a clear margin for all values of $k$. These results demonstrate the effectiveness of Co-rewarding in enhancing the reasoning ability of LLMs during inference, especially when applied to more powerful models such as Qwen3-8B-Base. Moreover, the higher avg@$k$ scores reflects that Co-rewarding enhances the robustness and stability of the LLM in generating correct answers across multiple rollouts.

\begin{figure}[t!]
  \centering
  \includegraphics[width=0.31\textwidth]{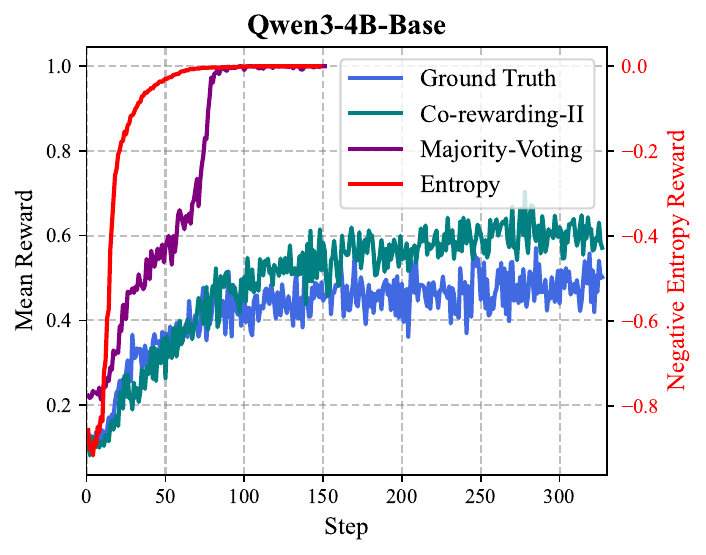}
  \includegraphics[width=0.31\textwidth]{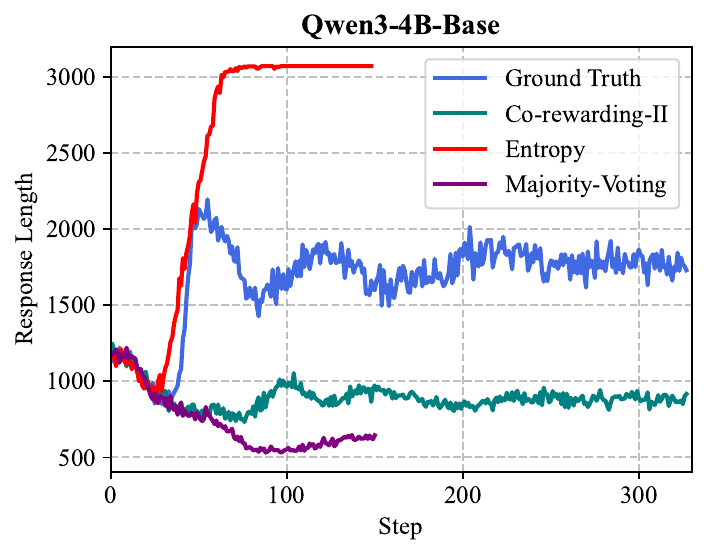}
  \includegraphics[width=0.31\textwidth]{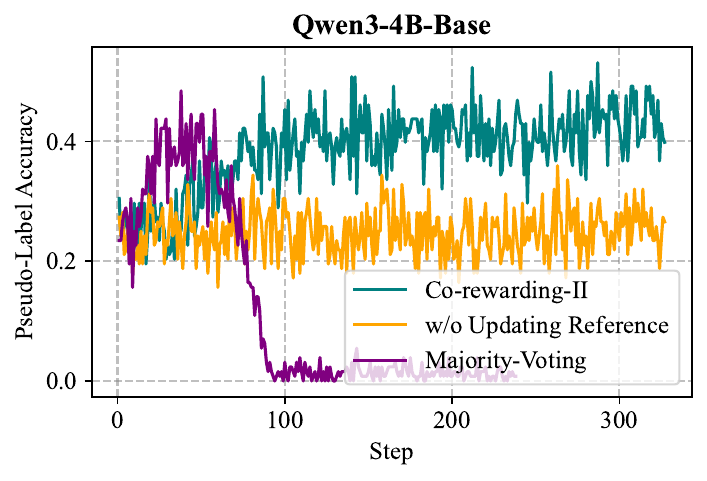}
  \caption{\textbf{Curves of reward (Left), response length (Middle), and pseudo label accuracy (Right) of Qwen3-4B-Base} trained on DAPO-14k. Entropy reward is plotted on the right $y$-axis due to its different reward scale. Note that entropy minimization is to maximizing the negative entropy.}
  \label{fig:reward_length_qwen3-4b}
\end{figure}

\begin{table}[t!]
    \centering
    \caption{\textbf{Detailed MMLU-Pro performance on Qwen3-4B-Base and Llama-3.2-3B-Instruct traind on DAPO-14k.} Results are reported for each of the 14 categories in MMLU-Pro.}
    \label{Tble:MMLU-Pro_LlamaQwen4B_DAPO14k}
    \vspace{-2mm}
    \resizebox{0.95\textwidth}{!}{
    \begin{tabular}{l|ccccccc}
    \toprule[1.6pt]
        \multicolumn{8}{c}{\textit{\textbf{Qwen3-4B-Base}}} \\
    \midrule
        \textbf{Methods} & \textbf{biology} & \textbf{business} & \textbf{chemistry} & \textbf{computer sci.} & \textbf{economics} & \textbf{engineering} & \textbf{health} \\
        \midrule
        - GT-Reward        & 73.50 & 63.49 & 59.71 & 56.34 & 65.05 & 42.93 & 50.86 \\
        \midrule
        - Self-Certainty   & 71.41 & 54.37 & 45.93 & 50.73 & 63.27 & 35.91 & 50.12 \\
        - Entropy          & 70.99 & 56.02 & 50.44 & 48.29 & 63.15 & 34.37 & 48.41 \\
        - Majority-Voting  & 70.43 & 55.77 & 52.83 & 53.41 & 62.79 & 38.09 & 50.61 \\
        - Co-rewarding-I   & 73.92 & 59.82 & 50.71 & 54.15 & 64.93 & 41.49 & 49.76 \\
        - Co-rewarding-II  & 72.66 & 59.95 & 55.65 & 53.41 & 64.10 & 39.73 & 50.61 \\
        \specialrule{1pt}{0.4ex}{0.4ex}
        \textbf{Methods} & \textbf{history} & \textbf{law} & \textbf{math} & \textbf{other} & \textbf{philosophy} & \textbf{physics} & \textbf{psychology} \\
        \midrule
        - GT-Reward        & 44.88 & 26.34 & 69.80 & 48.81 & 44.69 & 57.04 & 65.79 \\
        \midrule
        - Self-Certainty   & 39.63 & 24.43 & 59.44 & 43.94 & 40.08 & 47.04 & 59.65 \\
        - Entropy          & 40.68 & 26.43 & 60.99 & 45.13 & 43.69 & 50.89 & 61.90 \\
        - Majority-Voting  & 40.94 & 23.43 & 64.17 & 43.39 & 44.09 & 50.73 & 63.66 \\
        - Co-rewarding-I   & 40.94 & 23.25 & 63.73 & 44.91 & 42.69 & 50.58 & 60.78 \\
        - Co-rewarding-II  & 42.26 & 24.79 & 67.58 & 44.59 & 41.88 & 54.19 & 62.91 \\
    \specialrule{1pt}{0.4ex}{0.4ex}
    
        \multicolumn{8}{c}{\textit{\textbf{Llama3.2-3B-Instruct}}} \\
    \midrule
        \textbf{Methods} & \textbf{biology} & \textbf{business} & \textbf{chemistry} & \textbf{computer sci.} & \textbf{economics} & \textbf{engineering} & \textbf{health} \\
        \midrule
        - GT-Reward        & 54.81 & 36.25 & 25.18 & 33.41 & 42.65 & 21.57 & 39.36 \\
        \midrule
        - Self-Certainty   & 55.23 & 32.95 & 27.21 & 31.95 & 42.77 & 20.54 & 39.12 \\
        - Entropy          & 52.86 & 31.05 & 23.94 & 32.93 & 41.71 & 20.43 & 38.02 \\
        - Majority-Voting  & 56.07 & 32.95 & 22.79 & 30.98 & 44.19 & 18.99 & 39.61 \\
        - Co-rewarding-I   & 51.88 & 34.22 & 22.88 & 34.88 & 44.67 & 19.09 & 38.63 \\
        - Co-rewarding-II  & 56.21 & 34.35 & 27.03 & 35.61 & 43.01 & 19.92 & 40.34 \\
        \specialrule{1pt}{0.4ex}{0.4ex}
        \textbf{Methods} & \textbf{history} & \textbf{law} & \textbf{math} & \textbf{other} & \textbf{philosophy} & \textbf{physics} & \textbf{psychology} \\
        \midrule
        - GT-Reward        & 30.18 & 22.71 & 34.20 & 34.74 & 32.06 & 28.33 & 50.38 \\
        \midrule
        - Self-Certainty   & 30.45 & 24.98 & 33.38 & 31.60 & 29.86 & 28.56 & 50.50 \\
        - Entropy          & 33.86 & 21.89 & 32.35 & 33.01 & 29.46 & 24.25 & 47.50 \\
        - Majority-Voting  & 32.02 & 25.25 & 34.35 & 34.20 & 29.86 & 24.79 & 48.25 \\
        - Co-rewarding-I   & 33.86 & 23.25 & 32.12 & 33.01 & 31.86 & 25.40 & 48.75 \\
        - Co-rewarding-II  & 32.28 & 24.34 & 35.83 & 36.26 & 33.27 & 28.18 & 49.12 \\

    \specialrule{1pt}{0.4ex}{0.4ex}
    \end{tabular}}
\end{table}

\subsection{More Curves of Reward, Response Length and Pseuo Label Accuracy}
\label{app:reward_response_length}
Figure~\ref{fig:reward_length_qwen3-4b} supplements the reward and response curves of Qwen3-4B-Base trained on DAPO-14k. The trends are consistent with Qwen3-8B-Base and Llama-3.2-3B-Instruct in Figure~\ref{fig:reward_length_qwen3-8b_llama3b}: Majority-Voting and Entropy rapidly increase rewards at early stage and quickly peak, a clear sign of reward hacking. In contrast, GT-Reward and Co-rewarding-II exhibit smoother, steadily rising rewards, indicating genuine learning of reasoning ability. Moreover, Co-rewarding-II maintains moderate response lengths on Qwen3-4B-Base, further demonstrating its generality in balancing the exploration–exploitation trade-off during reasoning training, which is a core principle of RL~\citep{wang2018exploration}.

\par
Additionally, the right panel of Figure~\ref{fig:reward_length_qwen3-4b} presents the pseudo-label accuracy of Qwen3-4B-Base, showing trends consistent with Qwen3-8B-Base and Llama-3.2-3B-Instruct in Figure~\ref{fig:pseudo_acc_corewardv2}. As training progresses, Co-rewarding-II steadily improves pseudo-label accuracy, while ``w/o Updating Reference'' remains around 25\%. Majority-Voting briefly increases accuracy but soon collapses to zero, clearly indicating reward hacking. This highlights our design philosophy of pairing a fast policy student with a slowly updated teacher, which decouples supervision from the online policy while enabling the teacher to co-evolve with the student, thereby sustaining improvements in pseudo-label quality.

\subsection{More Results of Validation Performance Curves}
\label{Appe:Validation_curve}
As a supplement to Figure~\ref{fig:curve_validation}, Figure~\ref{fig:val_curve_qwen3-4b_2.5-3b} presents validation performance curves for {Qwen2.5-3B, Qwen3-4B-Base, Qwen3-8B-Base, Llama-3.2-3B-Instruct} trained on MATH, as well as Qwen3-4B-Base trained on DAPO-14k. Self-Certainty and Entropy collapse rapidly across all settings, as their supervision signals are tied to internal confidence or entropy and are easily exploited. Majority-Voting also collapses in several cases, reflecting that sampling pseudo labels from outputs cannot prevent hacking. By contrast, Co-rewarding-I maintains stability across MATH-trained models through data-side contrastive agreement, while Co-rewarding-II consistently provides stability across all models and datasets by disentangling supervision with a slowly updated teacher, making hacking substantially harder and optimization more reliable.

\subsection{Results of Test-time Training (TTT)}
Thanks to the label-free nature of self-supervised methods, which do not require GT labels, they are naturally compatible with test-time training (TTT), enabling further refinement of the model during inference. Table~\ref{Tble:app_exp_TTT} reports the TTT results on the challenging competition-level benchmark AMC across Co-rewarding and other self-rewarding baselines. We observe that Co-rewarding matches or even surpasses existing methods, achieving the best results on 11 out of 18 metrics. These findings broaden the applicability of self-supervised RL: beyond post-training for reasoning elicitation, it can also be leveraged at inference time to further improve performance on specific benchmarks.

\begin{figure}[t!]
  \centering
  \includegraphics[width=0.23\textwidth]{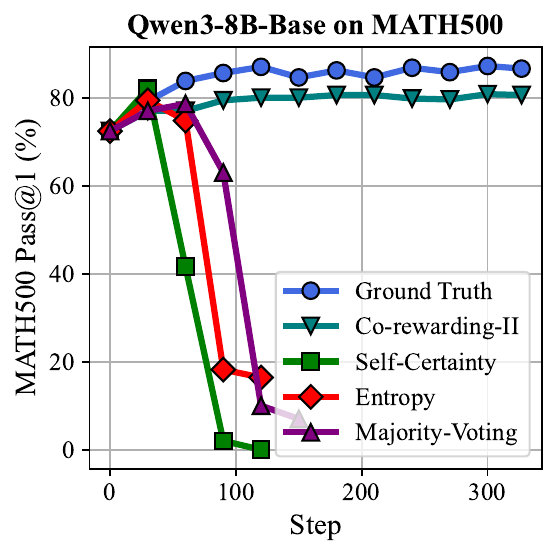}
  \includegraphics[width=0.23\textwidth]{fig/bench_curve/gsm8k_qwen38_markers.pdf}
  \includegraphics[width=0.23\textwidth]{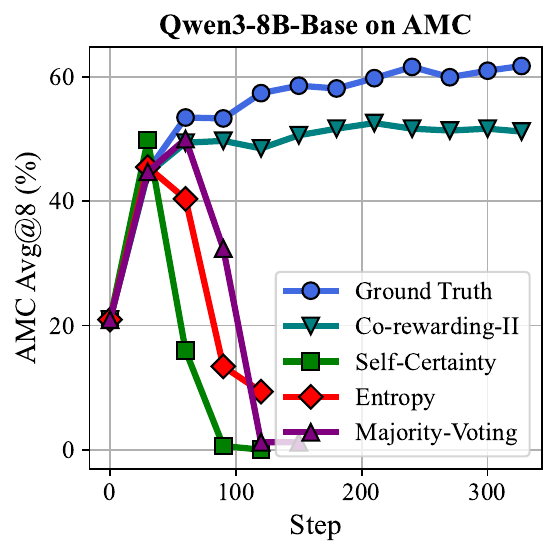} 
  \includegraphics[width=0.23\textwidth]{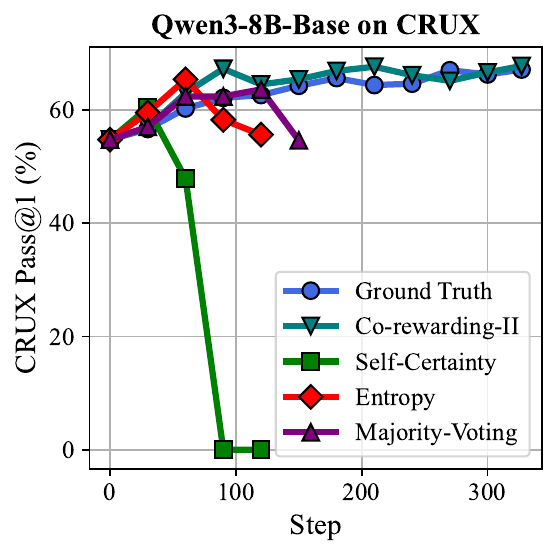} \\
  \includegraphics[width=0.23\textwidth]{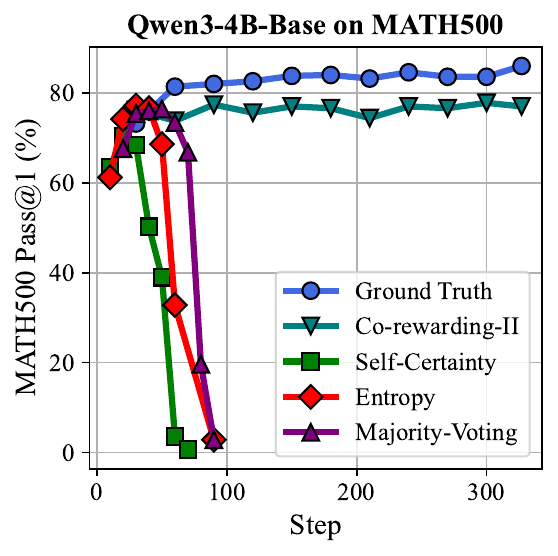}
  \includegraphics[width=0.23\textwidth]{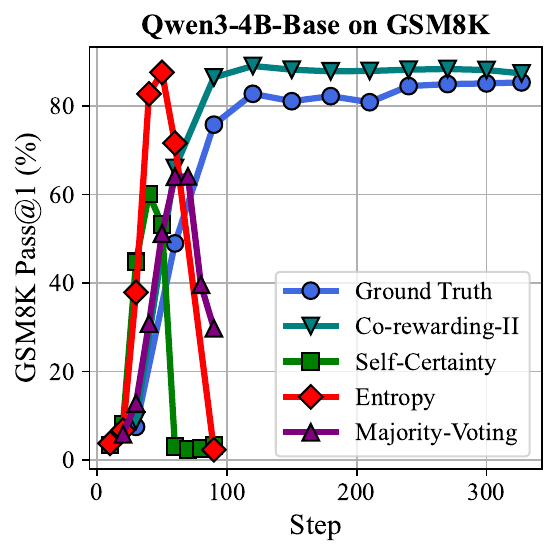} 
  \includegraphics[width=0.23\textwidth]{fig/bench_curve/AMC_qwen34_markers.pdf}
  \includegraphics[width=0.23\textwidth]{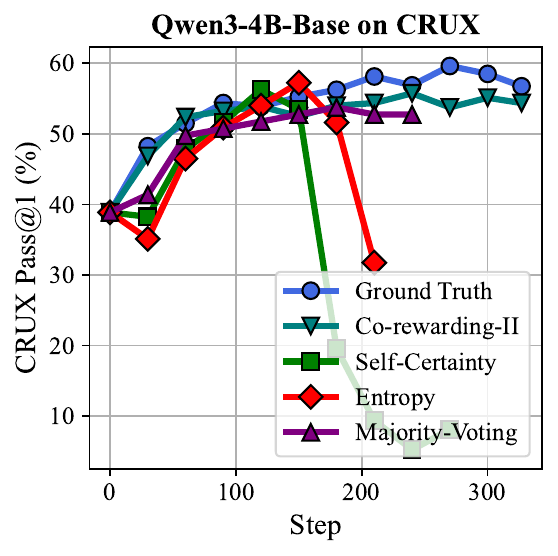} \\
  \includegraphics[width=0.23\textwidth]{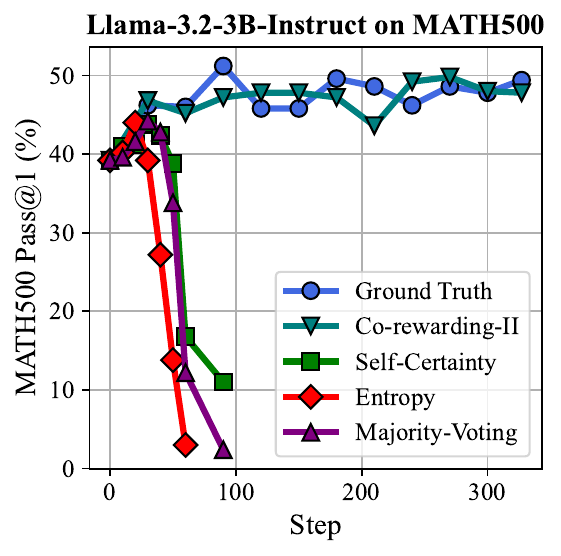}
  \includegraphics[width=0.23\textwidth]{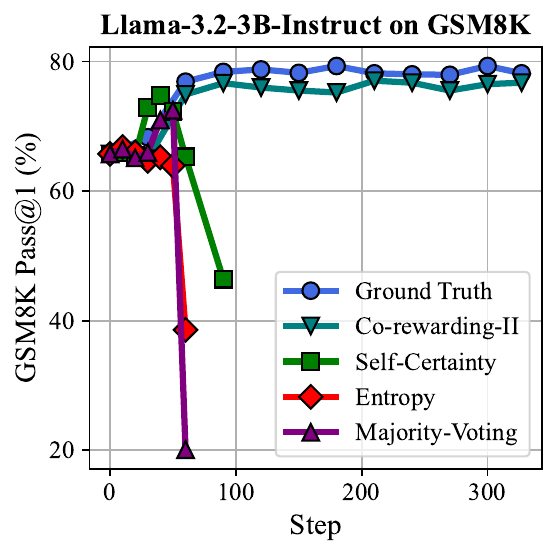}
  \includegraphics[width=0.23\textwidth]{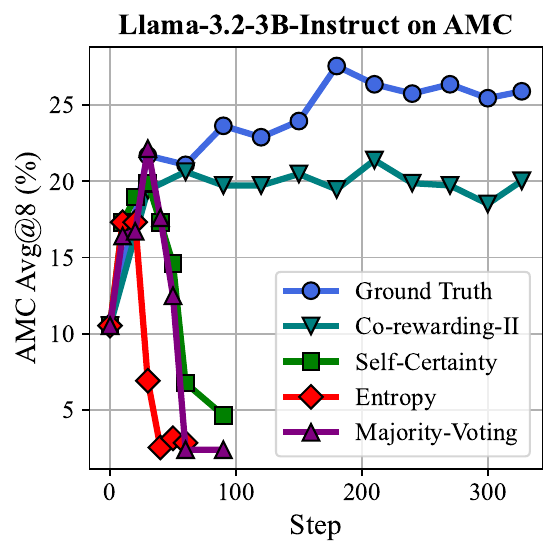}
  \includegraphics[width=0.23\textwidth]{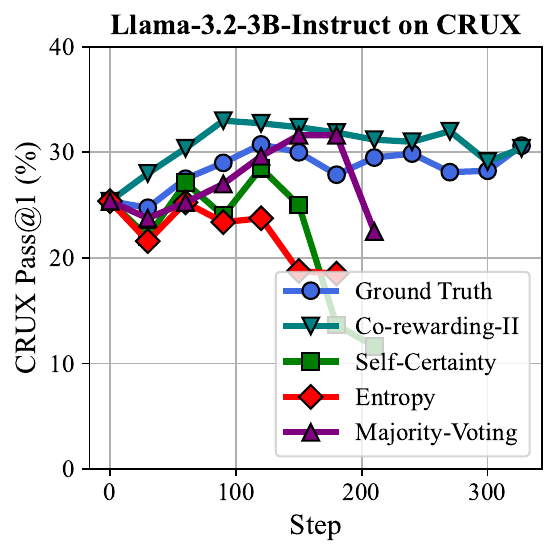}
  \caption{\textbf{Performance curves on benchmarks of MATH500, GSM8K, AMC and CRUX} across Qwen3-8B-Base, Qwen3-4B-Base, and Llama-3.2-3B-Instruct trained on DAPO-14k.}
  \label{fig:curve_math500_gsm8k_amc_crux}
\end{figure}

\begin{table}[t!]
    \centering
    \caption{\textbf{Impact of math training collapse on code and multi-task performance.} Results are evaluated on models before and after training collapse.}
    \label{Tble:Result_before_after_collapse}
    \vspace{-2mm}
    \resizebox{0.99\textwidth}{!}{
    \begin{tabular}{l|l|ccccccc}
    \toprule[1.6pt]
        \multicolumn{9}{c}{\textit{\textbf{Qwen3-4B-Base}}} \\
        \midrule
        \multirow{2}{*}{\textbf{Training stage}} & \multirow{2}{*}{\textbf{Methods}} & \multicolumn{4}{c}{\textbf{Mathematics}} & \multicolumn{2}{c}{\textbf{Code}} & \textbf{Multi-task} \\
        \cmidrule{3-9}
        ~ & ~ & MATH500 & GSM8K & AMC & AIME24 & LiveCode & CRUX & MMLU-Pro \\
        \midrule
        \multirow{3}{*}{Before training collapse} & - Self-Certainty & 68.4 & 44.81 & 35.39 & 8.85 & 25.88 & 50.12 & 48.84 \\
        ~ & - Entropy & 76.6 & 82.79 & 43.37 & 12.81 & 26.35 & 50.75 & 50.22 \\
        ~ & - Majority-Voting & 73.4 & 64.06 & 40.81 & 9.17 & 26.16 & 53.00 & 51.06 \\
        \midrule
        \multirow{3}{*}{After training collapse} & - Self-Certainty & 2.8 & 3.34 & 2.71 & 0.00 & 14.22 & 8.12 & 29.71 \\
        ~ & - Entropy & 2.8 & 2.35 & 3.46 & 0.00 & 18.60 & 31.75 & 28.13 \\
        ~ & - Majority-Voting & 2.8 & 4.85 & 1.36 & 0.00 & 24.36 & 52.75 & 50.19 \\
    \specialrule{1.6pt}{0.4ex}{0.4ex}
    \end{tabular}}
\end{table}

\subsection{More Results of Benchmark Performance Curves}
As a supplement to Figure~\ref{fig:curve_validation} and Figure~\ref{fig:performance_stability}, Figure~\ref{fig:curve_math500_gsm8k_amc_crux} presents performance curves on MATH500, GSM8K, AMC, and CRUX with Qwen3-8B-Base, Qwen3-4B-Base, and Llama-3.2-3B-Instruct. Consistent with earlier findings, Self-Certainty, Entropy, and Majority-Voting rapidly collapse across benchmarks and models, while Co-rewarding-II and GT-Reward sustain continued and stable improvements. These results underscore the link between performance and training stability: stable training enables models to continue improving by effectively learning knowledge from more data.

\subsection{Impact of Math Training Collapse on Other Tasks}
We investigate how training collapse occurring on math-oriented training sets impacts the model's performance on code-generation and multi-task benchmarks. To this end, we evaluate models trained with existing self-rewarding methods (Self-Certainty, Entropy, and Majority-Voting) both before and after training collapse. Table~\ref{Tble:Result_before_after_collapse} summarizes the results. We observe that training collapses on math-related training sets affect other tasks (LiveCode, CRUX and MMLU-Pro) in different way for certainty- or entropy-based methods (Self-Certainty and Entropy) compared with consensus-based methods (Majority-Voting). When collapse occurs on math-oriented training sets, all three methods show substantial performance degradation on the four math benchmarks (MATH500, GSM8K, AMC, and AIME24). However, their impacts on other tasks differ:

For certainty- or entropy-based methods, the performance on LiveCode, CRUX, and MMLU-Pro also declines after collapse on math training sets. This arises from their reward objectives: maximizing self-certainty or minimizing entropy, result in the decoding probability mass becoming highly concentrated on a very subset of tokens. Consequently, the model produces repetitive outputs, and this repetitive decoding behavior transfers across tasks, leading to degraded performance beyond the math domain.

For the consensus-based method, Majority-Voting shows similar performance before and after training collapse on math-oriented training sets. This may be because its collapses stem from reward hacking at the answer format: the model exploits the \mbox{\textbackslash boxed\{\}} structure by consistently inserting an incorrect but self-consistent answer to maximize reward. This type of collapse weakly affects the intermediate reasoning trace, which largely remains structured. Since code-generation and multi-task benchmarks do not rely on boxed-answer extraction, this type of collapse has limited impact on their performance.

\begin{table}[t!]
\caption{\textbf{Difference between original and rephrased questions} from background richness, vocabulary complexity, and sentence complexity.}
\vspace{-2mm}
\begin{center}
\resizebox{0.99\textwidth}{!}{
    \begin{tabular}{l|cccc}
    \toprule[1.5pt]
    \textbf{Training Set} & \textbf{\# Data Size} & \textbf{Background richness} & \textbf{Vocabulary complexity} & \textbf{Sentence complexity} \\
    \midrule
    MATH & 7,500 & +4.91\% & +4.79\% & +9.05\% \\
    DAPO-14k & 14,100 & +4.65\% & 1.95\% & +4.19\% \\
    \bottomrule[1.5pt]
    \end{tabular}}
\label{Tble:background_vocabulary_sentence}
\end{center}
\end{table}

\begin{table}[t!]
\caption{\textbf{Success rate of different rephraser LLMs:} MATH training set rephrased by Qwen3-32B, Qwen3-8B, and Qwen3-1.7B, respectively.}
\begin{center}
\vspace{-2mm}
\resizebox{0.99\textwidth}{!}{
    \begin{tabular}{l|cccc}
    \toprule[1.5pt]
    \textbf{Rephraser LLM} & \textbf{Training Set} & \textbf{\# Original questions} & \textbf{\# Rephrased questions} & \textbf{Success rate (\%)} \\
    \midrule
    Qwen3-32B & MATH & 7,500 & 7,498 & 99.97\% \\
    Qwen3-8B & MATH & 7,500 & 7,477 & 99.69\% \\
    Qwen3-1.7B & MATH & 7,500 & 2,060 & 27.47\% \\
    \bottomrule[1.5pt]
    \end{tabular}}
\label{Tble:different_rephraser}
\end{center}
\end{table}

\begin{table}[t!]
    \centering
    \caption{\textbf{Impact of rephraser LLM for Co-rewarding-I.} Train Qwen3-8B-Base using data rephrased by Qwen3-32B, Qwen3-8B and Qwen3-1.7B, respectively.}
    \label{Tble:Result_different_rephraser}
    \vspace{-2mm}
    \resizebox{0.99\textwidth}{!}{
    \begin{tabular}{l|l|cccccccc}
    \toprule[1.6pt]
        \textbf{Trained Model} & \textbf{Rephraser LLM} &\textbf{MATH500} & \textbf{GSM8K} & \textbf{AMC} & \textbf{AIME24} & \textbf{LiveCode} & \textbf{CRUX} & \textbf{IFEval} & \textbf{MMLU-Pro} \\
        \midrule
        \multirow{3}{*}{Qwen3-8B-Base} & Qwen3-32B & 81.2 & 93.70 & 51.20 & 15.10 & 30.81 & 66.00 & 55.79 & 59.95 \\
        ~ & Qwen3-8B & 79.2 & 92.72 & 51.51 & 14.58 & 30.90 & 63.12 & 54.73 & 59.30 \\
        ~ & Qwen3-1.7B & 78.2 & 87.41 & 49.25 & 12.81 & 29.57 & 61.00 & 53.44 & 55.85 \\
    \specialrule{1.6pt}{0.4ex}{0.4ex}
    \end{tabular}}
\end{table}

\subsection{Discussion of MATH and DAPO-14k}
\label{app:discuss_MATH_DAPO14k}
We leverage Qwen3-235B-A22B to score the difference between original and rephrased questions from multiple perspectives, including background richness, vocabulary complexity, and sentence complexity, for MATH and DAPO-14k. From Table~\ref{Tble:background_vocabulary_sentence}, we observe that the rephrasing in MATH exhibits larger changes from the original to rephrased questions than DAPO-14k. This suggests that the questions in MATH may provide favorable conditions for promoting diverse rephrasing variability, which is beneficial for the effectiveness of contrastive agreement in Co-rewarding-I.

\subsection{Robustness Analysis of Different Rephraser LLMs}
To analyze the impact of different rephraser LLMs for Co-rewarding-I, we conduct additional experiments using smaller LLMs instead of Qwen3-32B for rephrasing. To control architectural variability in the rephraser LLMs, we employ two smaller LLMs from the same family, i.e., Qwen3-8B and Qwen3-1.7B, for rephrasing the MATH training set. Table~\ref{Tble:different_rephraser} reports the rephrasing success rate. We observe that rephrasing success rates drop as the model size decreases, which is expected: rephrasing math questions while preserving the analogical essence is a relatively challenging task, and weaker LLMs struggle to achieve this goal. This observation supports our choice of Qwen3-32B as the rephraser, as a sufficiently capable LLM is required to produce faithful rephrasing.

We then train Co-rewarding-I on Qwen3-8B-Base using rephrased data generated by Qwen3-32B, Qwen3-8B, and Qwen3-1.7B, respectively. The performance is summarized in Table~\ref{Tble:Result_different_rephraser}. From the results, it can be observed that performance gradually degrades as the size of the rephraser LLM decreases, but not always significantly. Rephrasing with Qwen3-8B maintains reasonably similar performance to using Qwen3-32B, indicating that Co-rewarding-I exhibits a certain degree of robustness under moderate reductions in rephrasing quality. Notably, rephrasing with Qwen3-1.7B leads to a substantial performance drop. This degradation is largely attributable to the significantly lower rephrasing success rate of Qwen3-1.7B, which results in a substantial reduction of usable training data and consequently weakens the effectiveness of Co-rewarding-I.

\begin{figure}[t]
  \centering
  \includegraphics[width=0.31\textwidth]{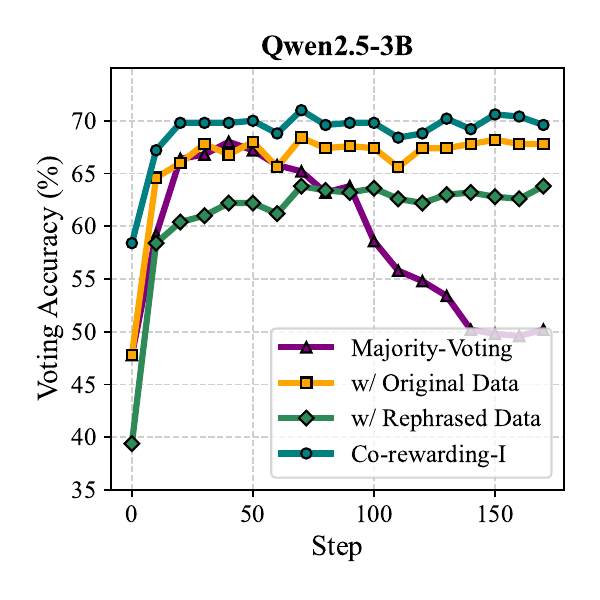}
  \includegraphics[width=0.31\textwidth]{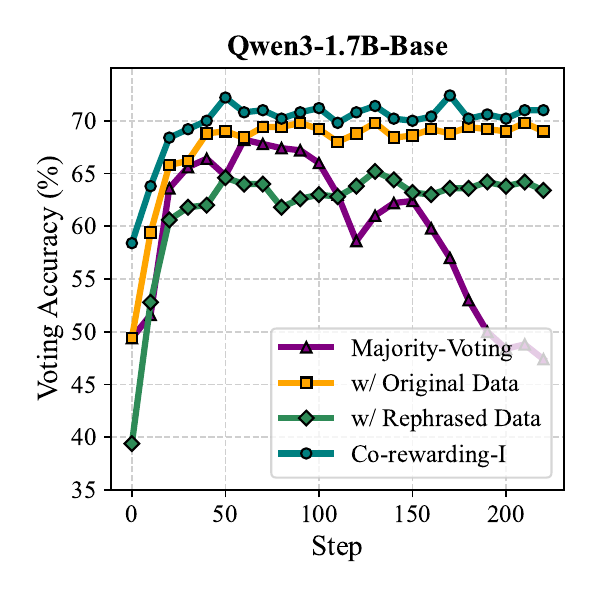}
  \includegraphics[width=0.31\textwidth]{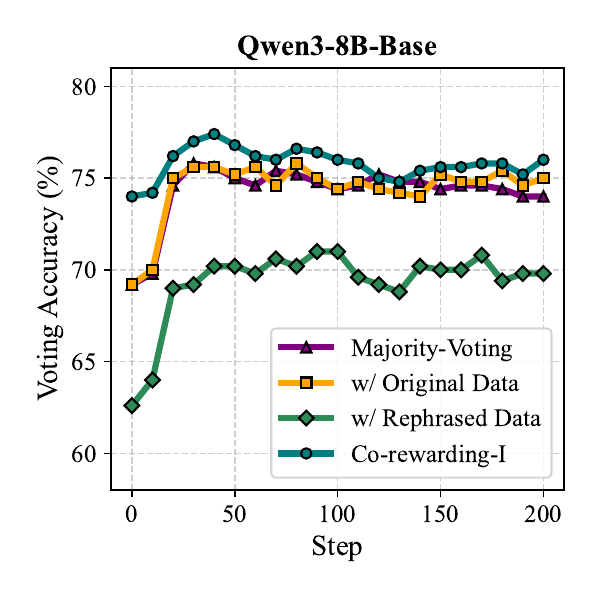}
  \caption{\textbf{Curves of voting accuracy of Majority-Voting, Co-rewarding-I and its ablations} with Qwen2.5-3B, Qwen3-1.7B-Base and Qwen3-8B-Base trained on MATH.}
  \label{fig:voting_acc_qwen2.5-3b_qwen3-1.7b-8b}
\end{figure}

\begin{figure}[t!]
  \centering
  \includegraphics[width=0.23\textwidth]{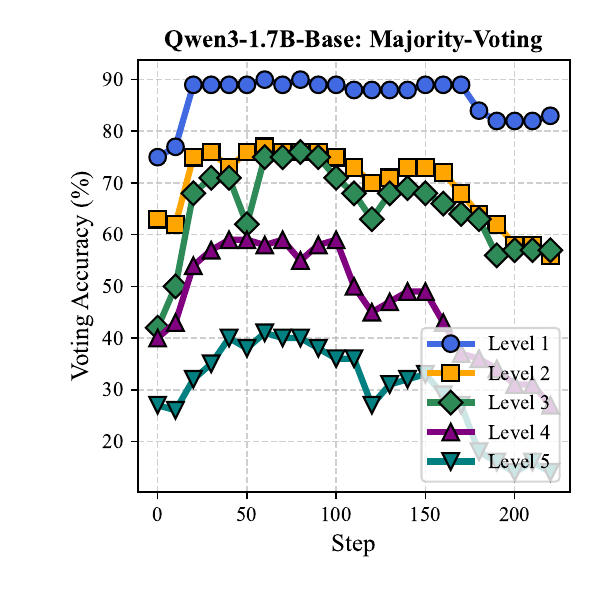}
  \includegraphics[width=0.23\textwidth]{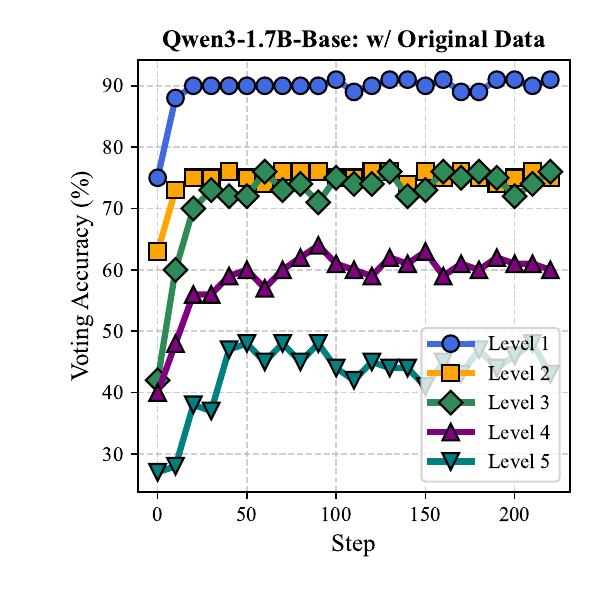}
  \includegraphics[width=0.23\textwidth]{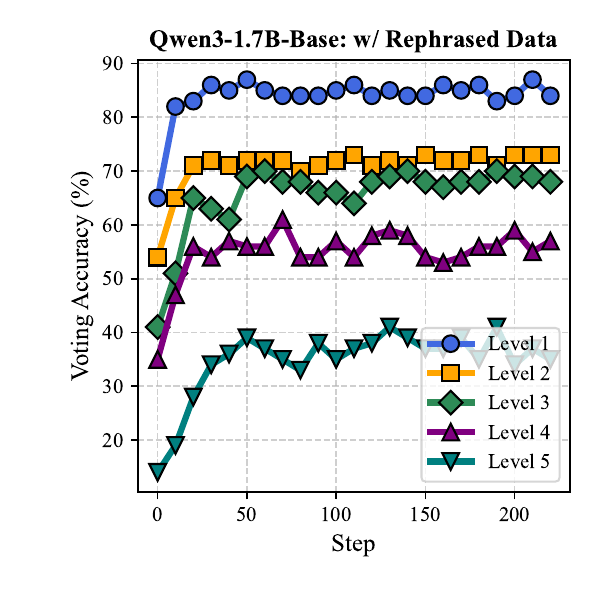}
  \includegraphics[width=0.23\textwidth]{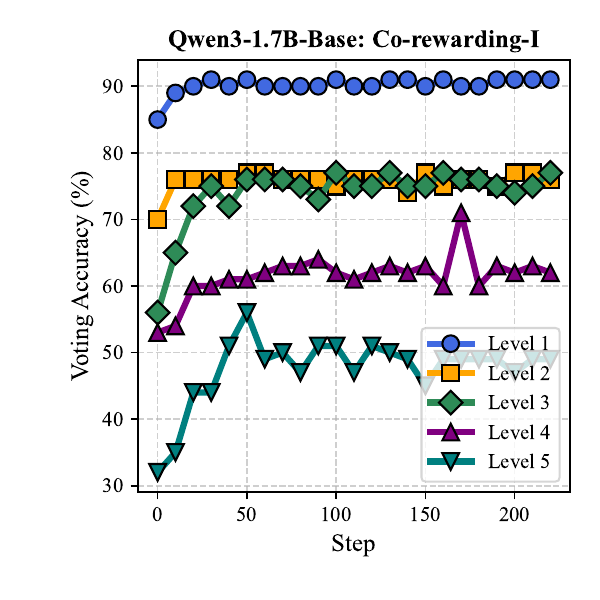} \\
  \includegraphics[width=0.23\textwidth]{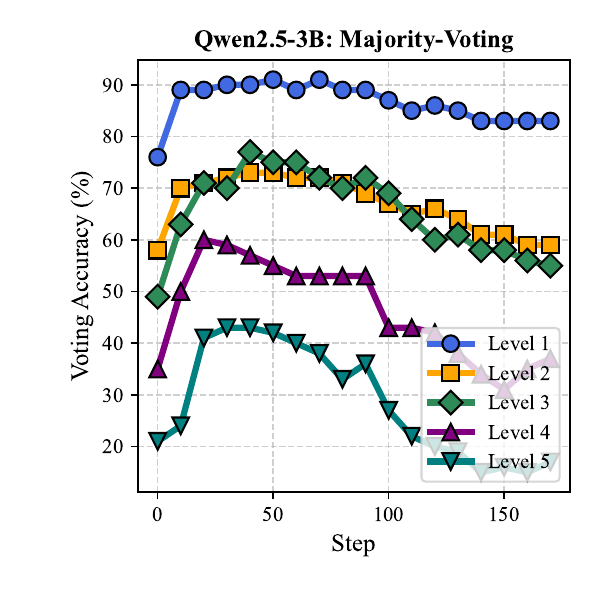}
  \includegraphics[width=0.23\textwidth]{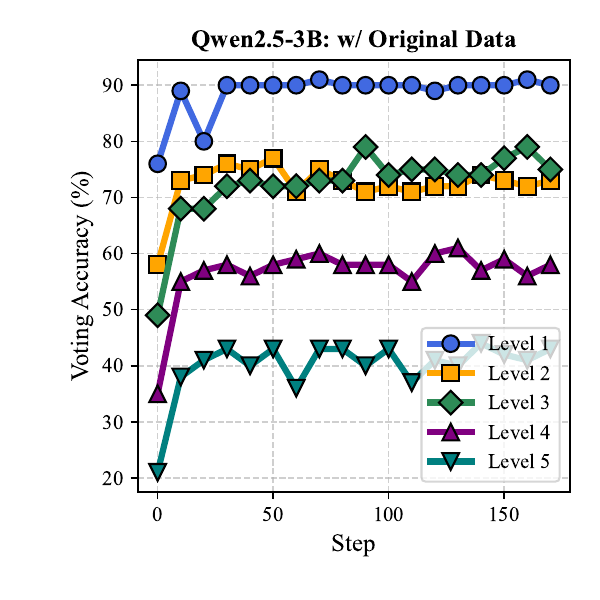}
  \includegraphics[width=0.23\textwidth]{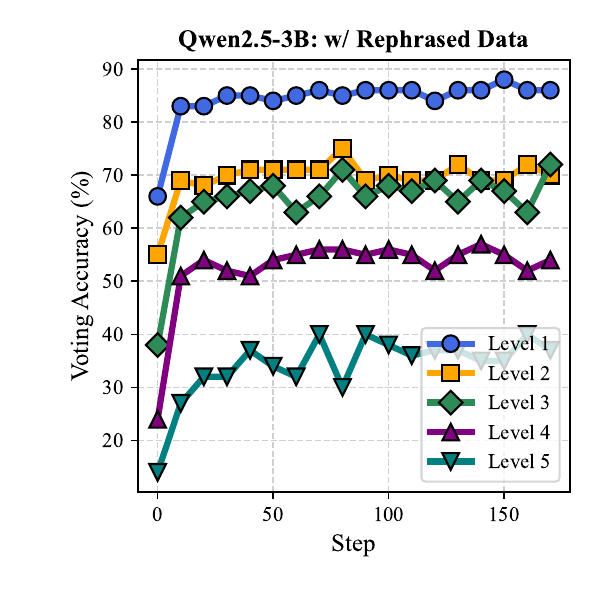}
  \includegraphics[width=0.23\textwidth]{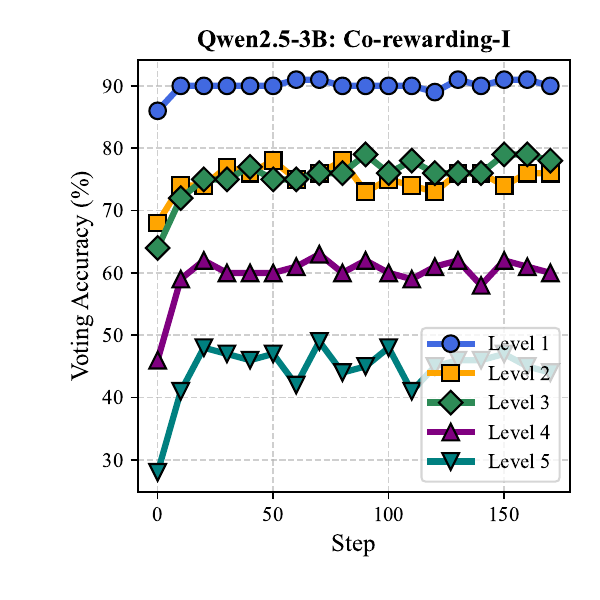} \\
  \caption{\textbf{Voting accuracy of Majority-Voting, Co-rewarding-I and its ablated variants across different difficulty levels of questions} \textit{Top:} Qwen3-1.7B-Base. \textit{Bottom:} Qwen2.5-3B.}
  \label{fig:voting_level}
\end{figure}

\subsection{Voting Accuracy Analysis of Co-rewarding-I}

To demonstrate the stability and efficiency of Co-rewarding-I, we compare its voting accuracy against that of Majority-Voting in Figure~\ref{fig:voting_acc_qwen2.5-3b_qwen3-1.7b-8b} and Figure~\ref{fig:voting_level}. These experiments are conducted on Qwen2.5-3B, Qwen3-1.7B-Base and Qwen3-8B-Base models, all trained on the MATH dataset. Across all settings, the Majority-Voting method exhibits reward hacking, where its performance sharply declines after reaching an early peak, particularly on more difficult questions (levels 2 to 5). In contrast, Co-Rewarding-I maintains a stable voting accuracy on both original and rephrased data. Ultimately, it achieves the highest overall voting accuracy across all models and dataset configurations.

%  ----------------------------------------
\begin{table}[tbh]
    \centering
    \caption{\textbf{Detailed MMLU-Pro performance on Qwen3-8B-Base and Qwen3-4B-Base trained on OpenRS.} Results are reported for each of 14 categories in MMLU-Pro.}
    \label{Tble:MMLU-Pro_Qwen8BQwen4B_OpenRS}
    \vspace{-2mm}
    \resizebox{0.95\textwidth}{!}{
    \begin{tabular}{l|ccccccc}
    \toprule[1.6pt]
        \multicolumn{8}{c}{\textit{\textbf{Qwen3-4B-Base}}} \\
    \midrule
        \textbf{Methods} & \textbf{biology} & \textbf{business} & \textbf{chemistry} & \textbf{computer sci.} & \textbf{economics} & \textbf{engineering} & \textbf{health} \\
        \midrule
        - GT-Reward & 70.99 & 59.82 & 52.30 & 54.63 & 65.05 & 39.01 & 51.22 \\
        \midrule
        - Self-Certainty & 69.87 & 54.50 & 44.08 & 49.27 & 63.63 & 37.36 & 50.24 \\
        - Entropy & 70.71 & 58.68 & 49.03 & 51.22 & 63.39 & 37.46 & 49.63 \\
        - Majority-Voting & 69.60 & 55.77 & 47.17 & 53.17 & 63.39 & 36.02 & 48.78 \\
        - Co-rewarding-I & 69.04 & 55.39 & 47.79 & 53.41 & 63.86 & 38.39 & 50.61 \\
        - Co-rewarding-II & 70.85 & 58.81 & 53.27 & 53.90 & 66.11 & 37.15 & 52.81 \\
        \specialrule{1pt}{0.4ex}{0.4ex}
        \textbf{Methods} & \textbf{history} & \textbf{law} & \textbf{math} & \textbf{other} & \textbf{philosophy} & \textbf{physics} & \textbf{psychology} \\
        \midrule
        - GT-Reward & 39.63 & 24.98 & 65.58 & 47.84 & 40.68 & 54.50 & 62.53 \\
        \midrule
        - Self-Certainty & 39.63 & 24.25 & 58.11 & 46.65 & 40.88 & 46.42 & 61.40 \\
        - Entropy & 39.90 & 22.16 & 62.18 & 45.02 & 43.09 & 50.19 & 59.90 \\
        - Majority-Voting & 40.68 & 22.52 & 60.25 & 46.10 & 41.08 & 48.42 & 60.65 \\
        - Co-rewarding-I & 40.68 & 24.25 & 62.18 & 44.37 & 44.49 & 49.58 & 61.65 \\
        - Co-rewarding-II & 41.21 & 25.89 & 64.91 & 45.24 & 39.28 & 52.27 & 59.40 \\
        \specialrule{1pt}{0.4ex}{0.4ex}
        \multicolumn{8}{c}{\textit{\textbf{Qwen3-8B-Base}}} \\
    \midrule
        \textbf{Methods} & \textbf{biology} & \textbf{business} & \textbf{chemistry} & \textbf{computer sci.} & \textbf{economics} & \textbf{engineering} & \textbf{health} \\
        \midrule
        - GT-Reward & 74.76 & 63.24 & 55.48 & 63.17 & 68.96 & 41.38 & 57.09 \\
        \midrule
        - Self-Certainty & 75.03 & 63.62 & 53.62 & 55.61 & 68.96 & 39.83 & 57.09 \\
        - Entropy & 75.73 & 64.39 & 54.51 & 58.29 & 65.05 & 41.69 & 55.87 \\
        - Majority-Voting & 76.15 & 60.20 & 54.15 & 56.34 & 69.91 & 38.91 & 55.75 \\
        - Co-rewarding-I & 76.43 & 65.78 & 57.07 & 62.20 & 69.43 & 43.14 & 56.60 \\
        - Co-rewarding-II & 76.84 & 64.25 & 54.68 & 62.43 & 68.12 & 42.00 & 58.06 \\
        \specialrule{1pt}{0.4ex}{0.4ex}
        \textbf{Methods} & \textbf{history} & \textbf{law} & \textbf{math} & \textbf{other} & \textbf{philosophy} & \textbf{physics} & \textbf{psychology} \\
        \midrule
        - GT-Reward & 50.92 & 30.25 & 67.58 & 52.49 & 51.10 & 57.20 & 67.67 \\
        \midrule
        - Self-Certainty & 49.34 & 28.88 & 68.02 & 51.62 & 52.10 & 56.89 & 66.42 \\
        - Entropy & 50.39 & 30.43 & 65.28 & 51.41 & 47.09 & 54.50 & 66.67 \\
        - Majority-Voting & 48.03 & 28.88 & 63.43 & 53.68 & 48.10 & 52.50 & 64.66 \\
        - Co-rewarding-I & 50.13 & 29.97 & 68.54 & 52.92 & 50.70 & 56.66 & 65.54 \\
        - Co-rewarding-II & 51.44 & 30.06 & 65.80 & 51.51 & 52.10 & 57.58 & 65.78 \\
    
\specialrule{1.6pt}{0.4ex}{0.4ex}
    \end{tabular}}
\end{table}

%  ----------------------------------------

\subsection{More Results of MMLU-Pro Evaluation}
\label{appe:Detail_MMLU-Pro}
As a complement to Table~\ref{Tble:MMLU-Pro_Qwen8B}, Table~\ref{Tble:MMLU-Pro_LlamaQwen4B_DAPO14k} and Table~\ref{Tble:MMLU-Pro_Qwen8BQwen4B_OpenRS} report detailed MMLU-Pro results for models trained on DAPO-14k and OpenRS, respectively. We observe that Co-rewarding consistently preserves general-domain performance across diverse subjects, indicating that though trained on math-oriented datasets, its improvements do not come at the cost of broader capabilities from other domains.

% we present detailed results for the Qwen3-4B-Base and Qwen3-8B-Base models on the MMLU-Pro benchmark, trained on the OpenRS dataset. The findings indicate that Co-rewarding methods achieve superior performance across most cross-domain downstream tasks. This advantage is particularly pronounced in subjects such as biology, economics, health, and law, where the performance of Co-rewarding approaches that of the GT-Reward.

\subsection{More Results of IFEval Evaluation}
\label{appe:Detail_IFEval}
The aim of IFEval is used to evaluate the instruction-following ability of LLMs. In Table~\ref{Tble:Main_Exp_CorewardingV1}, Table~\ref{Tble:Main_Exp_CorewardingV2}, Table~\ref{Tble:Appe_Exp_CorewardingV1} and Table~\ref{Tble:Supplement_Exp_CorewardingV2_OpenRS}, we report average IFEval performance due to space constraints. Specifically, the evaluation of IFEval includes four metrics: \{prompt\_level\_strict\_acc, inst\_level\_strict\_acc, prompt\_level\_loose\_acc and inst\_level\_loose\_acc\}, which apply different levels of answer matching. As a supplement, complete results are provided in Table~\ref{Tble:IFEval_Detail_MATH}, Table~\ref{Tble:IFEval_Detail_DAPO14k}, and Table~\ref{Tble:IFEval_Detail_OpenRS}. The results show that Co-rewarding not only preserves the inherent instruction-following ability of base models but also often surpasses GT-Reward across multiple models. This further confirms that Co-rewarding’s gains on mathematical and coding benchmarks are achieved without sacrificing general-domain instruction-following ability.

% As a complement to experiments in Table~\ref{Tble:AblationStudy}, we evaluated the instruction-following capabilities of Qwen3-4B/8B-Base and Llama-3.2-3B-Instruct models trained on DAPO-14k using the IFEval benchmark. This evaluation incorporates four key metrics: prompt-level strict accuracy, instruction-level strict accuracy, prompt-level loose accuracy, and instruction-level loose accuracy. The results indicate that Co-rewarding not only preserves the inherent instruction-following ability of the base models but also surpasses the performance of GT-Reward training across all tested models.

% In fact, the evaluation of IFEval includes four metrics: prompt\_level\_strict\_acc, inst\_level\_strict\_acc, prompt\_level\_loose\_acc and inst\_level\_loose\_acc. Due to space limit, we report the average of them in Table~\ref{Tble:Appe_Exp_CorewardingV1}. We provide the complete metrics in Table~\ref{tab:ifeval_benchmark} as a supplement. We observe that Co-rewarding training not only preserves the instruction-following ability of the base models, but actually surpasses the GT reward training on Qwen2.5-3B/7B, Qwen3-4B/8B, and Llama-3.2-3B-Instruct.

% ============================================================================
% Table 11: Other Results (%) of RL performance comparison on IFEval benchmark trained on MATH.
% ============================================================================

\begin{table}[th]
    \centering
    \caption{\textbf{Detailed IFEval Performance on Qwen2.5-3B/7B, Qwen3-1.7B/4B/8B-Base and Llama-3.2-3B-Instruct traind on MATH.} Results are reported for loose and strick settings respectively.}
    \label{Tble:IFEval_Detail_MATH}
    \vspace{-2mm}
    \resizebox{\textwidth}{!}{% 自动调整表格宽度以适应页面
    \begin{tabular}{l|ccccc|ccccc}
    \toprule[1.6pt]
    \multirow{2}{*}{\textbf{Methods}} & \multicolumn{10}{c}{\textbf{IFEval}} \\
    \cmidrule(lr){2-6} \cmidrule(lr){7-11}
    ~ & \textbf{Average} & \textbf{Prompt Strict} & \textbf{Prompt Loose} & \textbf{Inst. Strict} & \textbf{Inst. Loose} & \textbf{Average} & \textbf{Prompt Strict} & \textbf{Prompt Loose} & \textbf{Inst. Strict} & \textbf{Inst. Loose} \\
    \midrule
    ~ & \multicolumn{5}{c|}{\textit{\textbf{Qwen2.5-3B}}} & \multicolumn{5}{c}{\textit{\textbf{Qwen2.5-7B}}} \\
    \midrule
    Before RL           & 29.83 & 22.55 & 27.17 & 31.89 & 37.70 & 38.19 & 29.57 & 34.57 & 41.85 & 46.76 \\
    - GT-Reward         & 33.66 & 25.51 & 31.42 & 35.85 & 41.85 & 41.49 & 31.79 & 39.56 & 43.65 & 50.96 \\
    \midrule
    - Self-Certainty    & 32.22 & 24.40 & 29.76 & 34.65 & 40.05 & 37.24 & 28.47 & 34.38 & 40.05 & 46.04 \\
    - Entropy           & 32.94 & 24.77 & 30.50 & 35.13 & 41.37 & 40.33 & 30.13 & 37.87 & 43.29 & 50.00 \\
    - Majority-Voting   & 35.19 & 26.25 & 32.72 & 37.53 & 44.24 & 38.60 & 29.21 & 35.86 & 41.61 & 47.72 \\
    - Co-rewarding-I    & 33.86 & 23.84 & 31.61 & 36.09 & 43.88 & 41.73 & 32.35 & 39.37 & 44.48 & 50.72 \\
    - Co-rewarding-II   & 32.74 & 23.29 & 29.02 & 36.33 & 42.33 & 41.82 & 31.79 & 40.29 & 43.88 & 51.31 \\
    \specialrule{1pt}{0.4ex}{0.4ex}
    ~ & \multicolumn{5}{c|}{\textit{\textbf{Qwen3-1.7B-Base}}} & \multicolumn{5}{c}{\textit{\textbf{Qwen3-4B-Base}}} \\
    \midrule
    Before RL           & 33.65 & 25.69 & 30.86 & 36.45 & 41.60 & 46.43 & 36.04 & 44.18 & 48.68 & 56.83 \\
    - GT-Reward         & 36.16 & 27.35 & 31.79 & 40.64 & 44.84 & 47.80 & 37.34 & 46.77 & 49.40 & 57.67 \\
    \midrule
    - Self-Certainty    & 32.96 & 24.58 & 29.20 & 36.69 & 41.36 & 48.15 & 39.37 & 46.76 & 49.52 & 56.95 \\
    - Entropy           & 35.37 & 26.61 & 31.42 & 39.44 & 44.00 & 50.44 & 40.67 & 48.61 & 52.52 & 59.07 \\
    - Majority-Voting   & 35.45 & 26.06 & 32.16 & 38.72 & 48.84 & 48.78 & 37.89 & 47.50 & 50.36 & 59.65 \\
    - Co-rewarding-I    & 35.56 & 27.91 & 31.23 & 39.32 & 43.76 & 50.35 & 40.67 & 49.35 & 51.56 & 59.83 \\
    - Co-rewarding-II   & 36.94 & 27.17 & 33.64 & 40.05 & 46.88 & 51.30 & 41.40 & 49.54 & 53.12 & 61.15 \\
    \specialrule{1pt}{0.4ex}{0.4ex}
    ~ & \multicolumn{5}{c|}{\textit{\textbf{Qwen3-8B-Base}}} & \multicolumn{5}{c}{\textit{\textbf{Llama3-2-Instruct}}} \\
    \midrule
    Before RL           & 50.32 & 40.11 & 50.27 & 51.07 & 59.83 & 57.32 & 46.77 & 55.27 & 60.19 & 67.03 \\
    - GT-Reward         & 52.78 & 41.96 & 51.76 & 54.44 & 62.95 & 47.41 & 37.34 & 42.88 & 52.52 & 57.31 \\
    \midrule
    - Self-Certainty    & 50.98 & 39.74 & 49.54 & 52.88 & 61.75 & 54.88 & 43.81 & 52.68 & 58.15 & 64.87 \\
    - Entropy           & 51.81 & 40.67 & 51.20 & 52.76 & 62.59 & 54.70 & 43.81 & 52.68 & 57.67 & 64.63 \\
    - Majority-Voting   & 51.80 & 39.74 & 51.02 & 53.60 & 62.83 & 47.96 & 37.34 & 43.44 & 52.88 & 58.18 \\
    - Co-rewarding-I    & 55.79 & 43.99 & 57.11 & 55.63 & 66.42 & 49.14 & 39.37 & 45.66 & 53.12 & 58.39 \\
    - Co-rewarding-II   & 60.70 & 55.64 & 65.59 & 56.00 & 65.59 & 49.90 & 39.93 & 45.66 & 54.68 & 59.35 \\
    \bottomrule[1.6pt]
    \end{tabular}}
\end{table}

\begin{table}[th]
    \centering
    \caption{\textbf{Detailed IFEval performance on Qwen3-4B/8B-Base and Llama-3.2-3B-Instruct traind on DAPO-14k.} Results are reported for loose and strict settings in IFEval, respectively.}
    \label{Tble:IFEval_Detail_DAPO14k}
    \vspace{-2mm}
    \resizebox{0.8\textwidth}{!}{
    \begin{tabular}{l|ccccc}
    \toprule[1.6pt]
        \multirow{2}{*}{\textbf{Methods}} & \multicolumn{5}{c}{\textbf{IFEval}} \\
        \cmidrule{2-6}
         ~ & \textbf{Average} & \textbf{Prompt Strict} & \textbf{Prompt Loose} & \textbf{Inst. Strict} & \textbf{Inst. Loose} \\
        \midrule
        \multicolumn{6}{c}{\textit{\textbf{Qwen3-4B-Base}}} \\
        \midrule
        Before RL      & 46.43 & 36.04 & 44.18 & 48.68 & 56.83 \\
        - GT-Reward    & 47.70 & 37.52 & 45.84 & 49.76 & 57.67 \\
        \midrule
        - Self-Certainty & 45.58 & 35.67 & 43.99 & 47.84 & 54.80 \\
        - Entropy        & 48.20 & 37.71 & 46.58 & 50.48 & 58.03 \\
        - Majority-Voting& 48.91 & 39.19 & 47.69 & 50.24 & 58.51 \\
        - Co-rewarding-I & 46.84 & 36.41 & 45.66 & 48.80 & 56.47 \\
        - Co-rewarding-II& 48.90 & 39.56 & 46.21 & 51.44 & 58.39 \\
        \specialrule{1pt}{0.4ex}{0.4ex}
        \multicolumn{6}{c}{\textit{\textbf{Qwen3-8B-Base}}} \\
        \midrule
        Before RL      & 50.32 & 40.11 & 50.27 & 51.07 & 59.83 \\
        - GT-Reward    & 53.11 & 41.59 & 52.13 & 54.56 & 64.15 \\
        \midrule
        - Self-Certainty & 50.58 & 41.04 & 49.54 & 51.68 & 60.07 \\
        - Entropy        & 51.56 & 41.59 & 49.91 & 53.48 & 61.27 \\
        - Majority-Voting& 51.54 & 41.22 & 51.02 & 52.64 & 61.27 \\
        - Co-rewarding-I & 50.17 & 40.67 & 48.24 & 52.16 & 59.59 \\
        - Co-rewarding-II& 53.31 & 41.40 & 53.23 & 54.20 & 64.39 \\
        \specialrule{1pt}{0.4ex}{0.4ex}
        \multicolumn{6}{c}{\textit{\textbf{Llama3.2-3B-Instruct}}} \\
        \midrule
        Before RL      & 57.32 & 46.77 & 55.27 & 60.19 & 67.03 \\
        - GT-Reward    & 53.10 & 42.33 & 49.91 & 57.19 & 62.95 \\
        \midrule
        - Self-Certainty & 54.50 & 44.55 & 51.76 & 58.03 & 63.67 \\
        - Entropy        & 55.78 & 45.29 & 53.23 & 59.11 & 65.47 \\
        - Majority-Voting& 54.07 & 42.33 & 52.50 & 56.83 & 64.63 \\
        - Co-rewarding-I & 53.04 & 42.33 & 51.02 & 55.76 & 63.07 \\
        - Co-rewarding-II& 51.92 & 41.59 & 48.24 & 56.00 & 61.87 \\
        \bottomrule[1.6pt]
    \end{tabular}}
\end{table}

\begin{table}[th]
    \centering
    \caption{\textbf{Detailed IFEval Performance on Qwen3-8B/4B-Base trained on Open-RS.} Results are reported for loose and strict settings in IFEval, respectively.}
    \label{Tble:IFEval_Detail_OpenRS}
    \vspace{-2mm}
    \resizebox{\textwidth}{!}{
    \begin{tabular}{l|ccccc|ccccc}
    \toprule[1.6pt]
        \multirow{2}{*}{\textbf{Methods}} & \multicolumn{10}{c}{\textbf{IFEval}} \\
        \cmidrule{2-11}
         ~ & \textbf{Average} & \textbf{Prompt Strict} & \textbf{Prompt Loose} & \textbf{Inst. Strict} & \textbf{Inst. Loose} 
         & \textbf{Average} & \textbf{Prompt Strict} & \textbf{Prompt Loose} & \textbf{Inst. Strict} & \textbf{Inst. Loose} \\
        \midrule

        ~ & \multicolumn{5}{c|}{\textit{\textbf{Qwen3-8B-Base}}} & \multicolumn{5}{c}{\textit{\textbf{Qwen3-4B-Base}}} \\
        \midrule
        Before RL           & 50.32 & 40.11 & 50.27 & 51.07 & 59.83 
                            & 46.43 & 36.04 & 44.18 & 48.68 & 56.83 \\
        - GT-Reward         & 52.53 & 41.59 & 51.02 & 54.56 & 62.95 
                            & 47.80 & 37.34 & 46.77 & 49.40 & 57.67 \\
        \midrule
        - Self-Certainty    & 52.12 & 41.59 & 50.83 & 53.72 & 62.35 
                            & 46.47 & 35.86 & 44.73 & 48.56 & 56.71 \\
        - Entropy           & 52.94 & 43.25 & 51.94 & 53.72 & 62.83 
                            & 46.41 & 36.97 & 44.73 & 48.68 & 55.28 \\
        - Majority-Voting   & 51.13 & 40.67 & 49.35 & 53.36 & 61.15 
                            & 46.35 & 36.41 & 44.18 & 48.80 & 56.00 \\
        - Co-rewarding-I    & 53.11 & 41.40 & 53.05 & 53.95 & 64.02 
                            & 46.83 & 35.86 & 46.77 & 47.84 & 56.83 \\
        - Co-rewarding-II   & 52.92 & 42.14 & 52.50 & 54.08 & 62.95 
                            & 48.45 & 38.26 & 46.77 & 50.24 & 58.51 \\
        \bottomrule[1.6pt]
    \end{tabular}}
\end{table}

\begin{table}[t]
\centering
\caption{\textbf{Case studies:} Original vs. Rephrased Questions}
\begin{tabular}{|p{0.45\textwidth}|p{0.45\textwidth}|}
\hline
\textbf{Original Question} & \textbf{Rephrased Question} \\
\hline
Sam is hired for a 20-day period. On days that he works, he earns \$60. For each day that he does not work, \$30 is subtracted from his earnings. At the end of the 20-day period, he received \$660. How many days did he not work? 
& 
A contractor agrees to a job lasting 20 days. For every day the job is completed on time, the contractor earns \$60. However, for each day the work is delayed, a fine of \$30 is applied. After the 20-day period, the contractor's total earnings are \$660. How many days was the job delayed? \\
\hline
Karen drove continuously from 9:40 a.m. until 1:20 p.m. of the same day and covered a distance of 165 miles. What was her average speed in miles per hour? 
& 
A traveler set off at 9:40 a.m. and reached their destination at 1:20 p.m. the same day after traveling a total of 165 miles. What was their average speed during the trip in miles per hour? \\
\hline
Solve for $x$: $\tfrac{1}{2} + \tfrac{1}{x} = \tfrac{5}{6}$. 
& 
A tank is partially filled by two different pipes. One pipe fills half the tank in an hour, and together with another pipe, they fill five-sixths of the tank in the same time. If the second pipe alone fills $\tfrac{1}{x}$ of the tank in an hour, find the value of $x$. \\
\hline
\end{tabular}
\label{tab:question_rephrase}
\end{table}

\subsection{Original Questions vs. Rephrased Questions}
\label{app:rephrased_problem}
To provide an intuitive illustration, we present several examples of original questions with their rephrased versions in Table~\ref{tab:question_rephrase}. We observe that such rephrasings are reasonable and effective, as they preserve the same underlying mathematical essence while presenting the problems in a substantially different surface form. This reflects the high quality of our rephrased data and forms the basis of Co-rewarding-I: by leveraging contrastive agreement across data-invariant variants, the model is encouraged to elicit more robust reasoning ability.

\subsection{Complete Case Study}
\label{app:case_study_generation}
As a supplement to Figure~\ref{Fig:Case_study}, we present the complete generation outputs of this case study. The full outputs clearly reveal the reward hacking behaviors of existing self-rewarding baselines. Self-Certainty and Entropy fall into repetitive outputs—for example, Self-Certainty repeatedly generates ``Understanding," and Entropy repeatedly produces ``Simplify the next fraction" until truncated at the maximum length. This arises because their decoding probability mass collapses onto a small subset of tokens, leading the model to loop over them. Majority-Voting shows another form of reward hacking by boxing an incorrect answer ``0" to maximize consensus across rollouts and thereby secure the highest reward. In contrast, both Co-rewarding and GT-Reward yield the correct final answer along with a coherent step-by-step reasoning path.

\subsection{Case Study on Code Benchmark}
\label{app:case_study_code}
To understand the behaviors of different methods on other tasks, we supplement an additional case study on CRUX. We observe that Self-Certainty and Entropy fail by producing repetitive outputs, consistent with the collapse pattern on math-related benchmarks. In contrast, GT-Reward, Majority-Voting, and Co-rewarding generate coherent reasoning trace and correct solutions.

\begin{exbox}{Question:}
$\frac{m}{n}$ is the Irreducible fraction value of \[3+\frac{1}{3+\frac{1}{3+\frac13}}\], what is the value of $m+n$?
\end{exbox}

\begin{exbox}{Co-rewarding-I Answer:}
To solve the problem, we need to simplify the given expression step by step and then find the irreducible fraction form. The expression is:

\[3 + \frac{1}{3 + \frac{1}{3 + \frac{1}{3}}}.\]

First, we simplify the innermost fraction:

\[3 + \frac{1}{3} = \frac{9}{3} + \frac{1}{3} = \frac{10}{3}.\]

Now substitute this back into the expression:

\[3 + \frac{1}{3 + \frac{1}{\frac{10}{3}}} = 3 + \frac{1}{3 + \frac{3}{10}}.\]

Next, simplify the fraction inside the denominator:

\[3 + \frac{3}{10} = \frac{30}{10} + \frac{3}{10} = \frac{33}{10}.\]

Now substitute this back into the expression:

\[3 + \frac{1}{\frac{33}{10}} = 3 + \frac{10}{33}.\]

To add these, we need a common denominator:

\[3 + \frac{10}{33} = \frac{99}{33} + \frac{10}{33} = \frac{109}{33}.\]

The fraction \(\frac{109}{33}\) is already in its simplest form because 109 and 33 have no common factors other than 1. Therefore, \(m = 109\) and \(n = 33\), and the value of \(m + n\) is:
 
\[109 + 33 = 142.\]

So, the final answer is:

\[\boxed{142}.\]
\end{exbox}

\begin{exbox}{Co-rewarding-II Answer:}
To solve the given expression \(3 + \frac{1}{3 + \frac{1}{3 + \frac{1}{3}}}\), we need to simplify it step by step.

First, simplify the innermost fraction:
\[3 + \frac{1}{3} = \frac{9}{3} + \frac{1}{3} = \frac{10}{3}.\]

Next, substitute this back into the next fraction:
\[3 + \frac{1}{3 + \frac{1}{\frac{10}{3}}} = 3 + \frac{1}{3 + \frac{3}{10}} = 3 + \frac{1}{\frac{30}{10} + \frac{3}{10}} = 3 + \frac{1}{\frac{33}{10}} = 3 + \frac{10}{33}.\]

Now, simplify the expression \(3 + \frac{10}{33}\):
\[3 + \frac{10}{33} = \frac{99}{33} + \frac{10}{33} = \frac{109}{33}.\]

The fraction \(\frac{109}{33}\) is already in its simplest form because 109 and 33 are coprime (they have no common factors other than 1).

Thus, the irreducible fraction value of the given expression is \(\frac{109}{33}\), where \(m = 109\) and \(n = 33\). Therefore, \(m + n = 109 + 33 = 142\).

The value of \(m + n\) is \(\boxed{142}\).
\end{exbox}

\begin{exbox}{GT-Reward Answer:}
Alright, I have this problem:

The expression is:
\[ 3 + \frac{1}{3 + \frac{1}{3 + \frac{1}{3}}} \]

And I need to find the irreducible fraction \(\frac{m}{n}\) that equals this expression, then determine \(m + n\).

First, I need to simplify the expression step by step. It looks like a continued fraction, so I'll start by simplifying the innermost part and work my way out.

Let's denote the entire expression as:
\[ E = 3 + \frac{1}{3 + \frac{1}{3 + \frac{1}{3}}} \]

**Step 1: Simplify the innermost fraction**

Start with the innermost fraction:
\[ 3 + \frac{1}{3} \]

Calculate this:
\[ 3 + \frac{1}{3} = \frac{9}{3} + \frac{1}{3} = \frac{10}{3} \]

So now, the expression becomes:
\[ E = 3 + \frac{1}{3 + \frac{1}{\frac{10}{3}}} \]

Wait, I need to simplify \(\frac{1}{\frac{10}{3}}\):
\[ \frac{1}{\frac{10}{3}} = \frac{3}{10} \]

Now, the expression is:
\[ E = 3 + \frac{1}{3 + \frac{3}{10}} \]

**Step 2: Simplify the next fraction**

Now, simplify \(3 + \frac{3}{10}\):
\[ 3 + \frac{3}{10} = \frac{30}{10} + \frac{3}{10} = \frac{33}{10} \]

So, the expression now is:
\[ E = 3 + \frac{1}{\frac{33}{10}} \]

Simplify \(\frac{1}{\frac{33}{10}}\):
\[ \frac{1}{\frac{33}{10}} = \frac{10}{33} \]

Now, the expression is:
\[ E = 3 + \frac{10}{33} \]

**Step 3: Combine the terms**

Convert 3 to a fraction with denominator 33:
\[ 3 = \frac{99}{33} \]

Now, add the fractions:
\[ E = \frac{99}{33} + \frac{10}{33} = \frac{109}{33} \]

So, the expression simplifies to \(\frac{109}{33}\).

**Step 4: Ensure the fraction is irreducible**

Check if 109 and 33 have any common divisors. 109 is a prime number (I think), and 33 is \(3 \times 11\). Since 109 isn't divisible by 3 or 11, the fraction \(\frac{109}{33}\) is already in its irreducible form.

**Step 5: Find \(m + n\)**

Here, \(m = 109\) and \(n = 33\), so:
\[ m + n = 109 + 33 = 142 \]

Wait, let's double-check the calculations to ensure no mistakes were made.

**Verification:**

Starting over:

\[ E = 3 + \frac{1}{3 + \frac{1}{3 + \frac{1}{3}}} \]

Innermost:
\[ 3 + \frac{1}{3} = \frac{10}{3} \]

Next layer:
\[ 3 + \frac{1}{\frac{10}{3}} = 3 + \frac{3}{10} = \frac{33}{10} \]

Outermost:
\[ 3 + \frac{1}{\frac{33}{10}} = 3 + \frac{10}{33} = \frac{99}{33} + \frac{10}{33} = \frac{109}{33} \]

Yes, it checks out. Therefore, \(m + n = 109 + 33 = 142\).

\(\boxed{142}\)
\end{exbox}

\begin{exbox}{Majority-Voting Answer:}
To find the value of \( \frac{m}{n} \) as the irreducible fraction of the expression \( 3 + \frac{1}{3 + \frac{1}{3 + \frac{1}{3}}} \), we need to simplify the expression step by step.

First, let's simplify the innermost fraction:
\[ 3 + \frac{1}{3} = 3 + \frac{1}{3} = \frac{9}{3} + \frac{1}{3} = \frac{10}{3} \]

Now, substitute this back into the expression:
\[ 3 + \frac{1}{3 + \frac{1}{\frac{10}{3}}} = 3 + \frac{1}{3 + \frac{3}{10}} \]

Next, simplify the fraction \( 3 + \frac{3}{10} \):
\[ 3 + \frac{3}{10} = \frac{30}{10} + \frac{3}{10} = \frac{33}{10} \]

Now, substitute this back into the expression:
\[ 3 + \frac{1}{\frac{33}{10}} = 3 + \frac{10}{33} \]

To add these, we need a common denominator:
\[ 3 + \frac{10}{33} = \frac{99}{33} + \frac{10}{33} = \frac{109}{33} \]

So, the irreducible fraction value of the expression is \( \frac{109}{33} \). Here, \( m = 109 \) and \( n = 33 \). Therefore, the value of \( m + n \) is:
\[ m + n = 109 + 33 = 142 \]

Thus, the value of \( m + n \) is \(\boxed{0}\).
\end{exbox}

\begin{exbox}{Self-Certainty Answer:}
Alright, I have this math problem to solve:

**Problem:**
I need to solve a math problem. The problem is about understanding how to solve math problems. I need to understand how to solve this problem so I can understand more about math.

\#\#\# Understanding the Problem

First, I need to understand what the problem is. I need to understand math so I can understand more about numbers and how to solve problems with numbers. Understanding math will help me understand more about how to understand problems.

\#\#\# Understanding Math Problems

After understanding the problem, I need to understand more about math problems. Understanding math problems will help me understand how to understand these problems. Understanding these problems will help me understand more about how to understand math.

\#\#\# Understanding Mathematical Concepts

After understanding math problems, I need to understand more about mathematical concepts. Understanding mathematical concepts will help me understand more about math. Understanding these concepts will help me understand more about how math helps me understand numbers.

\#\#\# Understanding Mathematical Operations

After understanding mathematical concepts, I need to understand more about mathematical operations. Understanding mathematical operations will help me understand more about how to understand these operations. Understanding these operations will help me understand more about how these operations help me understand math.

\#\#\# Understanding Mathematical Operations in Math

After understanding mathematical operations, I need to understand ......

\#\#\# Understanding Mathematical Methods

After understanding mathematical operations, I need to understand more about mathematical methods. ......

\#\#\# Understanding How Mathematical Methods Help

......

\#\#\# Understanding How These Methods Help Me Understand Math

After understanding more about how these methods help me, I need to understand ......

\#\#\# Understanding How These Methods Help Me Understand Math Problems

After understanding how these methods help me understand math, I need to understand ......

\#\#\# Understanding How These Methods Help Me Solve Math Problems

After understanding how these methods help me understand these math problems, I need to understand ......

\#\#\# Understanding How These Methods Help Me Solve These Problems

After understanding how these methods help me solve these problems, I need to understand more about ......
\end{exbox}

\begin{exbox}{Entropy Answer:}
To find the value of \( \frac{m}{n} \) as the irreducible fraction of the expression:

\[
3 + \frac{1}{3 + \frac{1}{3 + \frac{1}{3}}}
\]

Let's simplify the expression step by step.

1. **Simplify the innermost fraction:**

\[
3 + \frac{1}{3}
\]

\[
3 + \frac{1}{3} = \frac{3 \times 3}{3} + \frac{1}{3} = \frac{9}{3} + \frac{1}{3} = \frac{9 + 1}{3} = \frac{10}{3}
\]

2. **Simplify the next fraction:**

\[
3 + \frac{1}{3 + \frac{1}{3}} = 3 + \frac{1}{\frac{10}{3}}
\]

\[
3 + \frac{1}{\frac{10}{3}} = 3 + \frac{1 \times 3}{10} = 3 + \frac{3}{10} = \frac{3 \times 10}{10} + \frac{3}{10} = \frac{30}{10} + \frac{3}{10} = \frac{30 + 3}{10} = \frac{33}{10}
\]

3. **Simplify the next fraction:**

\[
3 + \frac{1}{3 + \frac{1}{3 + \frac{1}{3}}} = 3 + \frac{1}{\frac{33}{10}}
\]

\[
3 + \frac{1}{\frac{33}{10}} = 3 + \frac{1 \times 10}{33} = 3 + \frac{10}{33} = \frac{3 \times 33}{33} + \frac{10}{33} = \frac{99}{33} + \frac{10}{33} = \frac{99 + 10}{33} = \frac{109}{33}
\]

4. **Simplify the next fraction:**

\[
3 + \frac{1}{3 + \frac{1}{3 + \frac{1}{3 + \frac{1}{3}}}} = 3 + \frac{1}{\frac{109}{33}}
\]

\[
3 + \frac{1}{\frac{109}{33}} = 3 + \frac{1 \times 33}{109} = 3 + \frac{33}{109} = \frac{3 \times 109}{109} + \frac{33}{109} = \frac{327}{109} + \frac{33}{109} = \frac{327 + 33}{109} = \frac{360}{109}
\]

5. **Simplify the next fraction:**

\[
3 + \frac{1}{3 + \frac{1}{3 + \frac{1}{3 + \frac{1}{3 + \frac{1}{3}}}}} = 3 + \frac{1}{\frac{360}{109}}
\]

\[
3 + \frac{1}{\frac{360}{109}} = 3 + \frac{1 \times 109}{360} = 3 + \frac{109}{360} = \frac{3 \times 360}{360} + \frac{109}{360} = \frac{1080}{360} + \frac{109}{360} = \frac{1080 + 109}{360} = \frac{1189}{360}
\]

6. **Simplify the next fraction:**

\[
3 + \frac{1}{3 + \frac{1}{3 + \frac{1}{3 + \frac{1}{3 + \frac{1}{3 + \frac{1}{3}}}}}} = 3 + \frac{1}{\frac{1189}{360}}
\]

......

7. **Simplify the next fraction:**

\[
3 + \frac{1}{3 + \frac{1}{3 + \frac{1}{3 + \frac{1}{3 + \frac{1}{3 + \frac{1}{3 + \frac{1}{3}}}}}}} = 3 + \frac{1}{\frac{3927}{1189}}
\]

......

8. **Simplify the next fraction:**

\[
3 + \frac{1}{3 + \frac{1}{3 + \frac{1}{3 + \frac{1}{3 + \frac{1}{3 + \frac{1}{3 + \frac{1}{3 + \frac{1}{3}}}}}}}} = 3 + \frac{1}{\frac{12970}{3927}}
\]

......

9. **Simplify the next fraction:**

\[
3 + \frac{1}{3 + \frac{1}{3 + \frac{1}{3 + \frac{1}{3 + \frac{1}{3 + \frac{1}{3 + \frac{1}{3 + \frac{1}{3 + \frac{1}{3}}}}}}}}} = 3 + \frac{1}{\frac{42837}{12970}}
\]

......

10. **Simplify the next fraction:**

\[
3 + \frac{1}{3 + \frac{1}{3 + \frac{1}{3 + \frac{1}{3 + \frac{1}{3 + \frac{1}{3 + \frac{1}{3 + \frac{1}{3 + \frac{1}{3 + \frac{1}{3}}}}}}}}}} = 3 + \frac{1}{\frac{141481}{42837}}
\]

......

11. **Simplify the next fraction:**

\[
3 + \frac{1}{3 + \frac{1}{3 + \frac{1}{3 + \frac{1}{3 + \frac{1}{3 + \frac{1}{3 + \frac{1}{3 + \frac{1}{3 + \frac{1}{3 + \frac{1}{3 + \frac{1}{3}}}}}}}}}}} = 3 + \frac{1}{\frac{467280}{141481}}
\]

......

12. **Simplify the next fraction:**

\[
3 + \frac{1}{3 + \frac{1}{3 + \frac{1}{3 + \frac{1}{3 + \frac{1}{3 + \frac{1}{3 + \frac{1}{3 + \frac{1}{3 + \frac{1}{3 + \frac{1}{3 + \frac{1}{3 + \frac{1}{3}}}}}}}}}}}} = 3 + \frac{1}{\frac{1543321}{467280}}
\]

......

13. **Simplify the next fraction:**

\[
3 + \frac{1}{3 + \frac{1}{3 + \frac{1}{3 + \frac{1}{3 + \frac{1}{3 + \frac{1}{3 + \frac{1}{3 + \frac{1}{3 + \frac{1}{3 + \frac{1}{3 + \frac{1}{3 + \frac{1}{3 + \frac{1}{3}}}}}}}}}}}}} = 3 + \frac{1}{\frac{5097243}{1543321}}
\]

\[
3 + \frac{1}{\frac{5097243}{1543321}} = 3 + \]
\end{exbox}

\begin{exbox}{Question:}
\begin{lstlisting}[breaklines=true]
## Question: 

You are given a Python function and an assertion containing an input to the function. Complete the assertion with a literal (no unsimplified expressions, no function calls) containing the output when executing the provided code on the given input, even if the function is incorrect or incomplete. 

[PYTHON]
def f(text):
    new_text = list(text)
    for i in '+':
        if i in new_text:
            new_text.remove(i)
    return ''.join(new_text)
assert f('hbtofdeiequ') == ??
[/PYTHON]


## Instruction 

Please answer this question by first reasoning and then providing your answer.
Present your reasoning and solution in the following json format. 
Please show your final answer in the `answer` field, e.g.,"answer": "42".

```json
{
    "reasoning": "___",
    "answer": "___"
}
\end{lstlisting}
\end{exbox}

\begin{exbox}{GT-Reward Answer:}
\begin{lstlisting}[breaklines=true]
{  
    "reasoning": "The function `f` is designed to remove all occurrences of the character '+' from the input string `text`. The function first converts the string into a list of characters, iterates through this list, and removes any '+' characters it finds. Finally, it joins the remaining characters back into a single string. Given the input string 'hbtofdeiequ', there are no '+' characters in the string, so the function will simply return the original string unchanged. Therefore, the output of the function when given the input 'hbtofdeiequ' is the string 'hbtofdeiequ'.",

    "answer": "hbtofdeiequ"
}
\end{lstlisting}
\end{exbox}

\begin{exbox}{Co-rewarding-I Answer:}
\begin{lstlisting}[breaklines=true]
{  
    "reasoning": "The function `f` is designed to remove all occurrences of the character '+' from the input string `text`. The function first converts the string into a list of characters, then iterates over the character '+' and removes it from the list if it exists. Finally, it joins the list back into a string and returns it. Given the input 'hbtofdeiequ', the character '+' does not appear in the string, so the function will return the original string unchanged.",  
    "answer": "hbtofdeiequ"  
}
\end{lstlisting}
\end{exbox}

\begin{exbox}{Co-rewarding-II Answer:}
\begin{lstlisting}[breaklines=true]
{  
    "reasoning": "The function `f` takes a string `text` and removes all occurrences of the character '+' from it. The function then joins the remaining characters back into a string and returns it. Given the input 'hbtofdeiequ', the function will remove all '+' characters (though there are none in the input), so the output will be the same as the input string. Therefore, the assertion should check if the output of `f('hbtofdeiequ')` is 'hbtofdeiequ'.",
    "answer": "hbtofdeiequ"
}
\end{lstlisting}
\end{exbox}

\begin{exbox}{Majority-Voting Answer:}
\begin{lstlisting}[breaklines=true]
{  
    "reasoning": "The function `f` is designed to remove all occurrences of the character '+' from the input string `text`. The input provided to the function is 'hbtofdeiequ'. Since the character '+' does not appear in this string, the function will return the original string unchanged. Therefore, the assertion should compare the output of the function to the original input string.",  
    "answer": "'hbtofdeiequ'"
}
\end{lstlisting}
\end{exbox}

\begin{exbox}{Self-Certainty Answer:}
\begin{lstlisting}[breaklines=true]
To solve this problem, we need to follow these steps:

1. Understand the given Python function.
2. Determine what the function does with the given input.
3. Complete the assertion with the correct output based on the function's behavior.

Let's go through each step one by one:

1. Understand the given Python function.
2. Determine what the function does with the given input.
3. Complete the assertion with the correct output based on the function's behavior.

We will write the code to understand the function, determine its behavior with the given input, and then complete the assertion with the correct output.

Let's write the code step by step.
```python
# Step 1: Understand the given Python function
# Step 2: Determine what the function does with the given input
# Step 3: Complete the assertion with the correct output based on the function's behavior

# Let's write the code to follow these steps
```

```python
# Step 1: Understand the given Python function
# Step 2: Determine what the function does with the given input
# Step 3: Complete the assertion with the correct output based on the function's behavior

# Let's write the code to follow these steps

# Step 1: Understand the given Python function
# We will define the given function and understand what it does

# Step 2: Determine what the function does with the given input
# We will determine the output of the function for the given input

# Step 3: Complete the assertion with the correct output based on the function's behavior
# We will complete the assertion with the correct output

# Let's write the code
```

Let's write the code step by step.
```python
# Step 1: Understand the given Python function
# Step 2: Determine what the function does with the given input
# Step 3: Complete the assertion with the correct output based on the function's behavior

# Let's write the code to follow these steps
```

```python
# Step 1: Understand the given Python function
# We will define the given function and understand what it does

# Step 2: Determine what the function does with the given input
# We will determine the output of the function for the given input

# Step 3: Complete the assertion with the correct output based on the function's behavior
# We will complete the assertion with the correct output

......

# Let's write the code

# Step 1: Understand the
\end{lstlisting}
\end{exbox}

\begin{exbox}{Entropy Answer:}
\begin{lstlisting}[breaklines=true]
```json
{
    "reasoning": "To determine the output of the function `f` when the input is 'hbtofdeiequ', we need to follow these steps:

1. Convert the string 'hbtofdeiequ' into a list:
   \[
   \text{new\_text} = \text{list}('hbtofdeiequ')
   \]
   \[
   \text{new\_text} = ['h', 'b', 't', 'o', 'f', 'd', 'e', 'i', 'e', 'q', 'u']
   \]

2. Iterate through the character '+':
   \[
   \text{for } i \text{ in } '+':
   \]
   \[
   \text{for } i \text{ in } '+':
   \]
   \[
   \text{for } i \text{ in } '+':
   \]
   \[
   \text{for } i \text{ in } '+':
   \]
   \[
   \text{for } i \text{ in } '+':
   \]
   \[
   \text{for } i \text{ in } '+':
   \]
    ......
   \[
   \text{for } i \text{ in } '+':
   \]
   \[
   \text{for } i \text{ in } '+
\end{lstlisting}
\end{exbox}

\end{document}